\documentclass[journal,10pt,twocolumn,twoside,compsoc]{IEEEtran}

\usepackage{booktabs}
\usepackage[lofdepth,lotdepth]{subfig}
\usepackage{epsfig}
\usepackage{graphicx}
\usepackage{xcolor,colortbl}
\usepackage{hyperref}
\usepackage{cancel}
\usepackage{etoolbox}
\usepackage{multirow}
\usepackage{amsfonts,amssymb,amsmath}
\usepackage{review}

\setrevision{0}
\newcommand{\comRev}[1]{#1}

\definecolor{darkgreen}{gray}{0.0}

\ifthenelse{\equal{\therevision}{1}}{\newcommand{\rowcolorrevone}{white}}{\newcommand{\rowcolorrevone}{white}}
\ifthenelse{\equal{\therevision}{2}}{\newcommand{\rowcolorrevoneSec}{darkgreen!20}}{\newcommand{\rowcolorrevoneSec}{white}}

\def\model{\bf}

\makeatletter
\patchcmd{\@verbatim}
  {}{}
  \makeatother

\definecolor{mygray}{gray}{0.7}
\newcommand{\ccg}{\cellcolor{mygray}}

\ifCLASSOPTIONcompsoc
  \usepackage[nocompress]{cite}
\else
  \usepackage{cite}
\fi

\ifCLASSINFOpdf

\else

\fi

\usepackage{ragged2e}

\begin{document}

%\title{Practical Guidelines for Deep Regression}
\title{A Comprehensive Analysis of Deep Regression}

\author{St\'{e}phane~Lathuili\`{e}re,
  Pablo~Mesejo,
  Xavier~Alameda-Pineda, \IEEEmembership{Member~IEEE},
  and~Radu~Horaud% <-this % stops a space
\IEEEcompsocitemizethanks{\IEEEcompsocthanksitem All authors are with the PERCEPTION team, INRIA Grenoble Rh\^{o}ne-Alpes and Universit\'e Grenoble Alpes. Pablo Mesejo is also with the Andalusian Research Institute in Data Science and Computational Intelligence (DaSCI), University of Granada. E-mail: firstname.lastname@inria.fr
%\protect\\
\IEEEcompsocthanksitem EU funding via the FP7 ERC Advanced Grant VHIA \#340113 is greatly acknowledged.
}% <-this % stops a space

}

\markboth{DRAFT}%
{S. Lathuili\`{e}re, P. Mesejo, X. Alameda-Pineda and R. Horaud: A Comprehensive Analysis of Deep Regression}

\makeatletter
\if@twocolumn
\newcommand{\whencolumns}[2]{
#2
}
\else
\newcommand{\whencolumns}[2]{
#1
}
\fi
\makeatother

\IEEEtitleabstractindextext{%
\begin{abstract}
Deep learning revolutionized data science, and recently its popularity has grown exponentially, as did the amount of papers employing deep networks. Vision tasks, such as human pose estimation, did not escape from this trend. There is a large number of deep models, where small changes in the network architecture, or in the data pre-processing, together with the stochastic nature of the optimization procedures, produce notably different results, making extremely difficult to sift methods that significantly outperform others. This situation motivates the current study, in which we perform a systematic evaluation and statistical analysis of vanilla deep regression, i.e. convolutional neural networks with a linear regression top layer. This is the first comprehensive analysis of deep regression techniques. We perform experiments on four vision problems, and report confidence intervals for the median performance as well as the statistical significance of the results, if any. Surprisingly, the variability due to different data pre-processing procedures generally eclipses the variability due to modifications in the network architecture. Our results reinforce the hypothesis according to which, in general, a general-purpose network (e.g. VGG-16 or ResNet-50) adequately tuned can yield results close to the state-of-the-art without having to resort to more complex and ad-hoc regression models.
\end{abstract}
\begin{IEEEkeywords}
Deep Learning, Regression, Computer Vision, Convolutional Neural Networks, Statistical Significance, Empirical and Systematic Evaluation, Head-Pose Estimation, Full-Body Pose Estimation, Facial Landmark Detection.
\end{IEEEkeywords}}

\maketitle

\IEEEdisplaynontitleabstractindextext

\IEEEpeerreviewmaketitle

\section{Introduction}
\label{sec:introduction}
%\ifCLASSOPTIONcompsoc
%\IEEEraisesectionheading{\section{Introduction}\label{sec:introduction}}
%\else

%\section{Introduction}
%\label{sec:introduction}
%\fi
%

\IEEEPARstart{R}{egression} 
techniques are widely employed to solve tasks where the goal is to predict continuous values. In computer vision, regression techniques span a large ensemble of applicative scenarios such as: head-pose estimation~\cite{fanelli2011real,zhu2012face}, facial landmark detection~\cite{burgos2013robust,dantone2012real}, human pose estimation~\cite{agarwal20043d,sun2012conditional}, age estimation~\cite{Yan07,guo2008image}, or image registration~\cite{chou20132d,niethammer2011geodesic}. 
 
For the last decade, deep learning architectures have overwhelmingly outperformed the state-of-the-art in many traditional computer vision tasks such as image classification~\cite{Krizhevsky2012,Szegedy2015} or object detection~\cite{Girshick2014,Sermanet2014}, and have also been applied  to newer tasks such as the analysis of image virality~\cite{alameda2017viraliency}. Roughly speaking, these architectures consist of several convolutional layers, generally followed by a few fully-connected layers, and a classification softmax layer with, for instance, a cross-entropy  loss; the overall architecture is referred to as convolutional neural network (ConvNet). Besides classification,  ConvNets are also used to solve regression problems. In this case, the softmax layer is commonly replaced with a fully connected regression layer with linear or sigmoid activations. We refer to such architectures as \textit{vanilla deep regression}. Following this approach, many deep models were proposed, obtaining state-of-the-art results in classical vision regression problems such as head pose estimation~\cite{Liu2016}, human pose estimation~\cite{Toshev2014,Belagiannis2015} or facial landmark detection~\cite{Sun2013}. Interestingly, \emph{vanilla deep regression} is often used as the main building block in cascaded approaches, where such regression methods are iteratively used to refine the predictions. Indeed, \cite{Toshev2014,Sun2013,Belagiannis2015} demonstrate the effectiveness of this strategy. Likewise for robust regression~\cite{lathuiliere2018deepgum}, often implying probabilistic models~\cite{hueber2015speaker}. Recently, more complex computer vision methods exploiting regression include multi-stage deep regression using clustering, auto-routing and pseudo-labels for deep fashion landmark detection~\cite{liu16FashionLandmark}, inverse piece-wise affine probabilistic model for head pose estimation~\cite{Lathuiliere2017} or a head-map estimator from which facial landmarks are extracted~\cite{bulat2017far}. However, the architectures developed in these studies are designed for a specific task, and how to adapt them to a different task is a non-trivial research question.

Regression could also be formulated as classification~\cite{rothe2016deep,Rogez_2017_CVPR}. In that case, the output space is generally discretized in order to obtain class labels, and a multi-class loss is minimized. However, this approach suffers from important drawbacks. First, there is an inherent trade-off between the complexity of the optimization problem and the accuracy of the method due to the discretization process. Second, in absence of a specific formulation, a confusion between two classes has the same cost independently of the corresponding error in the target space. Last, the discretization method that is employed must be designed specifically for each task, and therefore general-purpose methodologies must be used with care. In this study, we focus on generic vanilla deep regression so as to escape from the definition of a task-dependent trade-off between complexity and accuracy, as well as from a penalization loss that may not take the spatial proximity of the labels into account.
% 
% Conversely, a regression model, trained with a $L_2$ loss, penalizes more large errors than small errors. Last, the discretization method that is employed must be designed specifically for each task. For all these reasons, we do not include classification formulations in our study. %Interested readers are invited to refer to the literature review detailed below for more information about deep networks for classification. 

Thanks to the success of deep neural architectures, based on empirical validation, much of the scientific work currently available in computer vision exploits their representation power. Unfortunately, the immense majority of these works provide neither a statistical evaluation of the performance nor a rigorous justification of the methodological choices. The main consequence of this lack of systematic evaluation is that researchers proceed by trial-and-error experimentation because the scientific evidence behind the superiority of newly introduced techniques is neither sufficiently clear nor statistically grounded.  

There are a few studies devoted to the extensive and/or systematic evaluation of deep architectures, focusing on different aspects. For instance, seminal papers exploring efficient back-propagation strategies~\cite{LeCunBOM98}, or evaluating  ConvNets  for  visual recognition~\cite{Nebauer1998}, were already published twenty years ago. Some articles provide general guidance and understanding on appropriate architectural choices~\cite{SzegedyVISW16,SmithT16,IthapuRS17} or on gradient-based training strategies for deep architectures~\cite{bengio2012}, while others try to delve into the differences in performance of the variants of a specific (recurrent) model~\cite{GreffSKSS17}, or the differences in performance of several models applied to a specific problem~\cite{ChandrasekharLM16,MISHKIN2017,chatfield2014return}. Automatic ways to overcome the problem of choosing the optimal network architecture were also devised~\cite{SaxenaV16,XieY17}. Overall, there is very little guidance on the plethora of design choices and hyper-parameter settings for deep learning architectures, let alone ConvNets for regression problems. 

In summary, the analysis of prior work shows that, first, the absence of a systematic evaluation of deep learning advances in regression, and second, an over abundance of papers based on deep learning (for instance, $\approx2000$ papers were uploaded on arXiv only in March 2017, in the categories related to machine learning, computer vision and pattern recognition, computation and language, and neural and evolutionary computing). This highlights again the importance of serious comparative empirical studies to discern which are the key blocks in deep regression. Finally, in our opinion, there is a scientifically unjustified lack of statistical tests and confidence intervals in the experimental sections of the vast majority of works published in the field. Both the execution of a single run per benchmarked method and the succinct description of the preprocessing strategies being used limit the reliability of the results and method reproducibility. 

Even if some authors devote time and efforts to make their research reproducible~\cite{bulat2017far}, many published studies do not describe implementation, practical, or data pre-processing in detail~\cite{Sun2013,liu16FashionLandmark,Liu2016,Belagiannis2015}. One possible reason to explain this state of affairs may be the unavailability of systematic comparative procedures that highlight the most important details.
% The explanation for all this missing details could come from the fact that authors do not know which details to report without comparative studies that could highlight what are the most crucial ones.

We propose to fill in this gap with a systematic evaluation and a statistical analysis of the performance of vanilla deep regression for computer vision tasks. In order to conduct this study we take inspiration from two  recent papers. \cite{MISHKIN2017} presents an extensive evaluation of ConvNets on ImageNet for image categorization, including the type of non-linearity, pooling variants, network width, classifier design, image pre-processing, and learning parameters. %RADU: Does next phrase refer to~\cite{MISHKIN2017} or is it a general statement ? 
No statistical analysis accompanies the results reported in \cite{MISHKIN2017} such that one understands the differences in performance from a thorough statistical perspective. 
%RAdU: don't start with "on the other side"
\cite{GreffSKSS17} discusses the differences in performance of several variants of the long short-term memory based recurrent networks for categorical sequence modeling, thus addressing classification of sequential data: each variant is run several times, statistical tests are employed to compare each variant with a baseline architecture,  and the performance is discussed over box-plots offering some sort of graphical confidence intervals of the different benchmarked methods. 
%RADU: the motivation and contribution of the paper are not strong enough. The next phrase should be expanded into a paragraph.

% In particular, in this paper, we compare different ConvNet hyper parameters as in~\cite{MISHKIN2017}, using statistical tests (inspired from~\cite{GreffSKSS17}) for computer vision regression tasks: we test two different deep architectures (ResNet-50 and VGG-16) (Section \ref{sec:baselines}) on three different computer visions problems (head-pose estimation, facial landmark detection and full-body pose estimation) (Section \ref{sec:problems_datasets}), evaluate four different optimizers (adam, adagrad, rmsprop, adadelta) (Section \ref{sec:optimization}), four network variants (batch normalizer, dropout, pooling, layer to regress from, finetuning depth) (Section \ref{sec:variant}), several data preprocessing estrategies (Section \ref{sec:prepro}) . 

In this paper we carry out an in-depth analysis of vanilla deep regression methods on three standard computer vision problems: head pose estimation, facial landmark detection and full-body pose estimation. We first focus on the impact of different optimization techniques so as to guarantee that the networks are correctly optimized in the rest of the experiments. With the help of statistical tests we perform a systematic comparison between different network variants (e.g. the fine-tuning depth) and data pre-processing strategies. More precisely, we run each experiment five times and compute $95\%$ confidence intervals for the median performance as well as the Wilcoxon signed-rank test \cite{wilcoxon:test}. We claim that the combination of these two indicators is much more robust and reliable than the average performance over a single run as usually done in practice. Therefore, the main contribution is, not only a benchmarking methodology, but also the conclusions that stem out.  We compare the vanilla deep regression methods to some state-of-the-art regression algorithms specifically developed for the tasks just mentioned. Finally, we discuss which are the factors with high performance variability and which therefore should be a priority when exploring the use of deep regression in computer vision. 

The rest of the manuscript is organized as follows. Section~\ref{sec:protocol} discusses the experimental protocols (data sets and base architectures) used to benchmark the different choices. Section~\ref{sec:optimization} is devoted to find the best optimization strategy for each of the base architectures and data sets. The statistical tests (paired-tests and confidence intervals) are described in detail in Section~\ref{sec:stat}, and used to soundly benchmark the different network variants and data pre-processing strategies, respectively in Sections~\ref{sec:variant} and~\ref{sec:prepro}. We explore how deep regression is positioned with respect to the state-of-the-art in Section~\ref{sec:application}, before presenting an overall discussion in Section~\ref{sec:global_discussion}. Conclusions are drawn in Section~\ref{sec:conclusions}.

%DONE_RADU: paper organization is missing (it seems that there is some text below that was commented out)

%\input{related}
\section{Experimental Protocol}
\label{sec:protocol}
%\section{Experimental Protocol}
%\label{sec:protocol}
In this section, we describe the protocol adopted to evaluate the influence of different architectures and their variants, training strategies, and data pre-processing methods. We run the experiments using two common base architectures (see Section~\ref{sec:baselines}), on three standard computer vision problems that are often solved using regression (see Section~\ref{sec:problems_data sets}). The computational environment is described in~\ref{sec:comp_env}. %Importantly, the statistical tests we used are motivated and detailed in Section~\ref{sec:stat}.

\subsection{Base Architectures} % Baselines
\label{sec:baselines}
We choose to perform our study using two architectures that are among the most commonly referred in the recent literature: VGG-16~\cite{simonyan2014very} and ResNet-50~\cite{he2016deep}. The VGG-16 model was termed \emph{Model D} in~\cite{simonyan2014very}. We prefer VGG-16 to AlexNet~\cite{Krizhevsky2012} because it performs significantly better on ImageNet and has inspired several network architectures for various tasks~\cite{Ren2015FasterRT,liu16FashionLandmark}. ResNet-50 performs even better than VGG-16 with shorter training time, which is nice in general, and in particular for benchmarking.

VGG-16 is composed of 5 blocks containing two or three convolution layers and a max pooling layer. Let $CB^i$ denote the $i^{th}$ convolution block (see~\cite{simonyan2014very} for the details on the number of layers and of units per layer, as well as the convolution parameters). Let $FC^i$ denote the $i^{th}$ fully connected layer with a dropout rate of $50\,$\%. $Fl$ denotes a flatten layer that transforms a 2D feature map into a vector and $SM$ denotes a soft-max layer. With these notations, VGG-16 can be written as $CB^1-CB^2-CB^3-CB^4-CB^5-Fl-FC^1-FC^2-SM$.

The main novelty of ResNet-50 is the use of identity shortcuts, which augment the network depth and reduce the number of parameters. We remark that all the convolutional blocks of ResNet-50, except for the first one,  have identity connections, making them \textit{residual} convolutional blocks. However, since we do not modify the original ResNet-50 structure, we denote them $CB$ as well. In addition, $GAP$ denotes a global average pooling layer. According to this notation, ResNet-50 architecture can be described as follows: $CB^1-CB^2-CB^3-CB^4-CB^5-GAP-SM$.

Both networks are initialized by training on ImageNet for classification, as it is usually done. We then remove the last soft-max layer $SM$, employed in the context of classification, and we replace it with a fully connected layer with linear activations equal in number to the dimension of the target space. Therefore this last layer is a regression layer, denoted $REG$, whose output dimension corresponds to the one of the target space of the problem at hand.  

%RADU: the subsection title is misleading, there is almost nothing about "problems". Maybe "Problems" deserves a dedicated paragraph that justifies the choice of the problems. I would also mention that this problems can be address with multi-class classifiers. A discussion of regression against multi-class classifiers is totally missing...
\subsection{Data Sets}
\label{sec:problems_data sets}
In order to perform empirical comparisons, we choose three challenging problems: head-pose estimation, facial landmark detection and human-body pose estimation. For these three problems we selected  four data sets that are widely used, and that favor diversity in output dimension, pre-processing and data-augmentation requirements.  

The \textbf{Biwi} head-pose data set~\cite{Fanelli2013} consists of over $15,000$ RGB-D images corresponding to video recordings of 20 people (16 men and  4 women, some of whom recorded twice) using a Kinect camera. It is one of the most widely used data set for head-pose estimation~\cite{wang2013head,Mukherjee2015,Liu2016,drouard2017robust,Lathuiliere2017}. During the recordings, the participants freely move their head and the corresponding head orientations lie in the intervals $[-60^{\circ}, 60^{\circ}]$ (pitch), $[-75^{\circ}, 75^{\circ}]$ (yaw), and $[-20^{\circ}, 20^{\circ}]$ (roll). Unfortunately, it would be too time consuming to perform cross-validation in the context of our statistical study. Consequently, we employed the split used in~\cite{Lathuiliere2017}. Importantly, none of the participants appears both in the training and test sets.

We also use the LFW and NET facial landmark detection (\textbf{FLD})
data sets~\cite{Sun2013} that consist of 5590 and 7876 face images, respectively. We combined both data sets and employed the same data partition as in~\cite{Sun2013}. Each face/image is labeled with the pixel coordinates of five key-points,
namely left and right eyes, nose, and left and right mouth corners. %% The detection error is measured with the 
%% Euclidean distance between the estimated and the ground truth position of the landmark, divided by the width of the 
%% face image, as in~\cite{Sun2013}. The performance is measured with the failure rate of each landmark, where errors 
%% larger than $5\%$ are counted as failures.

Third, we use the \textbf{Parse} data set~\cite{ramanan2007learning} that is a standard data set used for human pose estimation. It is a relatively small data set as it contains only 305 images. Therefore, this data set challenges very deep architectures and requires a data augmentation procedure. Each image is annotated with the pixel coordinates of 14 joints. In addition, given the limited size of Parse, we consider all possible image rotations in the interval $[12^\circ,-12^\circ]$ with a step of $0.5^\circ$. This procedure is applied only to the training images.

\addnote[mpii-desc]{1}{Finally, we also use the \textbf{MPII} data set~\cite{andriluka14cvpr}, that is a larger data set for human pose estimation. The data set contains around 25K images of over 40K people describing more than 400 human activities. This dataset represents a significant advance in terms of diversity and difficulty with respect to previous datasets for human body pose estimation. Importantly, the test set is not available for this dataset and score on the test set can be obtained only by submitting predictions to the MPII dataset research group. In addition, the number of submissions is limited. Therefore the MPII test set cannot be used in the context of this study. Therefore, we split the MPII training set into training, validation and test using $70\%$, $20\%$ and $10\%$ of the original training set. Following MPII protocol, models are evaluated on the test set, consdering only sufficiently separated individuals. Only mirroring is employed for data-augmentation in order to keep training time tractable in the context of our statistical analysis.}

\subsection{Computational Environment}
\label{sec:comp_env}
All our experiments were ran on an Nvidia TITAN X (Pascal generation) GPU with Keras 1.1.1 (on the Theano 0.9.0 backend), thus favoring an easy to replicate hardware configuration.
%, except for the statistical analyses that were performed with MATLAB on CPU. 
%We remark that our computational environment is \textit{academic}, thus far from being populated with tens of GPUs. 
This computational environment is used to compare the many possible choices studied in this paper. In total, we summarize the results of more \comRev{ than 1000 experimental  runs  ($\approx$110 days  of  GPU  time).} % At a first stage we focus on selecting the right technique to optimize the network.
The code and the results of individual runs are available online.\footnote{\url{https://team.inria.fr/perception/deep-regression/}}

\section{Network Optimization}
\label{sec:optimization}
%\section{Network Optimization}
%\label{sec:optimization}
% \doubt{At this point we have no idea what part of the network we are fine-tuning...}\\
% \doubt{Also, if I understand correctly, no statistical tests are used in this section, so I think we should move the statistical test description after network optimization, and treat these as ``preliminary'' experiments.}
%RADU: what is "the training optimization"? Do you mean the optimizer used for training? Also, it is subsumed that optimization corresponds to stochastic gradient descent, it is worth mentioning it and that the optimizers that are analyzed are software packages (routines?) implementing SGD. What is a batch size? 

Optimizers for training neural networks are responsible for finding the free parameters $\theta$ (usually denoted weights) of a cost function $J(\theta)$ that, typically, includes a performance measure evaluated on the training set and additional regularization terms. Such a cost (also called loss) function lets us quantify the quality of any particular set of weights. In this paper, network optimization and network fine-tuning are used as synonym expressions, since we always start from ImageNet pre-trained weights when fine-tuning on a particular new task. The positive impact of pre-training in deep learning has been extensively studied and demonstrated in the literature \cite{HinSal06,Erhan2010,Yosinski2014,Tajbakhsh2016}. 

Gradient descent, a first-order iterative optimization algorithm for finding the minimum of a function, is the most common and established method for optimizing neural network loss functions \cite{goodfellow2016deep}. There are other methods to train neural networks, from derivative-free optimization \cite{yao1999evolving,mesejo2015artificial} to second-order methods \cite{goodfellow2016deep}, but their use is much less widespread, so they have been excluded from this paper. %Also, second-order methods make use of second derivatives to improve optimization, but only networks with a very small number of parameters can be practically trained using these approaches \cite{goodfellow2016deep}.  
 Attending to the number of training examples used to evaluate the gradient of the loss function, we can distinguish between batch gradient descent (that employs the entire training set at each iteration), mini-batch gradient descent (that uses several examples in each iteration), and stochastic gradient descent (also called sometimes on-line gradient descent, that employs a single example at each iteration). In this paper, and as is common practice in the deep learning literature, we use a mini-batch gradient descent whose batch size is selected through the preliminary experimentation described in Section~\ref{sec:res-batch-size}.

Many improvements of the basic gradient descent algorithm have been proposed. In particular, the need to set a learning rate (step size) has been recognized as crucial and problematic: setting this parameter too high can cause the algorithm to diverge; setting it too low makes it slow to converge. In practice, it is generally beneficial to gradually decrease the learning rate over time \cite{goodfellow2016deep}. Recently, a number of algorithms with adaptive learning rates have been developed, and represent some of the most popular optimization algorithms actively in use.

\whencolumns{\begin{figure}[t]}{\begin{figure*}[t]}
  \centering
  \includegraphics[width=0.40\textwidth]{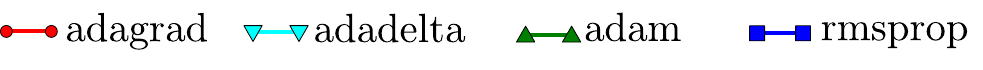}\\
  \vspace{-0.3cm}
  \subfloat[VGG-16 on Biwi]{\hspace{-2mm}
    \includegraphics[width=0.24\textwidth]{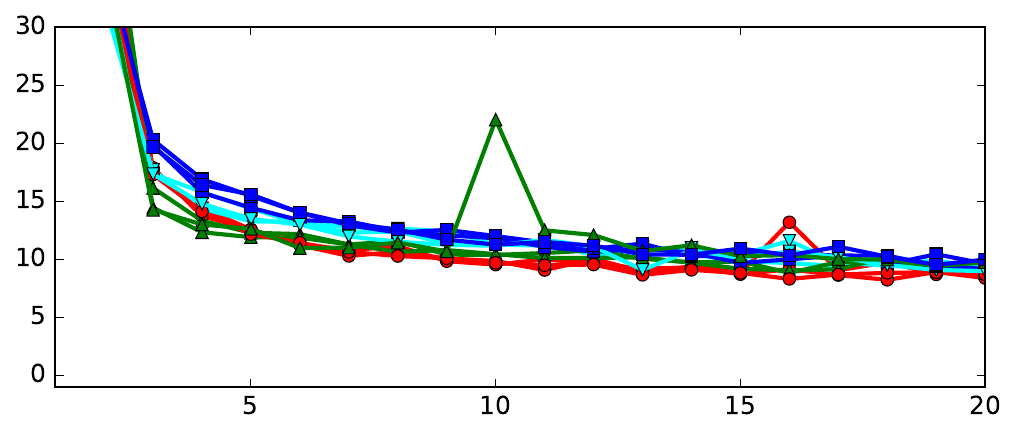}
    \label{lBiwi-VGG}
  }
  \subfloat[VGG-16 on FLD]{
  \includegraphics[width=0.24\textwidth]{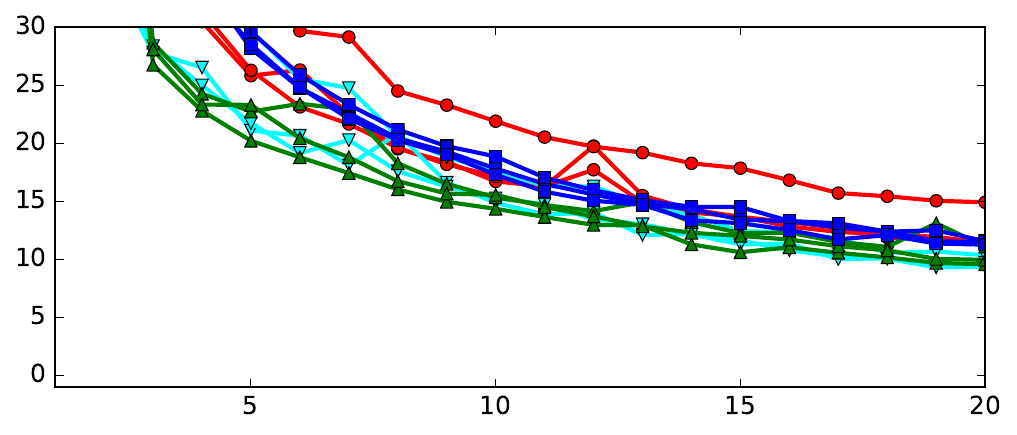}
    \label{lLand-VGG}
  }
  \subfloat[VGG-16 on Parse]{
  \includegraphics[width=0.24\textwidth]{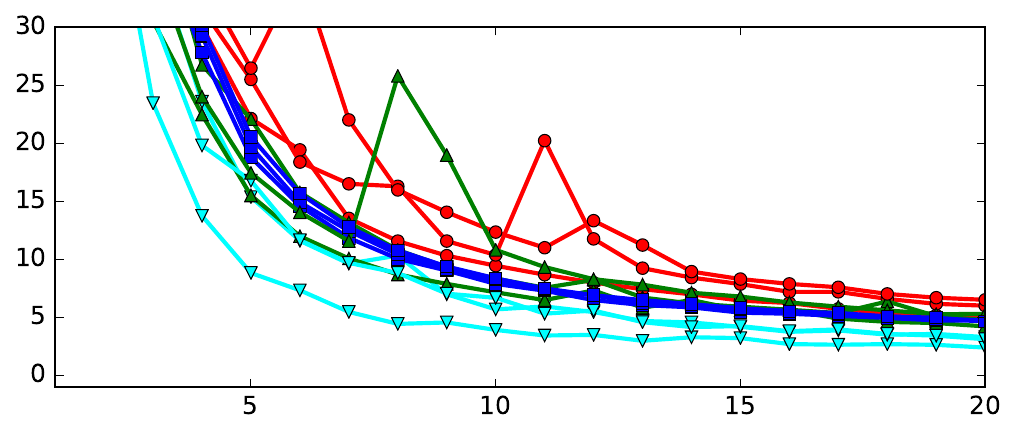}
    \label{lParse-VGG}
  }
  \subfloat[\comRev{VGG-16 on MPII}]{
  \includegraphics[width=0.24\textwidth]{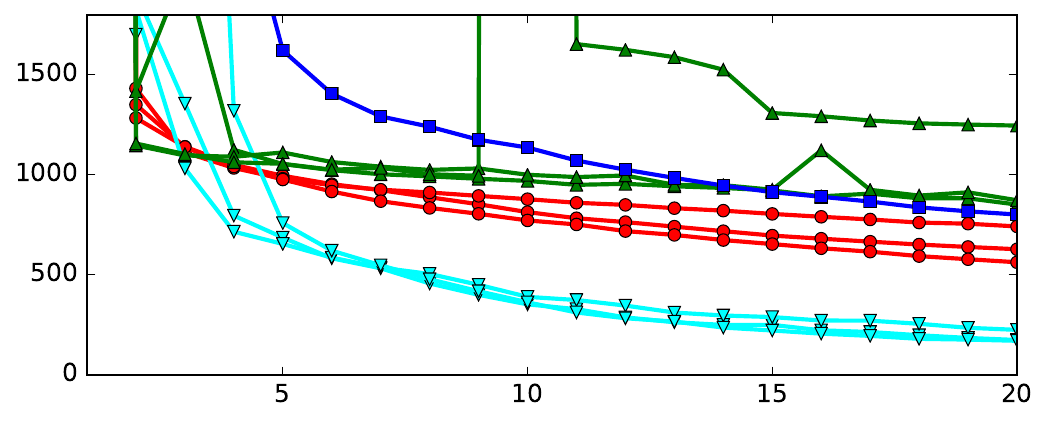}
    \label{lMpii-VGG}
  }\vspace{-4mm}\\
  \subfloat[ResNet-50 on Biwi]{\hspace{-2mm}
    \includegraphics[width=0.24\textwidth]{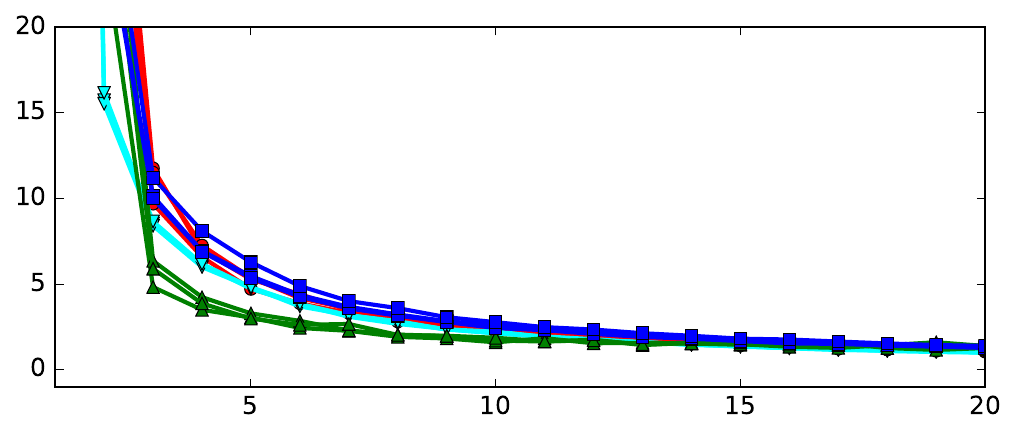}
    \label{lBiwi-ResNet}
  }
  \subfloat[ResNet-50 on FLD]{
  \includegraphics[width=0.24\textwidth]{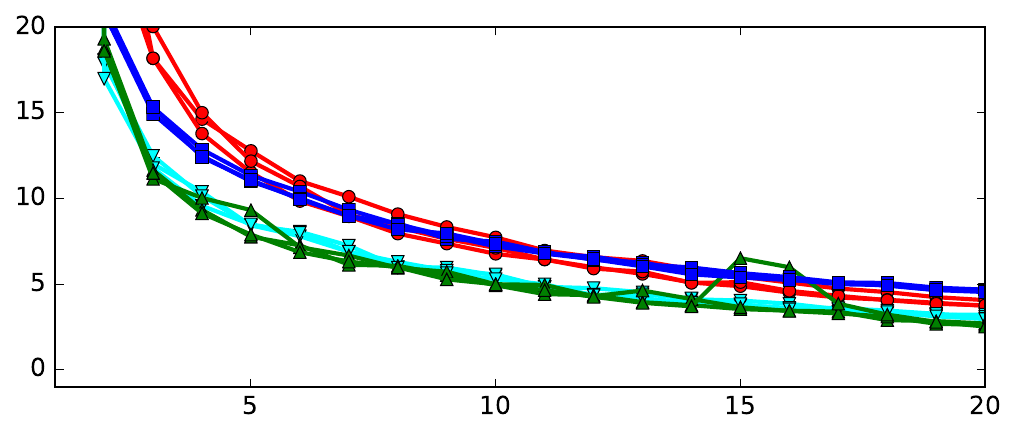}
    \label{lLand-ResNet}
  }
  \subfloat[ResNet-50 on Parse]{
  \includegraphics[width=0.24\textwidth]{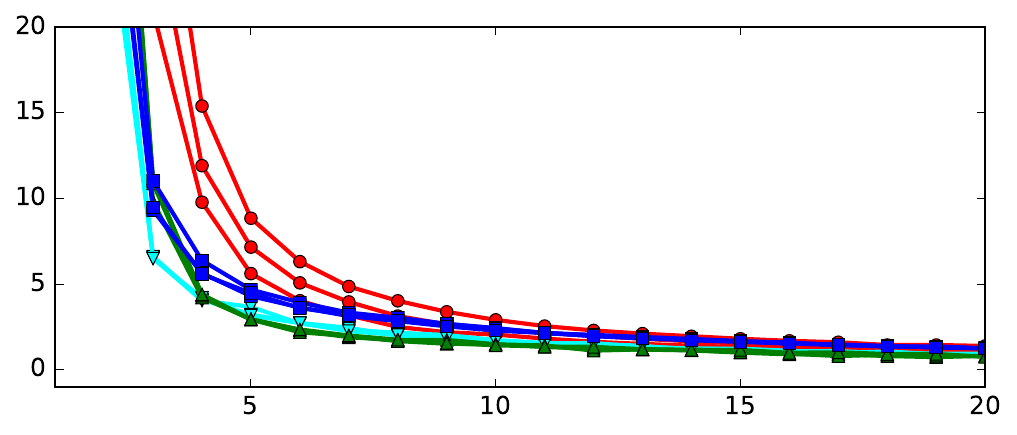}
    \label{lParse-ResNet}
  }
  \subfloat[\comRev{ResNet-50 on MPII}]{
  \includegraphics[width=0.24\textwidth]{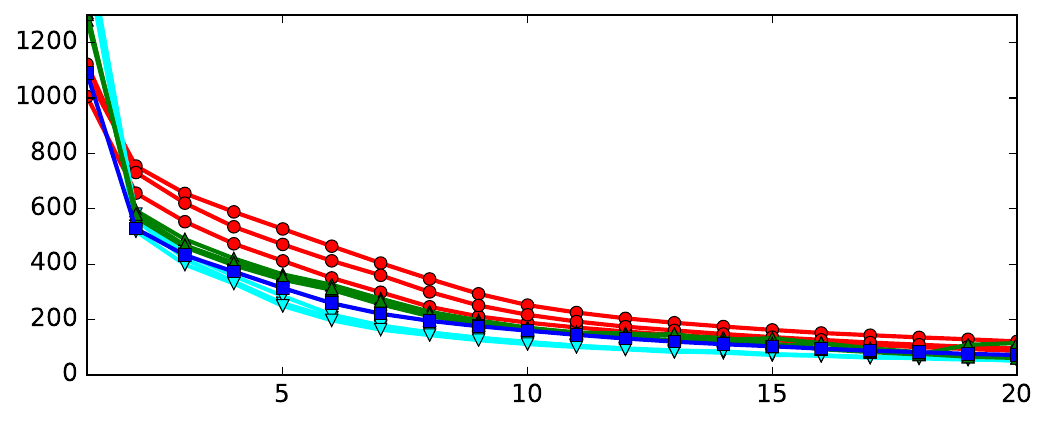}
    \label{lMpii-VGG}
  }\vspace{-2mm}
  \caption{Comparison of the training loss evolution with different optimizers for VGG-16 and ResNet-50.\vspace{-3mm}} 
% \label{fig:optimizers-VGG}
\label{fig:optimizers}
\whencolumns{\end{figure}}{\end{figure*}}

\whencolumns{\begin{figure}[t]}{\begin{figure*}[t]}
  \centering
  \includegraphics[width=0.40\textwidth]{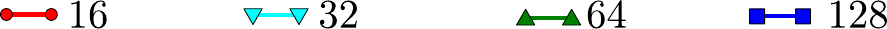}\\
  \vspace{-0.3cm}
  \subfloat[VGG-16 on Biwi]{\hspace{-2mm}
    \includegraphics[width=0.24\textwidth]{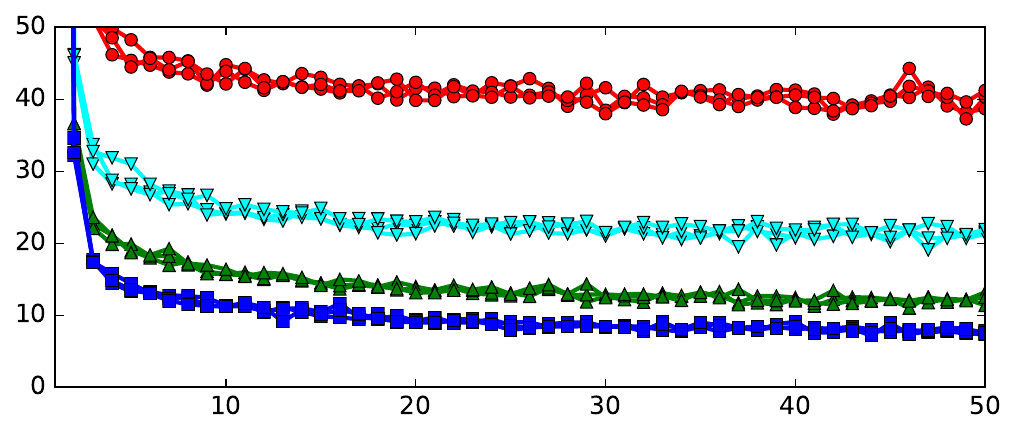}
    \label{lBiwiBS-VGG}
  }
  \subfloat[VGG-16 on FLD]{
  \includegraphics[width=0.24\textwidth]{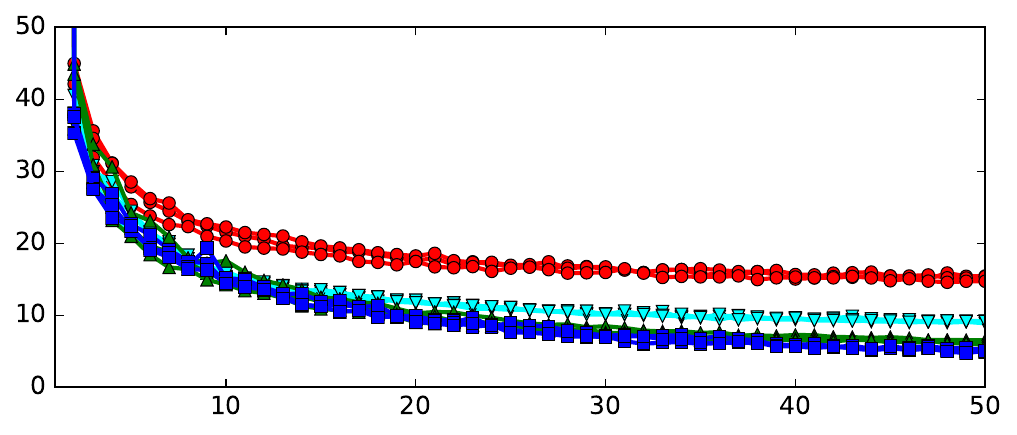}
    \label{lLandBS-VGG}
  }
  \subfloat[VGG-16 on Parse]{
  \includegraphics[width=0.24\textwidth]{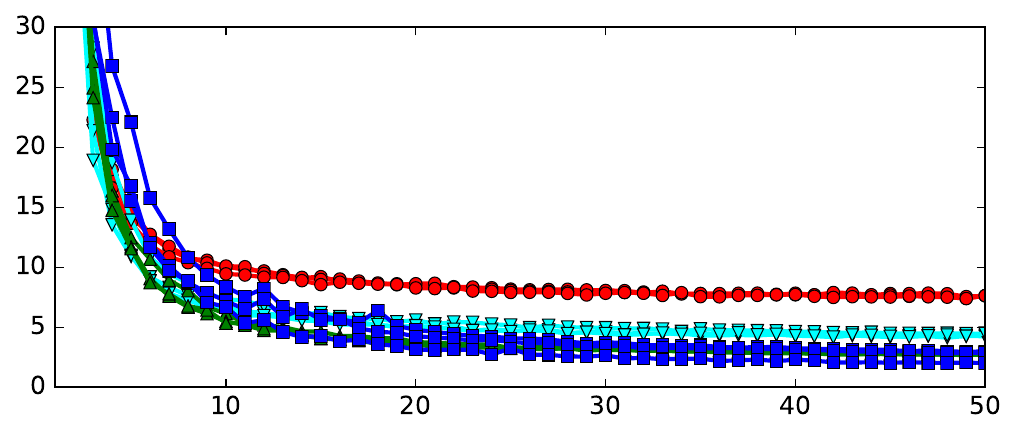}
    \label{lParseBS-VGG}
  } \subfloat[\comRev{VGG-16 on MPII}]{
  \includegraphics[width=0.24\textwidth]{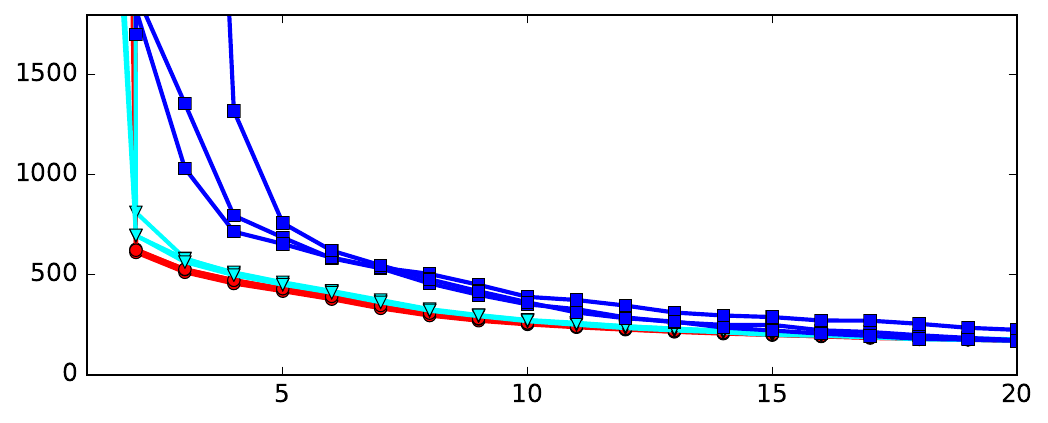}
    \label{lParseBS-VGG}
  }\vspace{-4mm}
  \subfloat[ResNet-50 on Biwi]{\hspace{-2mm}
    \includegraphics[width=0.24\textwidth]{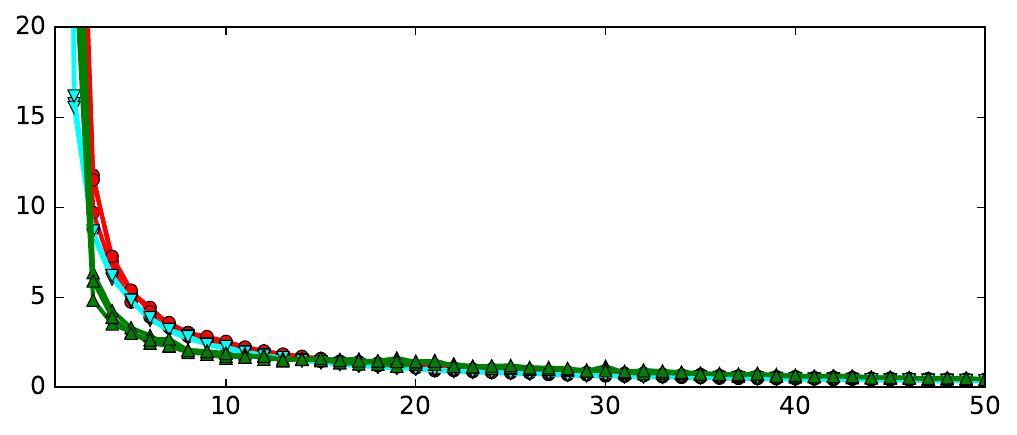}
    \label{lBiwiBS-ResNet}
  }
  \subfloat[ResNet-50 on FLD]{
  \includegraphics[width=0.24\textwidth]{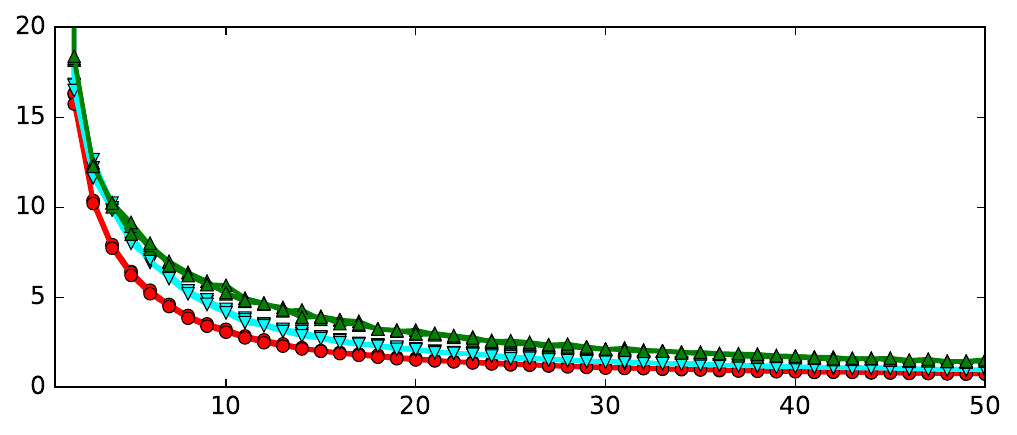}
    \label{lLandBS-ResNet}
  }
  \subfloat[ResNet-50 on Parse]{
  \includegraphics[width=0.24\textwidth]{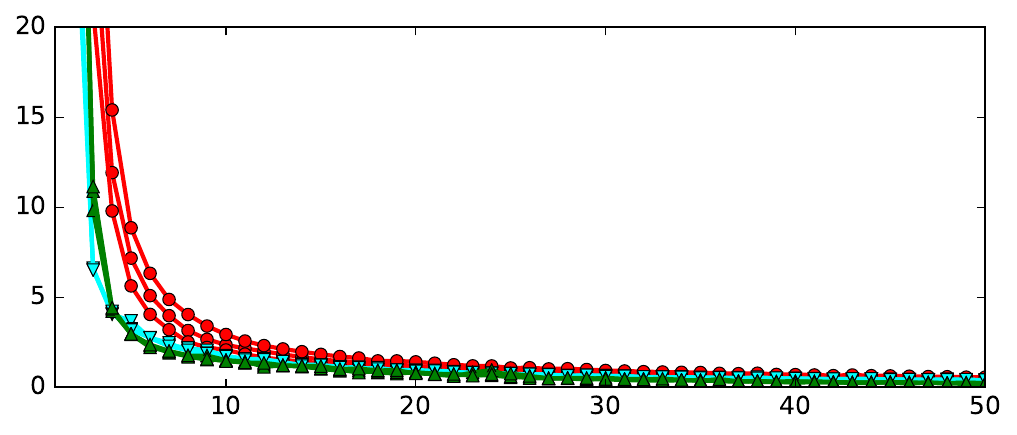}
    \label{lParseBS-ResNet}
  }\subfloat[\comRev{ResNet-50 on MPII}]{
  \includegraphics[width=0.24\textwidth]{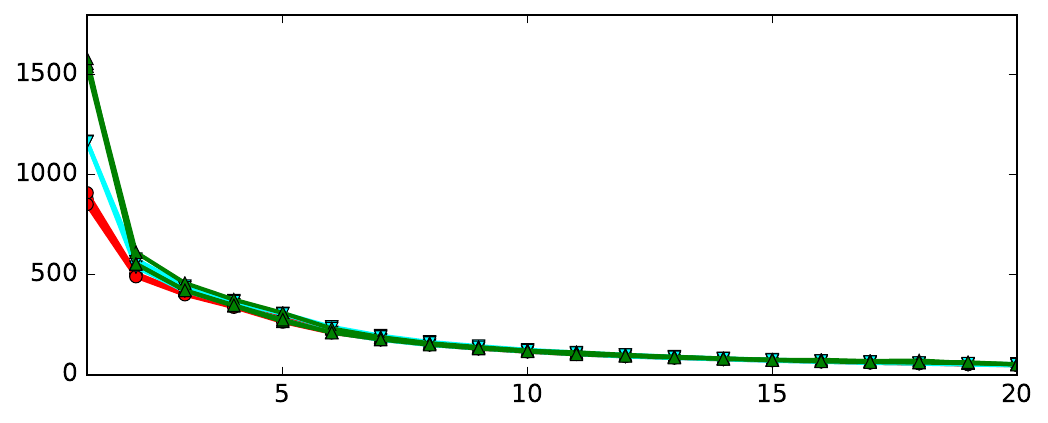}
    \label{lMpiiBS-ResNet}
  }\vspace{-2mm}
  \caption{Comparison of the training loss evolution with different
    batch size on VGG-16 and ResNet-50.\vspace{-3mm}}
\label{fig:batch}
\whencolumns{\end{figure}}{\end{figure*}}

\textbf{AdaGrad}~\cite{duchi2011adaptive}  (Adaptive Gradient) adapts the learning rate of every weight dividing it by the square root of the sum of their historical squared values. Weights with high gradients will have a rapid decrease of their learning rate, while weights with small or infrequent updates will have a relatively small decrease of their learning rate. 

\textbf{RMSProp}~\cite{tieleman2012lecture} (Root Mean Square Propagation)  modifies AdaGrad, to avoid lowering the learning rates very aggressively, by changing the gradient accumulation into an exponentially weighted moving average. AdaGrad shrinks the learning rate according to the entire history of the squared gradient, while RMSProp only considers recent gradients for that weight.

\textbf{AdaDelta}~\cite{zeiler2012adadelta} is also an extension of Adagrad, similar to RMSProp, that again dynamically adapts over time using only first order information and requires no manual tuning of a learning rate. 

\textbf{Adam}~\cite{kingma2014adam} (Adaptive Moments) is an update to the RMSProp optimizer where momentum \cite{sutskever13} is incorporated, i.e. in addition to store an exponentially decaying average of previous squared gradients (like in RMSProp), Adam also employs an exponentially decaying average of previous gradients (similar to momentum, where such moving average of previous gradients helps to dampen oscillations and to accelerate learning with intertia).

These four adaptive optimizers are evaluated in Section~\ref{sec:res-optimizer} and the two exhibiting the highest performance are used in the rest of the paper. We employ the mean square error (MSE) as the loss function to be optimized. As discussed in detail in Section~\ref{sec:variant}, VGG-16 and ResNet-50 are fine-tuned from (and including) the fifth and third convolutional block, respectively.

%In this section, we perform a preliminary study BLABLABLA, namely the optimizer and the batch size. We compare four different optimizers: AdaGrad~\cite{duchi2011adaptive}, AdaDelta~\cite{zeiler2012adadelta}, Adam~\cite{kingma2014adam}, RMSProp~\cite{tieleman2012lecture}. 
%RADU: I noticed that in this comment you refer to "choosing an optimization strategy". Indeed, it is not clear what you need by an optimizer, a routine that is called, a software package, etc. 
% Once an optimization strategy is chosen according to these experiments (see section \ref{sec:ResOptimizer}), we perform the other experiments related to network architecture and data pre-processing.{\color{red} I chose to speak about optimization here but should we have a separated subsection for that?}. In a second time, we compare different choices related to the network architecture and data pre-processing. Prior to that, we need to detail the problems and data sets we use (section \ref{sec:pbData}), the network variants we compare (section \ref{sec:variant}), the data pre-processing procedures (section \ref{sec:dataPreProc}), and the statistical method we chose for comparison (section \ref{sec:stat}).

%PABLO: This is the content you have to merge. In addition, the validation set is not $20\%$ of the train set for FLD, so you have to complete this info :)

\subsection{Impact of the Network Optimizer}
\label{sec:res-optimizer}

As  outlined above, we compare four optimizers: AdaGrad, AdaDelta, Adam and RMSProp. For each optimizer, we train the two networks (VGG-16 and ResNet-50) three times during 50 epochs with the default parameter values given in~\cite{chollet2015keras}. In this series of experiments, we choose a batch size of 128 and 64 for VGG-16 and ResNet-50, respectively (see Section~\ref{sec:res-batch-size}). \addnote[optim]{1}{The evolution of the loss value on the training set is displayed in Figure~\ref{fig:optimizers} for each one of the four data sets. We observe that, when employing VGG-16, the choice of the optimizer can lead to completely different loss values at convergence. Overall, we can state that the best training performance (in terms of loss value and convergence time) corresponds to AdaDelta and to Adam. 
In the light of the results described above, AdaGrad and RMSProp  do not seem to be the  optimizers of choice when training vanilla deep regression. While Adam and AdaDelta perform well, a comparative study prior to the selection of a particular optimizer is strongly encouraged since the choice may depend on the architecture and the problem at hand. Nevertheless, we observe that, with VGG-16, Adam is the best performing optimizer on FLD and one of the worst on MPII. Therefore, in case of optimization issues when tackling a new task, trying both Adam and AdaDelta seems to be a good option.}

\subsection{Impact of the Batch Size}
\label{sec:res-batch-size}

In this case, we test the previously selected optimizers (AdaDelta for VGG-16 and Adam for ResNet-50) with batch sizes of $16$, $32$, and $64$. A batch size of $128$ is tested on VGG-16 but not in ResNet-50 due to GPU memory limitations. We assess how the batch size impacts the optimization performance: Figure~\ref{fig:batch} shows the loss values obtained when training with different batch sizes.

First, we remark that the impact of the batch size in VGG-16 is more important than in ResNet-50. The latter is more robust to the batch size, yielding comparable results no matter the batch size employed. % both networks and in all the data sets, we observe a drastic impact of the batch size. This impact is clearly much bigger than the choice of the optimizer detailed previously. 
\addnote[BS]{1}{Second, except on MPII, we could conclude that using a larger batch size is a good heuristic towards good optimization (because in VGG-16 shows to be decisive, and in ResNet-50 does not harm performance). With VGG-16 on MPII, note that the batch size has an impact on convergence speed but not on the final loss value. With ResNet-50, the maximal batch size that can be used is 
%DONE_RADU: what is a "small GPU"? If you refer to memory, please repeat the word. Otherwise it's ambiguous. Unlike French (Spanish?) in English it is ok to repeat the same word for clarity.
%DONE_RADU: more slowly -->> slower. However, I that the whole phrase should be changed. What does it mean "train slower"? slow convergence rate? Be more precise.
constrained by the GPU memory being used. In our case, with an Nvidia TITAN X with 12 GB of memory, we could not train the ResNet-50 with a batch size of 128. 
% Therefore, a small GPU will not only train more slowly but will also show a worse performance. Obviously, this also depends on the number of network parameters being optimized: for ResNet-50 we could not test a batch size of 128 due to memory limitations. %, a batch size of 128 has not been evaluate it did not fit our GPU memory.
}
As a consequence, we choose 128 and 64 as batch sizes for VGG-16 and ResNet-50, respectively, and all subsequent experiments are performed using these batch sizes. As it is commonly found in the literature, the larger batch size the better (specially for VGG-16 according to our experiments). Importantly, when used for regression, the performance of ResNet-50 seems to be quite independent of the batch size.
%DONE_RADU: I would remove this phrase or make it more positive. It may go in some general conclusions concerning how one should report the results.
% In addition, we strongly encourage future authors to provide the batch size they employ in their papers to facilitate reproducibility.

%\subsection{Discussion on Optimization}

%DONE_RADU: This is too vague. What means "hardware capacity". There is no quantitative information here...
%DONE_RADU: I would avoid phrases like "research group with limited budget" , please turn this differently and say what is the minimal "hardware capacity. Also the ethical concern should be removed on my opinion.
% research group with a limited budget may just be unable to (i) investigate optimization settings leading to network convergence and/or (ii) produce results that lead to a scientific publication. We believe this raises a non-negligible ethical concern for the computer vision community.

\section{Statistical Analysis of the Results}
\label{sec:stat}

%{\color{red}
%\begin{enumerate}
%\item Identify sources of randomness: weights initialization for the last regression layer, optimization (data are shuffled), or data augmentation.  
%\item Justify why optimizers and the batch size are evaluated separately. We are mainly testing architectural choices, and not optimization choices.  
%\item Justify the selection of the baseline (standard configuration that works sufficiently well).
%\item Justify that we don't use specific metrics when selecting the model to use in the second block (evaluation).
%\item Justify that we use only 5 runs per configuration. 
%\item Mention that we use an all-vs-all comparison: since we don't have that many comparisons we do all tests and decide which method is the best (if one method/configuration is better than all others we put the asterisks).
%\item We are treating data, for the statistical tests, like if they were independent, but they are not (they are several repetitions on the same data). Is this a serious problem? Check the assumptions over which Wilcoxon is built to see if we're making some serious mistake...
%\item IMPORTANT NOTE: somewhere we should plainly say that ``We decide to report the median because the distribution of mean absolute and squared errors, as well as the paired differences, per image are not normally distributed, and the mean was affected by outliers. "
%\end{enumerate}
%}

Deep learning methods are generally based on stochastic optimization techniques, in which different sources of randomness, i.e.\ weight initialization, optimization and regularization procedures, have an impact on the results. In the analysis of optimization techniques and of batch sizes presented in the previous section we already observed some stochastic effects. While these effects did not forbid us to make reasonable optimization choices, other architecture design choices may be in close competition. In order to appropriately referee such competitions, one should draw conclusions based on rigorous statistical tests, rather than based on the average performance of a single training trial.
% draw conclusions from a single execution is not a rigorous scientific strategy. 
In this section we describe the statistical procedures implemented to analyze the results obtained after several training trials. We use two statistical tools widely used in many scientific domains.
% but barely used by the deep learning and computer vision communities. 

Generally speaking, statistical tests measure the probability of obtaining experimental results $D$ if hypothesis $H$ is correct, thus computing $P(D|H)$. The null hypothesis ($H_0$) refers to a general or default statement of a scientific experiment. It is presumed to be true until statistical evidence nullifies it for an alternative hypothesis ($H_1$). $H_0$ assumes that any kind of difference or significance observed in the data is due to chance. In this paper, $H_0$ is that none of the configurations under comparison in a particular experiment is any better, in terms of median performance, than other configurations. The estimated probability of rejecting $H_0$ when it is true is called \textit{p-value}. If the p-value is less than the chosen level of significance $\alpha$ then the null hypothesis is rejected. Therefore, $\alpha$ indicates how extreme observed results must be in order to reject $H_0$. For instance, if the p-value is less than the predetermined significance level (usually 0.05, 0.01, or 0.001, indicated with one, two, or three asterisks, respectively), then the probability of the observed results under $H_0$ is less than the significance level. In other words, the observed result is highly unlikely to be the result of random chance.  Importantly, the p-value only provides an index of the evidence against the null hypothesis, i.e. it is mainly intended to establish whether further research into a phenomenon could be justified. We consider it as one bit of evidence to either support or challenge accepting the null hypothesis, rather than as conclusive evidence of significance \cite{fisher1925,Nuzzo2014,Vidgen2016}, and a statistically insignificant outcome should be interpreted as ``absence of evidence, not evidence of absence" \cite{Sterne2001}. 

Statistical tests can be categorized into two classes: parametric and non-parametric. Parametric tests are  based on assumptions  (like normality or homoscedasticity) that are commonly violated when analyzing the performance of stochastic algorithms~\cite{DerracGMH11}. In our case, the visual inspection of the error measurements as well as the application of normality tests (in particular, the Lilliefors test) indicates a lack of normality in the data, leading to the use of non-parametric statistical tests. 

%Even if, in practice, the violation of the normality assumption  with large sample sizes does not cause major problems, 

% showed that these are not always uni-modal and bell-shaped. %, even commonly being very skewed. 
% In fact, the visualization through Quantile-Quantile plots displayed that data sometimes show a serious deviation from normality. Measures of central tendency (mean, median) and dispersion (std) also indicated that the distributions are commonly far from normality. Statistical tests employing the paired t-test and the Wilcoxon signed-rank test were performed leading to different conclusions, which in our opinion suggests a lack of normality in the data that produces inconsistent results when using a parametric test. As conclusion, due to the lack of normality, and for the sake of simplicity, the statistical analysis employed only uses non-parametric statistics.  \xavi{Rephrase the previous paragraph. It looks like the distribution of the error is not unimodal, then why using the median?}

Statistical tests can perform
two kinds of analysis: pairwise comparisons and multiple comparisons.
Pairwise statistical procedures perform comparisons
between two algorithms, obtaining in each application a
p-value independent from another one. Therefore, in order to carry
out a comparison which involves more than two algorithms, multiple
comparison tests should be used. If we try to draw a conclusion involving
more than one pairwise comparison, we will obtain an accumulated
error. In statistical terms, we are
losing control on the Family-Wise Error Rate (FWER), defined
as the probability of making one or more false discoveries (type I errors) among
all the hypotheses when performing multiple pairwise tests. Examples of post-hoc procedures, used to control the FWER, are Bonferroni-Dunn \cite{Dunn1961}, Holm \cite{Holm79}, Hochberg \cite{Hochberg1988}, Hommel \cite{Hommel1988}, Holland \cite{Holland1987}, Rom \cite{Rom1990}, or Nemenyi \cite{Nemenyi63}. Following the recommendation of Derrac et al. \cite{DerracGMH11}, in this paper we use the Holm procedure to control the FWER.%, can always be considered better than the Bonferroni-Dunn procedure, because it appropriately controls the FWER and it is more powerful than the Bonferroni-Dunn procedure. In relation to the post-hoc procedures, the differences of power between many methods are rather small, with some exceptions: Bonferroni-Dunn test should not be used in spite of its simplicity, because it is a very conservative test and many differences may not be detected, but procedures like Holm, Hochberg, Hommel, Holland, or Rom have a similar power. 

%According to \cite{DerracGMH11}, the Holm procedure, the one adopted in this paper, can always be considered better than the Bonferroni-Dunn procedure, because it appropriately controls the
%FWER and it is more powerful than the Bonferroni-Dunn procedure. In relation to the post-hoc procedures, the differences
%of power between many methods are rather small, with some
%exceptions: Bonferroni-Dunn test should not be used in
%spite of its simplicity, because it is a very conservative test and
%many differences may not be detected, but procedures like Holm, Hochberg, Hommel, Holland, or Rom have a similar
%power. 

Summarizing, once established that non-parametric statistics should be used, we decided to follow standard and well-consolidated statistical approaches: when pairwise comparisons have to be made, the Wilcoxon signed-rank test \cite{wilcoxon:test} is applied; when multiple comparisons have to be made (i.e. more than two methods are compared, thus increasing the number of pairwise comparisons), the FWER is controlled by applying the Bonferroni-Holm procedure (also called the Holm method) to multiple Wilcoxon signed-rank tests.  Finally, the $95\%$ confidence interval for the median of the MAE is reported.%obtained by each configuration is shown. 

\subsection{Wilcoxon Signed-rank Test}
The Wilcoxon signed-rank test \cite{wilcoxon:test} is a non-parametric statistical hypothesis test used to compare two related samples to assess the null hypothesis that the median difference between pairs of observations is zero. It can be used as an alternative to the paired Student's t-test, t-test for matched pairs,\footnote{Two data samples are matched/paired if they come from repeated observations of the same subject.} when the population cannot be assumed to be normally distributed. We use Wilcoxon signed-rank test to evaluate which method is the best (i.e. the most recommendable configuration according with our results) and the worst (i.e. the less recommendable configuration according with our results). The statistical significance is displayed on each table using asterisks, as commonly employed in the scientific literature: * represents a p-value smaller than 0.05 but larger or equal than 0.01, ** represents a p-value smaller than 0.01 but larger or equal than 0.001, and *** represents a p-value smaller than 0.001. 
%RADU: I don't understand the next phrase
When more than one configuration has asterisks it implies that these configurations are significantly better than the others but there are no statistically significant differences between them. 
The worst performing configurations are displayed using circles and following the same criterion.

\subsection{Confidence Intervals for the Median}
Importantly, with a sufficiently large sample, statistical significance tests may detect a trivial effect, or they may fail to detect a meaningful or obvious effect due to small sample size. In other words, very small differences, even if statistically significant, can be practically meaningless. Therefore, since we consider that reporting only the significant p-value for an analysis is not enough to fully understand the results, we decided to introduce confidence intervals as a mean to quantify the magnitude of each parameter of interest. 

Confidence intervals consist of a range of values (interval) that act as good estimates of the unknown population parameter.  Most commonly, the 95\% confidence interval is used. A confidence interval of 95\%  does not mean that for a given realized interval there is a 95\% probability that the population parameter lies within it (i.e.\ a 95\% probability that the interval covers the population parameter), but that there is a 95\% probability that the calculated confidence interval from some future experiment encompasses the true value of the population parameter. The 95\% probability relates to the reliability of the estimation procedure, not to a specific calculated interval. If the true value of the parameter lies outside the 95\% confidence interval, then a sampling event that has occurred with a probability of 5\% (or less) of happening by chance. 

We can estimate confidence intervals for medians and other quantiles  using the binomial distribution. The 95\% confidence interval for the $q$-th quantile can be found by applying the binomial distribution \cite{conover98}. The number of observations less than the $q$ quantile will be an observation from a binomial distribution with parameters $n$ and $q$, and hence has mean $nq$ and standard deviation $\sqrt{(nq(1-q)}$. We calculate $j$ and $k$ such that:
$j = nq - 1.96 \sqrt{nq(1-q)}$ and 
$k = nq + 1.96 \sqrt{nq(1-q)}$.
We round $j$ and $k$ up to the next integer. Then the 95\% confidence interval is between the $j^{th}$ and $k^{th}$ observations in the ordered data. 

\section{Network Variants}
\label{sec:variant}
%\section{Network Variants}
%\label{sec:variant}
%
\begin{table}[t]
\centering
\caption{Network baseline specification.%\vspace{-2mm}
}
 \resizebox{\linewidth}{!}{\begin{tabular}{lcccccc}
   \toprule
   Network &$\mathcal{L}$& BN & FT & DO & LR & TIR\\
   \midrule
   \rowcolor{mygray}
   VGG-16 & $\model MSE$  & $\model BN$ & $\model CB^4$ & $\model 10{-}\model DO$ & $\model \rho(FC^2)$ & $\model FC^2$\\
%    \midrule
   \rowcolor{mygray}
   ResNet-50& $\model MSE$  & $\cancel{\model BN}$ & $\model CB^3$ &  - & $\model \rho(GAP)$ & $\model GAP$\\
  \bottomrule\vspace{-6mm}
 \end{tabular}}
 \label{tab:baselines}
\end{table}

The statistical tests described above are used to compare the performance of each choice on the three data sets for the two base architectures. Due to the amount of time necessary to train deep neural architectures, we cannot compare all possible combinations, and therefore we must evaluate one choice at a time (e.g.\ the use of batch normalization). In order to avoid over-fiting, we use holdout as model validation tecnique, and test the generalization ability with an independent data set. In detail, we use 20\% of training data for validation (26\% in the case of FLD, because the validation set is explicitly provided in~\cite{Sun2013}). We use early stopping with a patience equal to four epochs (an epoch being a complete pass through the entire training set). In other words, the network is trained until the loss on the validation set does not decrease during four consecutive epochs. The two baseline networks are trained with the optimization settings chosen in Section~\ref{sec:optimization}. In this section, we evaluate the performance of different network \textit{variants}, on the three problems. 

Since ConvNets have a high number of parameters, they are prone to over-fitting.  Two common regularization strategies are typically used~\cite{JMLRsrivastava14a}:

\whencolumns{\begin{table}[t]}{\begin{table*}[t]}
\centering
 \caption{Impact of the loss choice ($\mathcal{L}$) on VGG-16 and ResNet-50.\vspace{-2mm}
 }
 \resizebox{\textwidth}{!}{\begin{tabular}{ccllcccllccc}
 \toprule
Data & \multirow{3}{*}{$\mathcal{L}$} && \multicolumn{4}{ c }{VGG-16} && \multicolumn{4}{ c }{ResNet-50}\\
\cmidrule{4-12}
 Set &  &&MAE test&MSE train& MSE valid &MSE test &&MAE test&MSE train& MSE valid &MSE test\\
   \midrule
\rowcolor{\rowcolorrevone}& $\model MSE$ &&\ccg [3.66 3.79]*** &\ccg [4.33 4.41] &\ccg [12.18 12.56] &\ccg [18.77 20.20]  &&\ccg [3.60 3.71]$^{\circ\circ\circ}$ &\ccg [1.25 1.27] &\ccg [21.49 22.25] &\ccg [17.15 18.14] \\ 
\rowcolor{\rowcolorrevone}& $\model HUB$ && [4.08 4.23]$^{\circ\circ\circ}$ & [5.41 5.53] & [12.30 12.65] & [23.96 25.83]  && [3.46 3.55] & [0.80 0.82] & [19.02 19.64] & [16.43 17.28] \\ 
\rowcolor{\rowcolorrevone}\multirow{-3}{*}{Biwi}& $\model MAE$&& [4.00 4.11]*** & [5.56 5.68] & [12.67 13.08] & [22.84 24.16] && [3.27 3.37]* & [0.79 0.80] & [16.34 16.90] & [15.33 16.15] \\ 
   \midrule
\rowcolor{\rowcolorrevone}& $\model MSE$ &&\ccg [2.61 2.76]$^{\circ\circ\circ}$ &\ccg [9.43 9.54] &\ccg [10.77 11.03] &\ccg [10.55 11.62]  &&\ccg [1.96 2.05]$^{\circ\circ\circ}$ &\ccg [5.21 5.25] &\ccg [6.85 6.95] &\ccg [5.79 6.39] \\ 
\rowcolor{\rowcolorrevone}& $\model HUB$ && [2.32 2.43]*** & [7.67 7.76] & [8.98 9.15] & [8.24 9.03]  && [1.75 1.84]*** & [3.12 3.18] & [5.36 5.47] & [4.62 5.09] \\ 
\rowcolor{\rowcolorrevone}\multirow{-3}{*}{FLD}& $\model MAE$&& [2.36 2.51] & [8.24 8.34] & [9.54 9.71] & [8.81 9.65] && [1.74 1.82]*** & [2.99 3.04] & [5.04 5.13] & [4.65 5.10] \\ 
   \midrule
\rowcolor{\rowcolorrevone}& $\model MSE$ &&\ccg [4.90 5.59] &\ccg [2.38 2.41] &\ccg [27.74 28.48] &\ccg [41.35 52.13]  &&\ccg [4.86 5.68] &\ccg [0.64 0.64] &\ccg [29.40 30.21] &\ccg [43.58 55.71] \\
\rowcolor{\rowcolorrevone}& $\model HUB$ && [4.89 5.34] & [2.69 2.72] & [29.64 30.46] & [40.68 49.57]  && [4.83 5.66] & [0.38 0.38] & [27.97 28.79] & [42.90 54.76] \\ 
\rowcolor{\rowcolorrevone}\multirow{-3}{*}{Parse}& $\model MAE$&& [4.74 5.29] & [2.48 2.52] & [26.51 27.35] & [38.87 51.98] && [4.83 5.49] & [0.54 0.54] & [28.18 28.98] & [42.76 54.11] \\ 
   \midrule
\rowcolor{\rowcolorrevone}& $\model MSE$ &&\ccg [8.42 8.78]$^{\circ\circ\circ}$ &\ccg [73.96 74.62] &\ccg [165.3 170.4] &\ccg [130.7 142.3]  &&\ccg [8.71 9.06]$^{\circ\circ\circ}$ &\ccg [114.2 115.5] &\ccg [191.2 196.4] &\ccg [142.1 154.5] \\
\rowcolor{\rowcolorrevone}& $\model HUB$ && [8.26 8.62] & [60.07 60.77] & [157.9 162.8] & [125.6 138.5]  && [7.69 8.03] & [84.07 85.02] & [159.1 164.0] & [115.3 125.8] \\ 
\rowcolor{\rowcolorrevone}\multirow{-3}{*}{MPII}& $\model MAE$&& [7.36 7.69]*** & [47.22 47.73] & [138.9 143.2] & [103.7 113.8] && [6.42 6.71]*** & [43.32 43.84] & [121.9 126.0] & [81.65 91.07] \\ 
   \bottomrule\vspace{-5mm}
 \end{tabular}}
 \label{tab:VGG-resnet-loss}
\whencolumns{\end{table}}{\end{table*}}

\addnote[LN]{1}{{\bf Loss} ($\mathcal{L}$) denotes the loss function employed at training time. The baselines employ  the Mean Squared Error ($\model MSE$). Following \cite{iciploss}, we compare the ($\model MSE$) with the Mean Absolut Error ($\model MAE$) and the Huber loss ($\model HUB$).}% \item 

{\bf Batch Normalization} (BN) was introduced to lead to fast and reliable network convergence~\cite{ioffe2015batch}. In the case of VGG-16 (resp. ResNet-50), we cannot add (remove) a batch normalization layer deeply in the network since the pre-trained weights of the layers after this batch normalization layer were obtained without it. Consequently, in the case of VGG-16 we can add a batch normalization layer either right before $REG$ (hence after the activation of $FC^2$, denoted by $\model BN$), or before the activation of $FC^2$ (denoted by $\model BNB$), or we do not use batch normalization $\cancel{\model BN}$. In ResNet-50 we consider only $\model BN$ and $\cancel{\model BN}$, since the batch normalization layer before the activation of the last convolutional layer is there by default. Intuitively, VGG-16 will benefit from the configuration $\model BN$, but not ResNet-50. This is due to the fact that the original VGG-16 does not exploit batch normalization, while ResNet-50 does. Using $\model BN$ in ResNet-50 would mean finishing by convolutional layer, batch normalization, activation, GAP, batch normalization and REG. A priori we do no expect gains when using ResNet-50 with $\model BN$ (and this is why the ResNet-50 baselines do not use BN), but we include this comparison for completeness.
\addnote[LN]{1}{Finally, we compare the use of batch normalization with the more recent layer normalization\cite{ba2016layer} denoted by $\model LN$.}% \item 

{\bf Dropout} (DO) is a widely used method to avoid over-fitting~\cite{JMLRsrivastava14a}. Dropout is not employed in ResNet-50, and thus we perform experiments only on VGG-16. We compare different settings: no dropout (denoted by $\model 00{-}DO$), dropout in $FC^1$ but not in $FC^2$ ($\model 10{-}DO$), dropout in $FC^2$ but not in $FC^1$ ($\model 01{-}DO$) and dropout in both ($\model 11{-}DO$).% {\color{red} PABLO: should we say here what is the amount of dropout used by default? Because we use 0.5, right? But we don't say it anywhere. }
% These settings are referred to as $\model 0{-}DO$, $\model 1{-}DO$ and $\model 2{-}DO$ respectively. \xavi{Add dropout in second but not in first}
% \end{enumerate}

Other approaches consist in choosing a network architecture that is less prone to over-fitting, for instance by chaging the number of parameters that are learned (fine-tuned). We compare three different strategies:
%RADU_NEW: I think that we need to come back to the enumeration environment.
%RADU_NEW: still, it is unclear what is to be modified !
%  \begin{enumerate}
% \setcounter{enumi}{2}
%  \item 

{\bf Fine-tuning depth.} FT %. When a pre-trained model is used, only the last layers are modified. Deeper layers are considered frozen and consequently they are not updated. The underlying question is: how many layers should be fine-tuned?  Each model is denoted by the name of 
denotes the deepest block fixed during training, i.e. only the last layers are modified. For both architectures, we compare $\model CB^2$, $\model CB^3$, $\model CB^4$, $\model CB^5$. Note that the regression layer is always trained from scratch.
  
%  \item 
{\bf Regressed layer.} RL denotes the layer after which the $BN$ and $REG$ layers are added: that is the layer that is regressed. 
% . In previous variants of VGG-16, the regression layer is added after $FC^2$, but this is an arbitrary choice. Other alternatives could consist on removing a few layers before adding $BN$ and $REG$. 
$\rho(RL)$ denotes the model where the regression is performed on the output activations of the layer $RL$, meaning that on top of that layer we directly add batch normalization and linear regression layers. For VGG-16, we compare $\model \rho(CB^5)$, $\model \rho(FC^1)$ and $\model \rho(FC^2)$. For ResNet-50 we compare $\model \rho(CB^5)$ and $\model \rho(GAP)$.

%  \item 
\addnote[tir-desc]{1}{Target \& input representation (TIR). The target representation can be either a heatmap or a low-dimensional vector. While in the first case, the input representation needs to keep the spatial correspondence with the input image, in the second case one can choose an input representation that does not maintain a spatial structure. In our study, we evaluate the performance obtained when using heatmap regression ($\model HM$) and when the target is a low-dimensional vector and the input is either the output of a global average pooling ($\model GAP$), global max pooling ($\model GMP$) or, in the case of VGG-16, the second fully connected layer ($\model FC^2$). In the case of $\model HM$, we use architectures directly inspired by \cite{Belagiannis2017recurrent,cao2016realtime}. In particular, we add a $1\times1$ convolution layer with batch normalization and ReLU activations, followed by a $1\times1$ convolution with linear activations to $CB^3$ and $CB^2$ for VGG-16 and ResNet-50, respectively, predicting a heatmap of dimensions $56\times56$ and $55\times55$, respectively. Following common practice in heatmap regression, the ground-truth heatmaps are generated according to a Gaussian-like function \cite{Belagiannis2017recurrent,newell2016stacked}, and training is performed employing the MSE loss. Importantly, for the experiments with heatmap regression, the FT configuration needs to be specifically adapted. Indeed, the baseline fine-tuning would lead to no fine-tuning, since $CB^3$ and $CB^4$ are not used. Consequently, we employ the $\model CB^2$ fine-tuning configuration for both VGG-16 and ResNet-50 networks. At test time, the landmark locations are obtained as the arguments of the maxima over the predicted heatmaps.}
% 
% Often, pooling is used to convert convolutional maps into vectors. In the case of VGG-16, the 2D feature maps of $CB^5$ are converted via a flattening layer $Fl$ that does not reduce the dimension. Alternatively, we could replace $Fl$ by global (max or average) pooling, denoted by $GMP$ and $GAP$ respectively. ResNet-50 already uses $GAP$ and hence we can only compare to $GMP$. The models are denoted $\model GAP$, $\model GMP$, for both, and $\model FC^2$ for VGG-16.
%  \end{enumerate} 

The settings corresponding to our baselines are detailed in Table~\ref{tab:baselines}. These baselines are chosen regarding common choices in the state-of-the-art (for ${\cal L}$, BN and DO) and in the light of preliminary experiments (not reported, for FR, LR and TIR). No important changes with respect to the original design were adopted. The background color is gray, as it will be for these two configurations in the remainder of the paper. We now discuss the results obtained using the previously discussed variants.

\whencolumns{\begin{table}[t]}{\begin{table*}[t]}
\centering
 \caption{Impact of the batch normalization (BN) layer on VGG-16 and ResNet-50.\vspace{-2mm}
 }
 \resizebox{\textwidth}{!}{\begin{tabular}{ccllcccllccc}
 \toprule
Data & \multirow{3}{*}{BN} && \multicolumn{4}{ c }{VGG-16} && \multicolumn{4}{ c }{ResNet-50}\\
\cmidrule{4-12}
 Set &  &&MAE test&MSE train& MSE valid &MSE test &&MAE test&MSE train& MSE valid &MSE test\\
   \midrule
		      & $\cancel{\model BN}$  && [5.04 5.23]$^{\circ\circ}$ & [2.52 2.56] & [20.68 21.45] & [35.68 37.81] && \ccg [3.60 3.71]*** & \ccg [1.25 1.27] & \ccg [21.49 22.25] &\ccg [17.15 18.14] \\ 
		      & $\model BN$ && \ccg [3.66 3.79]*** & \ccg [4.33 4.41] & \ccg [12.18 12.56] & \ccg [18.77 20.20] && [4.59 4.69]$^{\circ\circ\circ}$ & [2.18 2.22] & [22.93 23.63] & [28.56 30.07] \\
\rowcolor{\rowcolorrevone}& $\model LN$ && [3.93 4.06] & [5.76 5.88] & [14.38 14.84] & [21.21 22.51]  && [3.63 3.73] & [0.99 1.01] & [22.79 23.43] & [18.10 19.09] \\
\multirow{-4}{*}{Biwi}& $\model BNB$&& [4.63 4.76] & [8.11 8.29] & [16.69 17.32] & [30.49 32.57] && \multicolumn{1}{c}{--} & -- & -- & --  \\
   \midrule
		      & $\cancel{\model BN}$ &&  [3.67 3.90] & [21.19 21.45] & [22.26 22.76] & [19.70 22.25] &&\ccg [1.96 2.05]* & \ccg[5.21 5.25] & \ccg[6.85 6.95] & \ccg[5.79 6.39] \\ 
		      & $\model BN$ && \ccg [2.61 2.76] & \ccg[9.43 9.54] & \ccg[10.77 11.03] & \ccg[10.55 11.62] && [1.92 2.01]** & [4.81 4.86] & [6.83 6.94] & [5.70 6.34] \\  
\rowcolor{\rowcolorrevone}& $\model LN$ && [2.19 2.31]*** & [6.38 6.45] & [8.50 8.64] & [7.29 8.10] && [2.00 2.08]$^\circ$ & [4.34 4.39] & [6.60 6.72] & [6.16 6.73] \\  
\multirow{-4}{*}{FLD} & $\model BNB$&&  [15.53 16.65]$^{\circ\circ\circ}$ & [300.3 304.5] & [300.4 307.7] & [326.9 369.4] && \multicolumn{1}{c}{--} & -- & -- & --  \\
   \midrule
		      & $\cancel{\model BN}$  && [6.54 7.17] & [9.50 9.68] & [56.74 57.96] & [69.12 84.83] &&\ccg [4.86 5.68]*** &\ccg [0.64 0.64] &\ccg [29.40 30.21] &\ccg [43.58 55.71] \\ 
		      & $\model BN$ &&\ccg [4.90 5.59] &\ccg [2.38 2.41] &\ccg [27.74 28.48] &\ccg [41.35 52.13] && [5.72 6.25]$^{\circ\circ\circ}$ & [0.89 0.89] & [36.24 37.22] & [55.57 69.71] \\  
\rowcolor{\rowcolorrevone}& $\model LN$ && [4.73 5.31]** & [1.71 1.74] & [27.11 27.78] & [38.72 46.95] && [5.25 6.17] & [1.32 1.33] & [31.88 32.62] & [49.64 62.11] \\ 
\multirow{-4}{*}{Parse}& $\model BNB$ && [11.20 12.68]$^{\circ\circ\circ}$ & [152.2 154.0] & [142.9 146.2] & [184.9 238.1]  && \multicolumn{1}{c}{--} & -- & -- & --  \\
   \midrule
\rowcolor{\rowcolorrevone} & $\cancel{\model BN}$  && [11.43 11.87]$^{\circ\circ\circ}$ & [208.4 210.6] & [284.5 291.8] & [232.0 251.5] &&\ccg [8.71 9.06] &\ccg [114.2 115.5] &\ccg [191.2 196.4] &\ccg [142.1 154.5] \\ 
\rowcolor{\rowcolorrevone} & $\model BN$ &&\ccg [8.42 8.78] &\ccg [73.96 74.62] &\ccg [165.3 170.4] &\ccg [130.7 142.3] && [8.37 8.77]*** & [63.44 64.14] & [175.8 181.2] & [131.2 143.3] \\
\rowcolor{\rowcolorrevone} & $\model LN$ && [7.81 8.18]*** & [38.67 39.04] & [150.2 154.9] & [114.1 123.8] && [9.69 10.10]$^{\circ\circ\circ}$ & [157.3 159.1] & [218.9 224.3] & [172.1 185.6] \\ 
\rowcolor{\rowcolorrevone}\multirow{-4}{*}{MPII} & $\model BNB$ && [8.76 9.15] & [64.69 65.33] & [194.4 200.1] & [141.2 154.8]  && \multicolumn{1}{c}{--} & -- & -- & --  \\
   \bottomrule\vspace{-5mm}
 \end{tabular}}
 \label{tab:VGG-resnet-BN}
\whencolumns{\end{table}}{\end{table*}}

\subsection{Loss}
\label{sec:loss}
\addnote[results-loss]{1}{Table~\ref{tab:VGG-resnet-loss} presents the results obtained when changing the training loss of the two base architectures and the four datasets. The results of Table 2 are quite unconclusive, since the choice of the loss to be used highly depends on the dataset, and varies also with the base architecture. Indeed, while for MPII the best and worst choices are MAE and MSE respectively, MSE could be the best or worst for Biwi, depending on the architecture, and the Huber loss would be the optimal choice for FLD. All three loss functions are statistically equivalent for Parse, while for the other datasets significant differences are found. Even if the standard choice in the state-of-the-art is the $L_2$ loss, we strongly recommend to try all three losses when exploiring a new dataset.}

\subsection{Batch Normalization}
\label{sec:res-bn}
%Table \ref{tab:VGG-resnet-BN} shows the results obtained with the various choices of batch normalization. In the case of VGG-16, we observe that the impact of adding a batch normalization layer after the activations (i.e.\ $\model BN$) is significant and beneficial compared to the other two options. In the case of FLD, we notice that the problem with $\cancel{\model BN}$ and $\model BNB$ occurs at training since the final training MSE score is much higher than the one obtained with $\model BN$. Interestingly, on the Biwi data set we observe that the training MSE is better with $\cancel{\model BN}$ but $\model BN$ performs better on the validation and the test sets. We conclude that in the case of Biwi, $\model BN$ does not help the optimization, but increases the generalization ability. The high beneficial impact observed in VGG-16 justifies that we use it in the baseline network. The exact opposite trend is observed when running the same experiments with ResNet-50. Indeed, $\model BN$  neither improves the optimization (with the exception of FLD, which appears to be quite insensitive to BN using ResNet-50), nor the generalization ability. As discussed, this is expected because ResNet-50 uses already batch normalization (before the activation).

Table \ref{tab:VGG-resnet-BN} shows the results obtained \addnote[BNLN]{1}{ employing several normalization strategies. In the case of VGG-16, we observe that the impact of employing layer normalization (i.e.\ $\model LN$) or adding a batch normalization layer after the activations (i.e.\ $\model BN$), specially the former, is significant and beneficial compared to the other two alternatives. In the case of FLD, we notice that the problem with $\cancel{\model BN}$ and $\model BNB$ occurs at training since the final training MSE score is much higher than the one obtained with $\model BN$. Interestingly, on the Biwi data set we observe that the training MSE is better with $\cancel{\model BN}$ but $\model BN$ performs better on the validation and the test sets. We conclude that in the case of Biwi, $\model BN$ does not help the optimization, but increases the generalization ability. The highly beneficial impact of normalization observed in VGG-16 clearly justifies its use. In particular, $\model LN$ has empirically shown a better performance than $\model BN$, so its use is strongly recommended. However, this paper considers $\model BN$ as a baseline, both for the positive results it provides and for being a more popular and consolidated normalization technique within the scientific community. }The exact opposite trend is observed when running the same experiments with ResNet-50. Indeed, $\model BN$  neither improves the optimization (with the exception of FLD, which appears to be quite insensitive to BN using ResNet-50), nor the generalization ability. As discussed, this is expected because ResNet-50 uses already batch normalization (before the activation). 

\begin{table}
\centering
 \caption{Impact of the dropout (DO) layer on VGG-16.\vspace{-2mm}
 }
\whencolumns{%
 \begin{tabular}{ccllccc}
\toprule
 Dataset & DO &&MAE test&MSE train& MSE valid &MSE test\\
   \midrule
   \multirow{4}{*}{Biwi}    & $\model 00{-}DO$&&  [4.47 4.60]$^{\circ\circ\circ}$ & [6.34 6.45] & [14.42 14.91] & [28.80 30.98]  \\
   & $\model 01{-}DO$ && [3.56 3.67] & [3.35 3.40] & [12.00 12.40] & [17.52 18.54] \\ 
   & $\model 10{-}DO$ && \ccg[3.66 3.79] &\ccg [4.33 4.41] & \ccg[12.18 12.56] & \ccg[18.77 20.20]  \\    
   & $\model 11{-}DO$&&  [3.37 3.48]*** & [3.46 3.53] & [11.66 12.04] & [15.39 16.38] \\  
   \midrule
   \multirow{4}{*}{FLD}  & $\model 00{-}DO$ &&[2.55 2.70] & [8.60 8.71] & [10.53 10.74] & [10.16 11.23] \\
   & $\model 01{-}DO$ && [2.26 2.38]*** & [7.81 7.90] & [9.19 9.38] & [7.87 8.68] \\ 
   & $\model 10{-}DO$ &&\ccg[2.61 2.76] & \ccg[9.43 9.54] & \ccg[10.77 11.03] & \ccg[10.55 11.62]     \\ 
   & $\model 11{-}DO$ &&[2.27 2.42]*** & [7.59 7.68] & [9.10 9.29] & [7.94 8.92]\\ 
   \midrule
   \multirow{4}{*}{Parse}   & $\model 00{-}DO$ && [4.83 5.44] & [1.39 1.41] & [27.28 28.09] & [40.38 51.11]\\
   & $\model 01{-}DO$ && [4.87 5.52] & [2.87 2.90] & [27.89 28.64] & [42.92 50.80] \\ 
   & $\model 10{-}DO$ && \ccg[4.90 5.59] & \ccg[2.38 2.41] & \ccg[27.74 28.48] & \ccg[41.35 52.13] \\ 
   & $\model 11{-}DO$ &&  [4.91 5.59] & [3.25 3.28] & [29.50 30.21] & [43.46 54.62] \\ 
   \bottomrule
    \end{tabular}
   }{%
 \resizebox{\linewidth}{!}{
 \begin{tabular}{cclccc}
\toprule
 Dataset & DO &MAE test&MSE train& MSE valid &MSE test\\
   \midrule
   \multirow{4}{*}{Biwi}    & $\model 00$&  [4.47 4.60]$^{\circ\circ\circ}$ & [6.34 6.45] & [14.42 14.91] & [28.80 30.98]  \\
   & $\model 01$ & [3.56 3.67] & [3.35 3.40] & [12.00 12.40] & [17.52 18.54] \\ 
   & $\model 10$ & \ccg[3.66 3.79] &\ccg [4.33 4.41] & \ccg[12.18 12.56] & \ccg[18.77 20.20]  \\    
   & $\model 11$ &  [3.37 3.48]*** & [3.46 3.53] & [11.66 12.04] & [15.39 16.38] \\  
   \midrule
   \multirow{4}{*}{FLD}  & $\model 00$ &[2.55 2.70] & [8.60 8.71] & [10.53 10.74] & [10.16 11.23] \\
   & $\model 01$ & [2.26 2.38]*** & [7.81 7.90] & [9.19 9.38] & [7.87 8.68] \\ 
   & $\model 10$ &\ccg[2.61 2.76] & \ccg[9.43 9.54] & \ccg[10.77 11.03] & \ccg[10.55 11.62]     \\ 
   & $\model 11$ &[2.27 2.42]*** & [7.59 7.68] & [9.10 9.29] & [7.94 8.92]\\ 
   \midrule
   \multirow{4}{*}{Parse}   & $\model 00$ & [4.83 5.44] & [1.39 1.41] & [27.28 28.09] & [40.38 51.11]\\
   & $\model 01$ & [4.87 5.52] & [2.87 2.90] & [27.89 28.64] & [42.92 50.80] \\ 
   & $\model 10$ & \ccg[4.90 5.59] & \ccg[2.38 2.41] & \ccg[27.74 28.48] & \ccg[41.35 52.13] \\ 
   & $\model 11$ &  [4.91 5.59] & [3.25 3.28] & [29.50 30.21] & [43.46 54.62] \\ 
   \midrule
\rowcolor{\rowcolorrevone}& $\model 00$ & [8.33 8.72] & [59.29 59.98] & [162.5 166.7] & [128.8 141.0] \\ 
\rowcolor{\rowcolorrevone}& $\model 01$ & [8.46 8.82] & [68.03 68.65] & [163.2 167.6] & [132.1 143.7] \\ 
\rowcolor{\rowcolorrevone}& $\model 10$ & \ccg[8.42 8.78] & \ccg[73.96 74.62] & \ccg[165.3 170.4] & \ccg[130.7 142.3] \\
\rowcolor{\rowcolorrevone}\multirow{-4}{*}{MPII}& $\model 11$ & [8.42 8.77] & [64.37 65.07] & [167.1 172.3] & [129.9 142.8] \\ 
   \bottomrule\vspace{-7mm}
    \end{tabular}
  }
  }
    \label{tab:VGG-dropout}
\end{table}

\subsection{Dropout Ratio}
\label{sec:res-do}
Table~\ref{tab:VGG-dropout} shows the results obtained when comparing different dropout strategies with VGG-16 (ResNet-50 does not have fully connected layers and thus 
%RADU: I don't understand "it is encouraged to use dropout in both fully connected layers". Do you means "both first and second fully connected layers"?
%XAVI: This is standard wording for deep architectures, I believe.
no dropout is used). In the light of these results, it is encouraged to use dropout in both fully connected layers. However, for Parse \comRev{ and MPII}, this does not seem to have any impact, and 
%DONE_RADU: "seems to be optional" is too vague
on FLD the use of dropout in the first fully connected layer seems to have very mild impact, at least in terms of MAE performance on the test set. Since on the Biwi data set the $\model 
%DONE_RADU: "significantly worse" than what?
00{-}DO$ strategy is significantly worse than the other strategies, and not especially competitive in the other two data sets, we would suggest not to use this strategy. Globally, $\model 11{-}DO$ is the safest option (best for Biwi/FLD, equivalent for Parse).

\begin{figure}
 \centering
 \includegraphics[width=\linewidth]{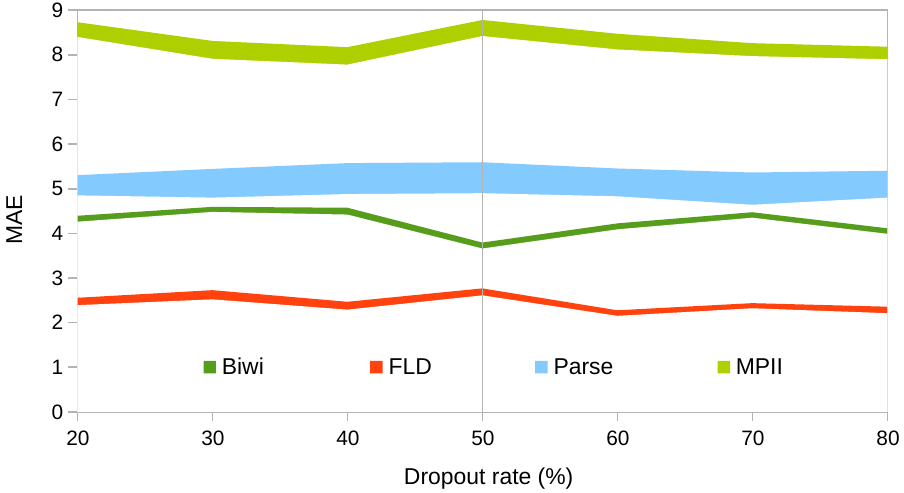}
 \caption{\comRev{Confidence intervals for the median MAE as a function of the dropout rate (\%) for the four datasets (using VGG-16). The vertical gray line is the default configuration.}}
 \label{fig:dropout-rate}
\end{figure}

\addnote[dropout-rate]{1}{We have also analyzed the impact of the dropout rate, from 20\% to 80\% with 10\% steps. We report the confidence intervals for the median of the MAE in the form of a graph, as a function of dropout rate, e.g. Figure~\ref{fig:dropout-rate}. We observe that varying the dropout rate has very little effect on Parse, while some small differences can be observed with the other datasets. While the standard configuration appears to be optimal for Biwi, it is not the optimal one for FLD and MPII. However, the optimal choice for MPII and FLD (40\% and 60\% respectively) are a very bad choice for Biwi. We conclude that the dropout rate should be in the range 40\% -- 60\%, and that rate values outside this range don't provide a significant advantage, at least in our settings.}

\whencolumns{\begin{table}[h!]}{\begin{table*}[t]}
\centering
 \caption{Impact of the finetunig depth (FT) on VGG-16 and ResNet-50.\vspace{-2mm}}
 \resizebox{\linewidth}{!}{\begin{tabular}{ccllcccllccc}
\toprule
 Data & \multirow{2}{*}{FT} & & \multicolumn{4}{c}{VGG-16} && \multicolumn{4}{c}{ResNet-50}\\
\cmidrule{3-12}
 Set & &&MAE test&MSE train& MSE valid &MSE test &&MAE test&MSE train& MSE valid &MSE test\\
    \midrule
   \multirow{4}{*}{Biwi} & $\model CB^5$ && [5.13 5.27] & [4.12 4.20] & [29.18 30.44] & [37.10 39.03] && [8.69 8.90]$^{\circ\circ\circ}$ & [32.57 33.17] & [140 145.1] & [102.5 107.8] \\ 
   & $\model CB^4$ && \ccg [3.66 3.79]*** & \ccg [4.33 4.41] & \ccg [12.18 12.56] & \ccg [18.77 20.20] && [3.40 3.51]*** & [0.87 0.89] & [17.85 18.43] & [15.98 16.86] \\  
   & $\model CB^3$&&  [4.88 5.03] & [4.32 4.40] & [15.96 16.54] & [30.95 32.71] && \ccg [3.60 3.71] & \ccg [1.25 1.27] & \ccg [21.49 22.25] & \ccg [17.15 18.14] \\ 
   & $\model CB^2$ &&  [5.33 5.46]$^{\circ\circ\circ}$ & [9.30 9.51] & [33.48 34.52] & [38.66 40.34] && [4.17 4.30] & [1.41 1.43] & [26.31 27.18] & [24.19 25.34]\\ 
   \midrule
   \multirow{4}{*}{FLD}  & $\model CB^5$ && [3.32 3.47] & [4.00 4.04] & [16.02 16.29] & [16.56 18.25] && [8.79 9.30]$^{\circ\circ\circ}$ & [114.6 116.2] & [120.2 123.1] & [113.1 127.1]\\ 
    & $\model CB^4$ && \ccg [2.61 2.76]*** & \ccg[9.43 9.54] & \ccg[10.77 11.03] & \ccg[10.55 11.62] && [2.21 2.31] & [4.48 4.52] & [8.10 8.24] & [7.36 8.01] \\ 
    & $\model CB^3$ &&  [2.85 3.10] & [6.31 6.40] & [7.79 7.97] & [12.52 14.45]  && \ccg [3.60 3.71] & \ccg [1.25 1.27] & \ccg [21.49 22.25] & \ccg [17.15 18.14] \\ 
    & $\model CB^2$ &&  [3.48 3.74]$^{\circ\circ\circ}$ & [10.83 10.98] & [11.80 12.05] & [18.52 21.96] && [4.17 4.30] & [1.41 1.43] & [26.31 27.18] & [24.19 25.34]\\ 
   \midrule
   \multirow{4}{*}{Parse}& $\model CB^5$ &&  [5.50 6.31]$^{\circ}$ & [1.31 1.33] & [37.29 38.33] & [49.61 63.18] && [8.27 9.28]$^{\circ\circ\circ}$ & [54.22 54.99] & [77.32 78.88] & [102.5 132.5] \\ 
    & $\model CB^4$ && \ccg [4.90 5.59]*** &\ccg [2.38 2.41] &\ccg [27.74 28.48] &\ccg [41.35 52.13] && [5.07 5.86] & [0.90 0.91] & [31.74 32.54] & [44.85 56.78] \\ 
    & $\model CB^3$ && [4.92 5.87] & [2.42 2.46] & [29.91 30.61] & [43.04 57.09] && \ccg [4.86 5.68]*** & \ccg[0.64 0.64] & \ccg[29.40 30.21] & \ccg[43.58 55.71] \\ 
    & $\model CB^2$ &&  [5.38 6.13] & [2.90 2.95] & [37.49 38.26] & [49.21 65.75] && [5.02 5.84] & [0.84 0.85] & [30.48 31.29] & [45.65 61.04]\\ 
   \midrule
   \rowcolor{\rowcolorrevone}& $\model CB^5$ && [11.0 11.4]$^{\circ\circ\circ}$ & [56.45 57.06] & [245.2 251.2] & [211.8 232.1]  && [18.2 18.8]$^{\circ\circ\circ}$ & [534.6 539.5] & [540.7 551.6] & [528.6 563.5] \\ 
   \rowcolor{\rowcolorrevone}& $\model CB^4$ && \ccg[8.42 8.78] & \ccg[73.96 74.62] & \ccg[165.3 170.4] & \ccg[130.7 142.3]  && [8.45 8.85]*** & [53.90 54.41] & [184.1 189.5] & [131.3 144.3] \\ 
   \rowcolor{\rowcolorrevone}& $\model CB^3$ && [8.10 8.47]*** & [69.57 70.35] & [160.5 165.0] & [121.2 133.3]  && \ccg [8.71 9.06] & \ccg [114.2 115.5] & \ccg [191.2 196.4] & \ccg [142.1 154.5] \\ 
   \rowcolor{\rowcolorrevone}\multirow{-4}{*}{MPII} & $\model CB^2$ && [9.21 9.62] & [97.02 98.10] & [202.0 207.4] & [156.7 169.3] && [8.50 8.87]*** & [105.5 106.7] & [183.0 188.5] & [137.7 149.2] \\ 
   \bottomrule\vspace{-7mm}
    \end{tabular}}
    \label{tab:VGG-ResNet-FTDepth}
\whencolumns{\end{table}}{\end{table*}}

\subsection{Fine Tuning Depth}
\label{sec:res-ft}
Table~\ref{tab:VGG-ResNet-FTDepth} shows results obtained with various fine-tuning depth values, as described in Section~\ref{sec:variant}, for both VGG-16 and ResNet-50. In the case of VGG-16, we observe a behavior of $\model CB^4$ similar to the one observed for batch normalization. Indeed, $\model CB^4$ may not be the best choice in terms of optimization ($\model CB^5$ exhibits smaller training MSEs) but it is the best one in terms of generalization ability (for MSE validation and test as well as MAE test). \addnote[mpii-anal]{1}{For VGG-16, this result is statistically significant for all data sets but MPII, where $\model CB^3$ shows a statistically significant better performance}. In addition, the results shown in Table~\ref{tab:VGG-ResNet-FTDepth} also discourage to use $\model CB^2$. It is more difficult to conclude in the case of ResNet-50. While for two of the data sets the recommendation is to choose the baseline (i.e.\ fine tune from 
%RADU: please explain "ablation".
%XAVI: Radu, an ablation study is something pretty standard, at least in deep learning, and consists in studying how different components impact the performance.
$CB^3$), ResNet-50 on Biwi selects the model $\model CB^4$ by a solid margin in a statistically significant manner\addnote[mpii-anal2]{1}{, and the same happens with MPII where $\model CB^4$ and $\model CB^2$ represent better choices}. Therefore, we suggest than, when using ResNet-50 for regression, one still runs an ablation study varying the number of layers that are tuned. In this ablation study, the option $\model CB^5$ should not necessarily be included, since for all data sets this option is significantly worse than the other ones.

%DONE_RADU: The subsection name doesn't sound very English to me.
%DONE_RADU: we should find an alternative to "layer to regress from"
\subsection{Regression Layer}
\label{sec:res-lr}
Tables~\ref{tab:VGG-RegFrom} and~\ref{tab:resnet-RegFrom} show the results obtained when varying the regression layer, for VGG-16 and ResNet-50 respectively. In the case of VGG-16 we observe a strongly consistent behavior, meaning that the best method in terms of optimization performance is also the method that best generalizes (in terms of MSE validation and test as well as MAE test). Regressing from the second fully connected layer is a good choice when using VGG-16, whereas the other choices may be strongly discouraged depending on the data set. The results on ResNet-50 are a bit less conclusive. The results obtained are all statistically significant but different 
%DONE_RADU: please revise the expression "point to regress from"
depending on the data set. Indeed, experiments on Biwi\comRev{, MPII  } and Parse point to use {\model GAP}, while results on FLD point to use $\model CB^5$. However, we can observe that the confidence intervals of $\model \rho(GAP)$ and $\model \rho(CB^5)$ with ResNet-50 on Biwi are not that different. This means that the difference between the two models is small while consistent over the test set images.

\begin{table}[t]
\centering
 \caption{
 Impact of the regressed layer (RL) for VGG-16.\vspace{-2mm}}
 \whencolumns{%
 \begin{tabular}{cclccc}
\toprule
Data Set & RL &MAE test&MSE train& MSE valid &MSE test\\
   \midrule
   \multirow{3}{*}{Biwi} & $\model \rho(FC^2)$&\ccg [3.66 3.79]*** &\ccg [4.33 4.41] &\ccg [12.18 12.56] &\ccg [18.77 20.20]\\  
			 & $\model \rho(FC^1)$& [5.17 5.31]$^{\circ\circ\circ}$ & [9.49 9.66] & [17.84 18.40] & [36.32 38.53] \\ 
			 & $\model \rho(CB^5)$& [4.64 4.75] & [5.38 5.47] & [16.96 17.49] & [28.84 29.85] \\    
   \midrule
   \multirow{3}{*}{FLD}  & $\model \rho(FC^2)$&\ccg [2.61 2.76]*** & \ccg[9.43 9.54] & \ccg[10.77 11.03] & \ccg[10.55 11.62] \\ 
			 & $\model \rho(FC^1)$& [3.51 3.68] & [9.87 10.00] & [12.98 13.23] & [18.22 19.97] \\ 
			 & $\model \rho(CB^5)$& [3.61 3.82]$^{\circ\circ}$ & [16.55 16.74] & [18.49 18.84] & [19.00 21.38]  \\ 
   \midrule
   \multirow{3}{*}{Parse}& $\model \rho(FC^2)$&\ccg [4.90 5.59]** &\ccg [2.38 2.41] &\ccg [27.74 28.48] &\ccg [41.35 52.13] \\ 
			 & $\model \rho(FC^1)$& [4.99 5.66] & [2.61 2.64] & [30.13 30.74] & [43.25 55.67] \\ 
			 & $\model \rho(CB^5)$& [5.58 6.14]$^{\circ\circ\circ}$ & [2.83 2.85] & [34.46 35.12] & [52.62 61.34] \\ 

   \bottomrule
    \end{tabular}
 }{%
 \resizebox{\linewidth}{!}{
 \begin{tabular}{cclccc}
\toprule
Data & RL &MAE test&MSE train& MSE valid &MSE test\\
   \midrule
   \multirow{3}{*}{Biwi} & $\model \rho(FC^2)$&\ccg [3.66 3.79]*** &\ccg [4.33 4.41] &\ccg [12.18 12.56] &\ccg [18.77 20.20]\\  
			 & $\model \rho(FC^1)$& [5.17 5.31]$^{\circ\circ\circ}$ & [9.49 9.66] & [17.84 18.40] & [36.32 38.53] \\ 
			 & $\model \rho(CB^5)$& [4.64 4.75] & [5.38 5.47] & [16.96 17.49] & [28.84 29.85] \\    
   \midrule
   \multirow{3}{*}{FLD}  & $\model \rho(FC^2)$&\ccg [2.61 2.76]*** & \ccg[9.43 9.54] & \ccg[10.77 11.03] & \ccg[10.55 11.62] \\ 
			 & $\model \rho(FC^1)$& [3.51 3.68] & [9.87 10.00] & [12.98 13.23] & [18.22 19.97]  \\ 
			 & $\model \rho(CB^5)$& [3.61 3.82]$^{\circ\circ}$ & [16.55 16.74] & [18.49 18.84] & [19.00 21.38]  \\ 
   \midrule
   \multirow{3}{*}{Parse}& $\model \rho(FC^2)$&\ccg [4.90 5.59]** &\ccg [2.38 2.41] &\ccg [27.74 28.48] &\ccg [41.35 52.13] \\ 
			 & $\model \rho(FC^1)$& [4.99 5.66] & [2.61 2.64] & [30.13 30.74] & [43.25 55.67] \\ 
			 & $\model \rho(CB^5)$& [5.58 6.14]$^{\circ\circ\circ}$ & [2.83 2.85] & [34.46 35.12] & [52.62 61.34] \\ 
   \midrule
\rowcolor{\rowcolorrevone} & $\model \rho(FC^2)$&\ccg [8.42 8.78]*** &\ccg [73.96 74.62] &\ccg [165.3 170.4] &\ccg [130.7 142.3] \\ 
\rowcolor{\rowcolorrevone} & $\model \rho(FC^1)$& [9.75 10.15] & [168.9 170.7] & [208.7 214.0] & [171.5 185.2] \\ 
\rowcolor{\rowcolorrevone}\multirow{-3}{*}{MPII} & $\model \rho(CB^5)$& [10.60 11.08] & [199.5 201.3] & [252.4 257.7] & [200.7 216.4] \\ 
   \bottomrule
    \end{tabular}
    }
    }
    \label{tab:VGG-RegFrom}
\end{table}

\begin{table}[t]
\centering
 \caption{
 Impact of the regressed layer (RL) for ResNet-50.\vspace{-2mm}}
 \whencolumns{%
 \begin{tabular}{cclccc}
\toprule
Data Set & RL &MAE test&MSE train& MSE valid &MSE test\\
   \midrule
   \multirow{2}{*}{Biwi} & $\model \rho(GAP)$&\ccg [3.60 3.71]*** & \ccg [1.25 1.27] & \ccg [21.49 22.25] & \ccg [17.15 18.14] \\  
			 & $\model \rho(CB^5)$& [3.61 3.73] & [0.71 0.72] & [15.42 15.92] & [17.60 18.55]\\    
   \midrule
   \multirow{2}{*}{FLD}  & $\model \rho(GAP)$&\ccg [1.96 2.05] & \ccg[5.21 5.25] & \ccg[6.85 6.95] &\ccg [5.79 6.39]\\  
			 & $\model \rho(CB^5)$& [1.61 1.70]*** & [1.77 1.79] & [4.81 4.90] & [3.98 4.36] \\    
   \midrule
   \multirow{2}{*}{Parse}& $\model \rho(GAP)$&\ccg [4.86 5.68]*** & \ccg[0.64 0.64] & \ccg[29.30 30.09] &\ccg [43.58 55.71] \\  
			 & $\model \rho(CB^5)$&  [5.56 6.24] & [0.48 0.48] & [36.76 37.61] & [54.69 67.21] \\    
   \bottomrule
  \end{tabular}
   }{%
 \resizebox{\linewidth}{!}{
 \begin{tabular}{cclccc}
\toprule
Data & RL &MAE test&MSE train& MSE valid &MSE test\\
   \midrule
   \multirow{2}{*}{Biwi} & $\model \rho(GAP)$&\ccg [3.60 3.71]*** & \ccg [1.25 1.27] & \ccg [21.49 22.25] & \ccg [17.15 18.14] \\  
			 & $\model \rho(CB^5)$& [3.61 3.73] & [0.71 0.72] & [15.42 15.92] & [17.60 18.55]\\

   \midrule
   \multirow{2}{*}{FLD}  & $\model \rho(GAP)$&\ccg [1.96 2.05] & \ccg[5.21 5.25] & \ccg[6.85 6.95] &\ccg [5.79 6.39]\\  
			 & $\model \rho(CB^5)$& [1.61 1.70]*** & [1.77 1.79] & [4.81 4.90] & [3.98 4.36] \\

   \midrule
   \multirow{2}{*}{Parse}& $\model \rho(GAP)$&\ccg [4.86 5.68]*** & \ccg[0.64 0.64] & \ccg[29.30 30.09] &\ccg [43.58 55.71] \\  
			 & $\model \rho(CB^5)$&  [5.56 6.24] & [0.48 0.48] & [36.76 37.61] & [54.69 67.21] \\    
   \midrule
\rowcolor{\rowcolorrevone} & $\model \rho(GAP)$&\ccg [8.71 9.06]*** &\ccg [114.2 115.5] &\ccg [191.2 196.4] &\ccg [142.1 154.5] \\ 
\rowcolor{\rowcolorrevone}\multirow{-2}{*}{MPII} & $\model \rho(CB^5)$& [8.95 9.37] & [125.6 127.0] & [195.1 199.8] & [150.3 163.2] \\ 
   \bottomrule%\vspace{-7mm}
  \end{tabular}
  }
  }
    \label{tab:resnet-RegFrom}
\end{table}

\whencolumns{\begin{table}[h!]}{\begin{table*}[t]}
\centering
 \caption{Impact of the target and input representations (TIR) on VGG-16 and ResNet-50.\vspace{-2mm}}
\resizebox{\textwidth}{!}{
 \begin{tabular}{ccllcccllccc}
\toprule
 Data & \multirow{2}{*}{TIR} & & \multicolumn{4}{c}{VGG-16} && \multicolumn{4}{c}{ResNet-50}\\
\cmidrule{3-12}
 Set & &&MAE test&MSE train& MSE valid &MSE test &&MAE test&MSE train& MSE valid &MSE test\\
    \midrule
   \multirow{3}{*}{Biwi} & $\model GMP$ && [3.97 4.08] & [4.03 4.11] & [18.25 18.99] & [21.19 22.38] &&  [3.64 3.75] & [1.48 1.50] & [20.62 21.23] & [17.80 18.88]\\    
   & $\model GAP$ && [3.99 4.09] & [2.75 2.80] & [20.71 21.40] & [21.03 22.07] &&\ccg[3.60 3.71]*** & \ccg[1.25 1.27] & \ccg[21.49 22.25] &\ccg [17.15 18.14] \\    
   & $\model FC^2$ &&\ccg [3.66 3.79]*** &\ccg [4.33 4.41] &\ccg [12.18 12.56] &\ccg [18.77 20.20] && \multicolumn{1}{c}{--} & -- & -- & --  \\      
   \midrule
   \multirow{4}{*}{FLD}  & $\model GMP$ && [2.50 2.64]*** & [3.02 3.06] & [7.65 7.81] & [9.56 11.07] && [1.75 1.82]*** & [2.17 2.19] & [5.21 5.30] & [4.67 5.22]  \\    
    & $\model GAP$ &&   [2.53 2.67]*** & [5.18 5.23] & [9.06 9.22] & [9.69 10.64]  &&\ccg[1.96 2.05] & \ccg[5.21 5.25] & \ccg[6.85 6.95] & \ccg[5.79 6.39]  \\    
    & $\model FC^2$ &&\ccg [2.61 2.76] & \ccg[9.43 9.54] & \ccg[10.77 11.03] & \ccg[10.55 11.62]  && \multicolumn{1}{c}{--} & -- & -- & --  \\      
   \rowcolor{\rowcolorrevone}& $\model HM$ && [3.08 3.22]$^{\circ\circ\circ}$ & [17.94 18.16] & [18.08 18.46] & [14.74 16.60] && [2.25 2.35]$^{\circ\circ\circ}$ & [8.37 8.46] & [8.99 9.17] & [7.65 8.39] \\ 
   \midrule
   \multirow{4}{*}{Parse}& $\model GMP$ && [5.55 6.06] & [1.17 1.18] & [30.00 30.67] & [52.17 61.86]  &&  [5.74 6.51] & [0.89 0.90] & [36.39 37.17] & [54.63 69.10]  \\    
    & $\model GAP$ && [5.61 6.07] & [1.48 1.50] & [31.46 32.15] & [51.33 60.57]  &&\ccg  [4.86 5.68]*** & \ccg [0.64 0.64] & \ccg [29.30 30.09] & \ccg [43.58 55.71] \\    
    & $\model FC^2$ &&\ccg [4.90 5.59]*** &\ccg [2.38 2.41] &\ccg [27.74 28.48] &\ccg [41.35 52.13]  && \multicolumn{1}{c}{--} & -- & -- & --  \\      
    \rowcolor{\rowcolorrevone}& $\model HM$ && [19.06 21.19]$^{\circ\circ\circ}$ & [455.8 460.3] & [475.4 485.8] & [674.5 806.1] && [18.90 21.32]$^{\circ\circ\circ}$ & [358.6 363.1] & [458.1 469.7] & [631.8 841.5] \\
   \midrule
    \rowcolor{\rowcolorrevone}& $\model GMP$ && [9.55 9.90] & [73.96 74.72] & [215.4 221.0] & [166.5 179.4] && [8.71 9.06] & [114.2 115.5] & [191.2 196.4] & [142.1 154.5] \\ 
    \rowcolor{\rowcolorrevone}& $\model GAP$ && [9.60 9.98] & [63.56 64.21] & [212.0 217.0] & [163.5 178.7] &&\ccg [7.72 8.06]*** &\ccg [23.35 23.56] &\ccg [147.9 152.1] &\ccg [110.9 119.0] \\
    \rowcolor{\rowcolorrevone}& $\model FC^2$ &&\ccg [8.42 8.78]*** &\ccg [73.96 74.62] &\ccg [165.3 170.4] &\ccg [130.7 142.3] && \multicolumn{1}{c}{--} & -- & -- & --  \\      
    \rowcolor{\rowcolorrevone}\multirow{-4}{*}{MPII}& $\model HM$ && [27.50 28.28]$^{\circ\circ\circ}$ & [1489 1500] & [1419 1443] & [1384 1448] && [17.28 18.28]$^{\circ\circ\circ}$ & [885.0 896.4] & [993.1 1016] & [773.22 859.0] \\
   \bottomrule\vspace{-7mm}
    \end{tabular}
    }
    \label{tab:TIR}
\whencolumns{\end{table}}{\end{table*}}

\subsection{Target and Input Representations}
\label{sec:res-pl}
\addnote[results-tir]{1}{Table~\ref{tab:TIR} shows the results obtained for different target and input representations on VGG-16 and ResNet-50. Regarding VGG-16, we observed that except for FLD, the best 
%DONE_RADU: "is not to replace the flatten layer ($Fl$) by a pooling layer (neither $GMP$ nor $GAP$)" maybe you want to say:
% "the best option is to use the flatten layer instead of a pooling layer". Use of "neither ... nor" ambiguous.
option is to keep the flatten layer, and therefore to discard the use of pooling layers ($\model GMP$ or $\model GAP$). The case of FLD is different, since the results would suggest that it is significantly better to use either $\model GAP$ or $\model GMP$. Similar conclusions can be drawn from the results obtained with ResNet-50 in the sense that the standard configuration ($\model GAP$) is the optimal one for Biwi and Parse, but not for FLD. 
%DONE_RADU: strange phrase...
In the case of FLD, the results suggest to choose $\model GMP$. As indicated above, since heatmap regression represents, in general, the state of the art in human-pose estimation and landmark localization, we also tested HM on FLD, Parse and MPII. The first conclusion that can be drawn from Table~\ref{tab:TIR} is that HM produces worse results than the other TIR. A more detailed inspection of Table~\ref{tab:TIR} reveals an overall optimization problem: HM leads to less good local minima. Importantly, this occurs under a wide variety of optimization settings (tested but not reported). We also realized that, current deep regression strategies exploiting heatmaps do not exploit vanilla regressors, but ad-hoc, complex and cascaded structures~\cite{Belagiannis2017recurrent,Nie_2018_CVPR,tompson2014joint,Chen2017AdversarialPA,Chu2017multi}. We also evaluated an encoder-decoder strategy inspired by~\cite{newell2016stacked} with no success. We conclude that the combination of vanilla deep regressors and heatmaps is not well suited. The immediate consequence is twofold: first, the evaluation of state-of-the-art architectures exploiting heatmaps falls outside the scope of the paper, since these architectures are not vanilla deep regressors, and second, applications necessitating computationally efficient  regression algorithms discourage the use of complex cascaded networks, e.g. heatmaps.}
% 
% The results in FLD are reasonable, showing that our network is able to converge and solve the optimization problem. In the case of Parse and MPII, we hypothesize that the data processing used does not effectively contribute to the learning procedure and, even after testing with different learning rates and optimizers, the results are far from those obtained by the other alternatives. We cannot rule out that, to take advantage of HM, it is necessary to resort to more sophisticated and ad-hoc strategies (see \cite{Belagiannis2017recurrent,Nie_2018_CVPR,tompson2014joint,Chen2017AdversarialPA,Chu2017multi}), although an encoder-decoder strategy based on one of these approaches \cite{newell2016stacked} was also evaluated without producing results significantly different from those shown in the table.
% }

\subsection{Discussion on Network Variants}
This section summarized the results obtained with different network variants. Firstly, \addnote[BNandDO]{1}{the behavior of batch normalization is very stable for both networks, i.e. regardless of the problem being addressed, VGG-16 always benefits from the presence of batch normalization, while ResNet-50 always benefits from its absence. Therefore, the recommendation is that evaluating the use of batch normalization is mandatory, but it appears that conclusions are constant over different datasets for a given network, hence there is no need to run extensive experimentation to select the best option. With respect to other normalization techniques, it is worth mentioning that layer normalization statistically outperforms batch normalization in three out of four experiments. Therefore, its use is also highly recommended. In relation to dropout, as it is classically observed in the literature, one should use dropout with VGG-16 (this experiment does not apply to ResNet-50), since its use is always either beneficial or not harmful, and
preferably with $\model FC^2$ and $\model FC^1$. }  Second, regarding the number of layers to be finely tuned, we recommend to exploit the  $\model CB^5$ model for VGG-16 unless there are strong reasons in support of a different choice. For ResNet-50 the choice would be between $\model CB^3$ and $\model CB^4$ (but definitely not $\model CB^5$). Third, regression should be performed from $\model FC^2$ in VGG-16 (we do not recommend any of the tested alternatives) and either from $\model GAP$ or from $\model CB^5$ in ResNet-50. Finally, all the TIRs should be tested, since their optimality depends upon the networ and dataset in a statistically significant manner, except for heatmap regression, which shows consistently bad performance when used with vanilla deep regression.

% {\color{orange} Add HM discussion}

Very importantly, in case of limited resources (e.g. computing time), taking the suboptimal choice may not lead to a crucial difference with respect to the perfect combination of parameters. However, one must avoid the cases in which the method is proven to be significantly worse than the other, because in those cases performances (in train, validation and test) have proven to be evidently different.

\section{Data Pre-Processing}
\label{sec:prepro}
%\section{Data Pre-Processing}
%\label{sec:prepro}
In this section we discuss the different data pre-processing techniques that we consider in our benchmark. Because VGG-16 has two fully connected layers, we are constrained to use the pre-defined input size of $224\times224$ pixels. Hence, we systematically resize the images to this size. For the sake of a fair comparison, the very same input images are given to both networks. Firstly, we evaluate the impact of mirroring the training images (not used for test). Table~\ref{tab:VGG-resnet-mirroring} reports the results when evaluating the impact of mirroring. %% {\color{red} Numbers reported for training and validation are computed only on the non-mirrored images in order to compare scores computed on the exact same images. Therefore, in the case where mirroring is used, the values do not correspond exactly to the loss that has been optimized at training time. When comparing with Tables ~\ref{tab:VGG-resnet-BN}, to~\ref{tab:pooling}, we observe that these values are similar to those computed on the whole training set.}\doubtSte{I think we should report the numbers on full dataset. It sounds bullshit to me}
The conclusion is unanimous: mirroring is statistically better for all configurations. In addition, in most of the cases the confidence intervals are disjoint meaning that with high probability the output obtained when training with mirroring will have lower error than training without mirroring.

Since the three data sets used in our study have different characteristics, the pre-processing steps differ from data set to data set. Importantly, in all cases the pre-processing baseline technique is devised from the common usage of the data sets in the recent literature. Below we specify the baseline pre-processing as well as other tested pre-processing alternatives for each data set.

\whencolumns{\begin{figure}[t]}{\begin{figure}[t]}
  \centering
  \subfloat[Original]{
    \includegraphics[height=0.09\textwidth]{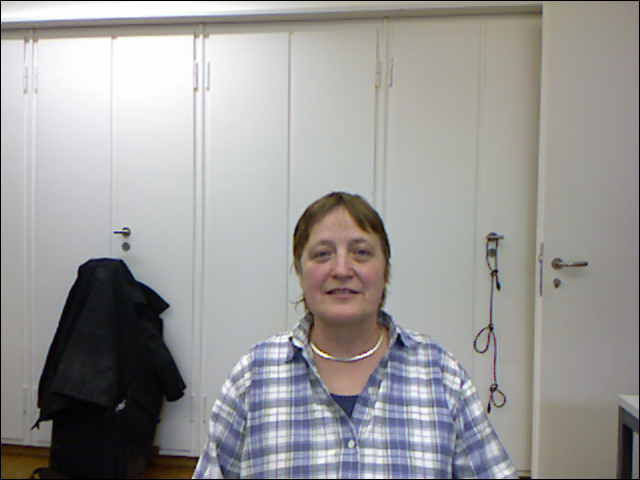}
    \label{biwi_orig}
  }
  \subfloat[\model 224]{
    \includegraphics[height=0.09\textwidth]{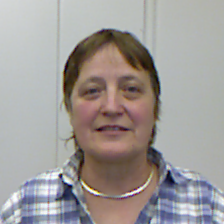}
    \label{biwi_224}
  }
    \subfloat[\model 64-$\boldsymbol\mu$Pad]{
  \includegraphics[height=0.09\textwidth]{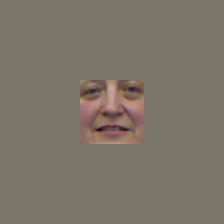}
    \label{biwi_64mp}}
  \subfloat[\model 64-0Pad]{\hspace{-2mm}
  \includegraphics[height=0.09\textwidth]{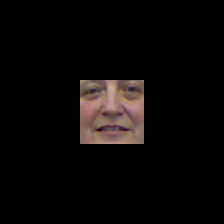}
    \label{biwi_64zp}
  }
  \\
    \subfloat[\model 64-Re]{\hspace{-2mm}
      \includegraphics[height=0.09\textwidth]{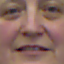}
    \label{biwi_64r}
    }
  \subfloat[\model 128-$\boldsymbol\mu$Pad]{
  \includegraphics[height=0.09\textwidth]{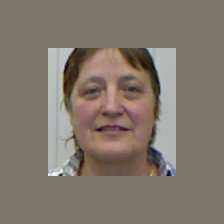}
    \label{biwi_128mp}
  }
  \subfloat[\model 128-0Pad]{\hspace{-2mm}
  \includegraphics[height=0.09\textwidth]{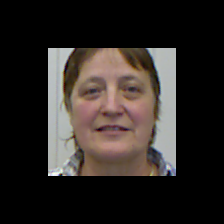}
    \label{biwi_128zp}
  }
  \subfloat[\model 128-Re ]{\hspace{-2mm}
  \includegraphics[height=0.09\textwidth]{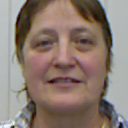}
    \label{biwi_128r}
  }
 % \vspace{-2mm}
  \\
%     \subfloat[Original]{\hspace{-2mm}
%       \includegraphics[height=0.09\textwidth]{"figures/paulaZahnOri"}
%     \label{FLD1_Ori}
%     }
%   \subfloat[$\model\epsilon{-}0$]{
%   \includegraphics[height=0.09\textwidth]{"figures/paulaZahn0"}
%     \label{FLD1_e0}}
%   \subfloat[$\model\epsilon{-}5$]{\hspace{-2mm}
%   \includegraphics[height=0.09\textwidth]{"figures/paulaZahn5"}
%     \label{FLD1_e5}
%   }
%     \subfloat[$\model\epsilon{-}15$]{\hspace{-2mm}
%   \includegraphics[height=0.09\textwidth]{"figures/paulaZahn15"}
%     \label{FLD1_e15}
%   }
%   \subfloat[$\model\epsilon{-}50$]{\hspace{-2mm}
%       \includegraphics[height=0.09\textwidth]{"figures/paulaZahn50"}
%     \label{FLD1_e50}
%   }
%   \vspace{-3mm}\\

    \subfloat[Original ]{\hspace{-2mm}
  \includegraphics[height=0.09\textwidth]{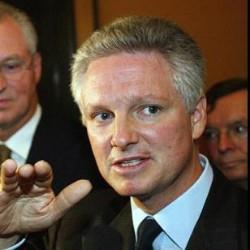}
    \label{FLD2_Ori}
  }
    \subfloat[$\model\epsilon{-}0$]{
  \includegraphics[height=0.09\textwidth]{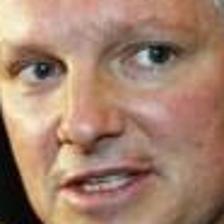}
    \label{FLD2_e0}}
  \subfloat[$\model\epsilon{-}5$]{\hspace{-2mm}
  \includegraphics[height=0.09\textwidth]{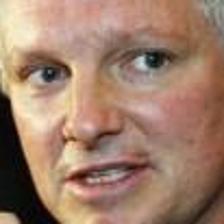}
    \label{FLD2_e5}
  }
  \subfloat[$\model\epsilon{-}15$]{\hspace{-2mm}
  \includegraphics[height=0.09\textwidth]{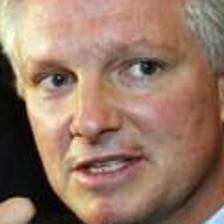}
    \label{FLD2_e15}
  }
  \subfloat[$\model\epsilon{-}50$]{\hspace{-2mm}
  \includegraphics[height=0.09\textwidth]{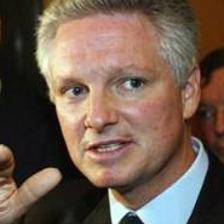}
    \label{FLD2_e50}
  }\\
  \subfloat[Original]{
    \begin{minipage}[c]{.15\textwidth}
      \centering
          \vspace{1mm}\includegraphics[width=1.7cm]{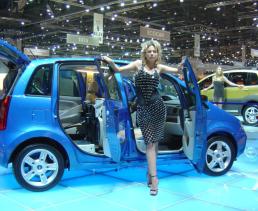}\vspace{2mm}
          \includegraphics[height=1.7cm]{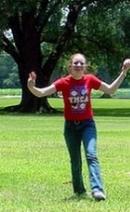}\vspace{1mm}

        \end{minipage}
    \label{lParse-Original}
  }\hspace{-1cm}
  \subfloat[\model 120x80]{
    \begin{minipage}[c]{.15\textwidth}
      \centering
          \includegraphics[height=1.7cm,width=1.7cm]{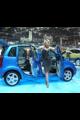}\vspace{1mm}
          \includegraphics[height=1.7cm,width=1.7cm]{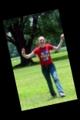}

        \end{minipage}
    \label{lParse-KAR120}
  }\hspace{-1.2cm}
  \subfloat[\model $\cancel{\boldsymbol\alpha}$-$\boldsymbol\mu$Pad]{
            \begin{minipage}[c]{.15\textwidth}
      \centering
    \includegraphics[height=1.7cm,width=1.7cm]{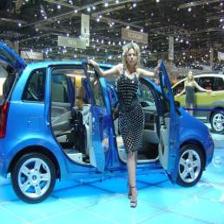}\vspace{1mm}
    \includegraphics[height=1.7cm,width=1.7cm]{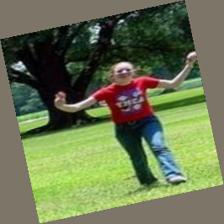}

        \end{minipage}
    \label{lParse-NKARmp}
  }\hspace{-1.2cm}
  \subfloat[\model $\boldsymbol\alpha$-$\boldsymbol\mu$Pad]{
    \centering
        \begin{minipage}[c]{.15\textwidth}
      \centering
    \includegraphics[height=1.7cm,width=1.7cm]{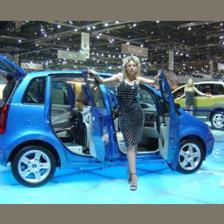}\vspace{1mm}
    \includegraphics[height=1.7cm,width=1.7cm]{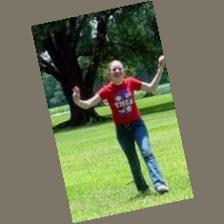}

    \end{minipage}
    \label{lParse-KARmp}
  }\\
      \subfloat[Original ]{\hspace{-2mm}
  \includegraphics[height=0.09\textwidth]{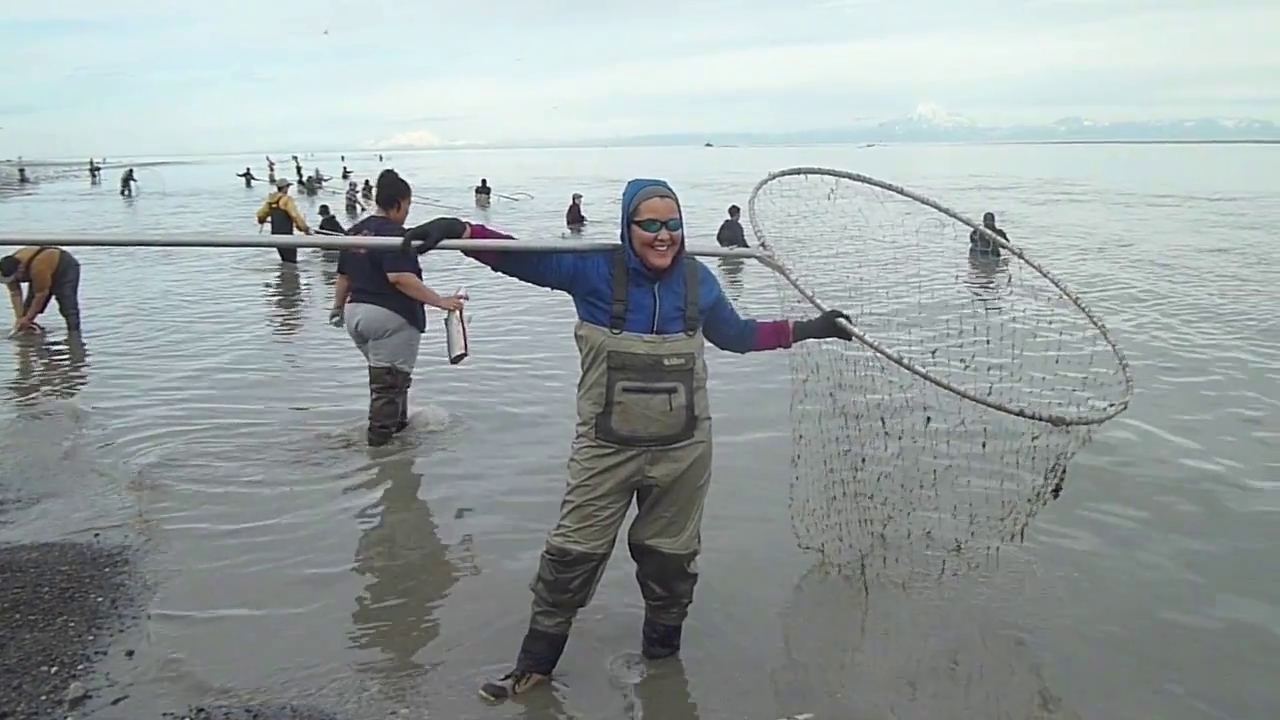}
    \label{MPII_Ori}
  }
    \subfloat[$\model\epsilon{-}0$]{
  \includegraphics[height=0.09\textwidth]{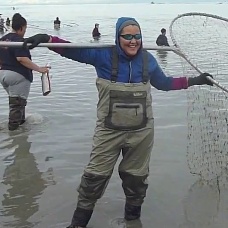}
    \label{MPII_e0}}
  \subfloat[$\model\epsilon{-}5$]{\hspace{-2mm}
  \includegraphics[height=0.09\textwidth]{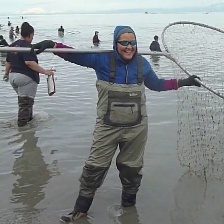}
    \label{MPII_e5}
  }
  \subfloat[$\model\epsilon{-}15$]{\hspace{-2mm}
  \includegraphics[height=0.09\textwidth]{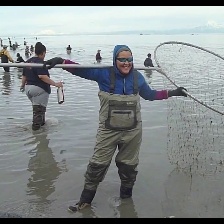}
    \label{MPII_e15}
  }
  \\

  % \hspace{-1.2cm}
%   \subfloat[\model $\cancel{\boldsymbol\alpha}$-0Pad]{
%             \begin{minipage}[c]{.15\textwidth}
%       \centering
%     \includegraphics[height=1.7cm,width=1.7cm]{"figures/PARSE_NOkeepAspectRatio_zeroPadding/im0005_angle_0"}\vspace{1mm}
%     \includegraphics[height=1.7cm,width=1.7cm]{"figures/PARSE_NOkeepAspectRatio_zeroPadding/im0001_angle_12"}
% 
%         \end{minipage}
%     \label{lParse-NKAR0p}
%   }\hspace{-1.2cm}
%   \subfloat[\model $\boldsymbol\alpha$-$\boldsymbol\mu$Pad]{
%     \centering
%         \begin{minipage}[c]{.15\textwidth}
%       \centering
%     \includegraphics[height=1.7cm,width=1.7cm]{"figures/PARSE_keepAspectRatio_meanPadding/im0005_angle_0"}\vspace{1mm}
%     \includegraphics[height=1.7cm,width=1.7cm]{"figures/PARSE_keepAspectRatio_meanPadding/im0001_angle_12"}
% 
%     \end{minipage}
%     \label{lParse-KARmp}
%   }\hspace{-1.2cm}
%   \subfloat[\model $\boldsymbol\alpha$-0Pad]{
%             \begin{minipage}[c]{.15\textwidth}
%       \centering
%     \includegraphics[height=1.7cm,width=1.7cm]{"figures/PARSE_keepAspectRatio_zeroPadding/im0005_angle_0"}\vspace{1mm}
%     \includegraphics[height=1.7cm,width=1.7cm]{"figures/PARSE_keepAspectRatio_zeroPadding/im0001_angle_12"}
% 
%         \end{minipage}
%     \label{lParse-KAR0p}
%   }
%  \vspace{-2mm}
  \caption{Pre-processed examples for the Biwi (first and second rows), FLD (third row), Parse (fourth row), and MPII (fifth row) data sets.\vspace{-5mm}}
\label{fig:pre-processing}
\whencolumns{\end{figure}}{\end{figure}}

% \begin{enumerate}
%  \item 
 \textbf{Biwi}. The baseline for the Biwi data set is inspired from~\cite{drouard2017head,Lathuiliere2017}, where the authors crop a $64\times64$ and a $224\times224$ window respectively, centered on the face. We investigate the use of three window sizes: $64\times64$, $128\times128$ and $224\times224$. The latter is referred to as {\bf 224}. When cropping windows smaller than $224\times224$ pixels, we investigate three possibilities: resize (denoted by {\bf 64-Re} and {\bf 128-Re}), padding with the mean value of ImageNet ({\bf 64-$\boldsymbol\mu$Pad} and {\bf 128-$\boldsymbol\mu$Pad}) and padding with zeros ({\bf 64-0Pad} and {\bf 128-0Pad}). Examples of these pre-processing steps are shown in Figures~\ref{biwi_orig}-\ref{biwi_128r}.
 
%  \item 
 \textbf{FLD}. In the case of the facial landmark data set, original images and face bounding boxes are provided. Similarly to the Biwi data set, the issue of the amount of context information is investigated. \cite{Sun2013} proposes to expand the bounding boxes and then to adopt a cascade strategy to refine the input regions of the networks. As we want to keep our processing as general as possible, we adopt the following procedure: the face bounding box is expanded by $\epsilon\%$ in each direction. We compare four different expanding ratios: $0\%$, $5\%$, $15\%$ and $50\%$. These are denoted with $\boldsymbol\epsilon${\model-0}, $\boldsymbol\epsilon${\model-5}, $\boldsymbol\epsilon${\model-15}, and $\boldsymbol\epsilon${\model-50} respectively, see Figures~\ref{FLD2_Ori} to~\ref{FLD2_e50}.
 
%  \item 
 \textbf{Parse}. When using this data set in~\cite{Belagiannis2015}, the images were resized to $120\times80$ pixels. In our case, we resize the images to fit into a rectangle of this size and pad to a squared image with zeros, followed by resizing to $224\times224$ pixels. This strategy is referred to as ${\model 120x80}$. We also consider directly resizing into $224\times 224$ images, hence without keeping the aspect ratio, and padding with the mean value of ImageNet ($\cancel{\boldsymbol\alpha}${\bf-}$\boldsymbol\mu${\bf Pad}), or keeping the aspect ratio ($\boldsymbol\alpha${\bf-}$\boldsymbol\mu${\bf Pad}). Examples are shown in Figures~\ref{lParse-Original}-\ref{lParse-KARmp}. The two last strategies are also employed using zero-padding ($\cancel{\boldsymbol\alpha}${\bf-0Pad} and $\boldsymbol\alpha${\bf-0Pad}).
 
\addnote[data-MPII]{2}{
\textbf{MPII}. In the case of the MPII data set, similarly to the Biwi and the FLD data sets, the question of the amount of context information is examined. Similarly to the FLD dataset, we adopt the following procedure: the bounding box is expanded by $\epsilon\%$ on each direction. We compare three different expanding ratios: $0\%$, $5\%$, $15\%$ denoted respectively by $\boldsymbol\epsilon${\model-0}, $\boldsymbol\epsilon${\model-5}, and $\boldsymbol\epsilon${\model-15}. Examples of these pre-processing steps are shown in Figures ~\ref{MPII_Ori} to~\ref{MPII_e15}. Note that, contrary to the FLD dataset, we do not evaluate larger expanding ratios, since the MPII dataset contains many images where several persons would appear within the bounding box if a larger expanding ratio were employed.}
 
%  and pad with zeros or the mean value of ImageNet (denoted by $\cancel{\boldsymbol\alpha}${\bf-0Pad} and $\cancel{\boldsymbol\alpha}${\bf-}$\boldsymbol\mu${\bf Pad}). Padding plays a role only in rotated images. Finally, we consider applying the same padding to images resized keeping the aspect ratio ($\boldsymbol\alpha${\bf0-Pad} and $\boldsymbol\alpha${\bf-}$\boldsymbol\mu${\bf Pad}). Examples of these strategies are shown in Figures~\ref{lParse-Original}-\ref{lParse-KARmp}.
% \end{enumerate}

\whencolumns{\begin{table}[h]}{\begin{table*}[t]}
\centering
 \caption{Impact of the mirroring (Mirr.) on VGG16 and ResNet50.\vspace{-2mm} }
 \resizebox{\textwidth}{!}{\begin{tabular}{ccllcccllccc}
 \toprule
Data & \multirow{3}{*}{Mirr.} && \multicolumn{4}{ c }{VGG-16} && \multicolumn{4}{ c }{ResNet-50}\\
\cmidrule{4-12}
 Set &  &&MAE test&MSE train& MSE valid &MSE test &&MAE test&MSE train& MSE valid &MSE test\\
   \midrule
   \multicolumn{1}{ c  }{\multirow{2}{*}{Biwi} } & \multicolumn{1}{ c }{Yes}  &&\ccg [3.66 3.79]*** & \ccg [4.33 4.41] & \ccg [12.18 12.56] &\ccg [18.77 20.20] &&\ccg  [3.60 3.71]*** & \ccg[1.25 1.27] & \ccg[21.49 22.25] &\ccg [17.15 18.14] \\ 
   \multicolumn{1}{ c  }{} & \multicolumn{1}{ c }{{No}} &&[5.49 5.67] & [9.09 9.36] & [24.22 25.61] & [42.55 45.73]  &&[4.46 4.57] & [1.39 1.42] & [19.70 20.83] & [27.34 28.60] \\ 

   \midrule
 \multicolumn{1}{ c  }{\multirow{2}{*}{FLD} } & \multicolumn{1}{ c }{Yes}  &&\ccg  [2.61 2.76]*** & \ccg[9.43 9.54] & \ccg[10.77 11.03] &\ccg [10.55 11.62]  &&\ccg  [1.96 2.05]*** & \ccg[5.21 5.25] & \ccg[6.85 6.95]  &\ccg [5.79 6.39]\\ 
   \multicolumn{1}{ c  }{} & \multicolumn{1}{ c }{{No}} && [3.06 3.24] & [14.13 14.38] & [15.29 15.73] & [14.28 15.68]  &&[2.05 2.13] & [4.41 4.46] & [7.04 7.20] & [6.43 7.15] \\ 
     
   \midrule
\multicolumn{1}{ c  }{\multirow{2}{*}{Parse} } & \multicolumn{1}{ c }{Yes}  &&\ccg  [4.90 5.59]*** &\ccg [2.38 2.41] &\ccg [27.74 28.48] &\ccg [41.35 52.13]  &&\ccg  [4.86 5.68]*** & \ccg [0.64 0.64] &\ccg [29.40 30.21] &\ccg [43.58 55.71] \\
   \multicolumn{1}{ c  }{} & \multicolumn{1}{ c }{{No}} && [5.08 5.76] & [2.31 2.36] & [28.98 30.14] & [45.15 57.08]  && [5.88 6.62] & [1.14 1.16] & [42.99 44.38] & [59.30 77.70] \\
   
   \midrule
\rowcolor{\rowcolorrevoneSec} & \multicolumn{1}{ c }{Yes}  &&\ccg [8.42 8.78]*** &\ccg [73.86 74.78] &\ccg [165.5 172.7] &\ccg [130.7 142.3]   &&\ccg [8.71 9.06]*** &\ccg [113.8 115.6] &\ccg [189.4 196.7] & \ccg [142.1 154.5] \\
  \rowcolor{\rowcolorrevoneSec} \multirow{-2}{*}{MPII}  & \multicolumn{1}{ c }{{No}} && [9.40 9.74] & [95.32 96.72] & [195.6 202.7] & [160.8 173.8]  && [9.59 10.12] & [81.99 83.36] & [216.8 225.5] & [172.7 188.7] \\     
   \bottomrule\vspace{-4mm}
 \end{tabular}}
 \label{tab:VGG-resnet-mirroring}
\whencolumns{\end{table}}{\end{table*}}

\whencolumns{\begin{table}[h!]}{\begin{table*}[t]}
\centering
 \caption{ Impact of the data pre-processing on VGG-16 and ResNet-50.\vspace{-2mm}\label{tab:data-preprocessing}}
 \resizebox{\textwidth}{!}{\begin{tabular}{ccllcccllccc}
\toprule
\multicolumn{2}{c}{Data Set \&} && \multicolumn{4}{c}{VGG-16} && \multicolumn{4}{c}{ResNet-50} \\
    \cmidrule{3-12}
\multicolumn{2}{c}{Pre-processing} && MAE test&MSE train& MSE valid &MSE test && MAE test&MSE train& MSE valid &MSE test \\
   \midrule
   \multirow{7}{*}{\rotatebox[origin=c]{90}{Biwi}} & {\bf 128-$\boldsymbol\mu$Pad} && [3.93 4.02] & [5.89 6.00] & [15.34 15.85] & [21.13 21.94] && [3.47 3.55] & [1.05 1.06] & [18.71 19.35] & [16.41 17.13] \\    
			 & {\bf 128-Re} && [3.87 3.97] & [4.98 5.06] & [13.23 13.75] & [19.80 20.79] && [3.18 3.25] & [0.97 0.98] & [16.37 16.91] & [13.63 14.25]  \\ 
			 & {\bf 128-0Pad} && [3.97 4.05] & [7.83 7.98] & [15.55 16.09] & [21.51 22.31] && [3.20 3.28] & [1.52 1.55] & [17.94 18.52] & [14.45 15.03] \\ 
			 & {\bf 64-$\boldsymbol\mu$Pad} && [4.48 4.63] & [6.09 6.22] & [21.52 22.31] & [27.91 29.77]  && [3.34 3.47] & [1.45 1.47] & [21.18 21.99] & [15.55 16.73] \\ 
			 & {\bf 64-Re} && [4.06 4.21] & [6.79 6.92] & [15.05 15.59] & [22.72 24.32] && [2.98 3.07]** & [0.99 1.00] & [13.18 13.66] & [12.20 12.90] \\ 
			 & {\bf 64-0Pad} &&  [4.80 5.02]$^{\circ\circ\circ}$ & [12.15 12.40] & [19.35 20.05] & [32.72 35.33]  && [3.29 3.42] & [3.34 3.40] & [19.21 20.01] & [14.91 16.02]  \\ 
			 & {\bf 224} && \ccg[3.66 3.79]*** & \ccg[4.33 4.41] & \ccg[12.18 12.56] & \ccg[18.77 20.20] &&\ccg[3.60 3.71]$^{\circ\circ\circ}$ & \ccg[1.25 1.27] & \ccg[21.49 22.25] & \ccg[17.15 18.14] \\  
  \midrule
    \multirow{4}{*}{\rotatebox[origin=c]{90}{FLD}} & $\boldsymbol\epsilon${\bf-0} && [2.25 2.36]*** & [7.61 7.69] & [8.94 9.11] & [7.69 8.57] && [1.63 1.73]*** & [2.41 2.44] & [5.07 5.16] & [4.23 4.63]\\
     & $\boldsymbol\epsilon${\bf-5} && \ccg[2.61 2.76] & \ccg[9.43 9.54] & \ccg[10.77 11.03] & \ccg[10.55 11.62]  && \ccg[1.96 2.05] & \ccg[5.21 5.25] & \ccg[6.85 6.95] & \ccg[5.79 6.39]\\
     & $\boldsymbol\epsilon${\bf-15} && [2.35 2.48] & [8.42 8.50] & [9.88 10.07] & [8.37 9.36]  && [2.00 2.08] & [4.48 4.53] & [6.59 6.70] & [5.95 6.43]\\
     & $\boldsymbol\epsilon${\bf-50} && [2.96 3.17]$^{\circ\circ\circ}$ & [11.68 11.82] & [13.33 13.60] & [13.32 15.31]  && [2.59 2.72]$^{\circ\circ\circ}$ & [7.12 7.18] & [9.92 10.07] & [10.11 11.20]\\
   \midrule
   \multirow{5}{*}{\rotatebox[origin=c]{90}{Parse}}& {\bf $\boldsymbol\alpha$-$\boldsymbol\mu$Pad} && [9.99 11.52]$^{\circ\circ\circ}$ & [8.47 8.62] & [114.5 117.5] & [161.4 226.1] && [11.61 13.63]$^{\circ\circ\circ}$ & [4.29 4.35] & [180.7 185.8] & [235.3 311.5] \\
			 & {\bf $\boldsymbol\alpha$-0Pad} && [9.72 11.13] & [9.77 9.95] & [116.1 119.2] & [165.3 212.9] && [10.90 12.54] & [3.34 3.39] & [158.5 163.7] & [207.9 279.6] \\ 
			 & {\bf $\cancel{\boldsymbol\alpha}$-$\boldsymbol\mu$Pad} && [8.54 9.56]*** & [6.22 6.35] & [106.9 109.4] & [125.5 158.7] && [9.30 11.28]*** & [2.83 2.87] & [140.3 144.5] & [160.8 218.3] \\ 
			 & {\bf $\cancel{\boldsymbol\alpha}$-0Pad} && [8.39 9.33]*** & [8.23 8.39] & [111.9 115.3] & [120.9 152.9] && [9.15 10.82]*** & [3.35 3.39] & [136.6 140.7] & [151.7 194.1] \\ 
			 & {\bf 120x80} && [9.55 11.34]  & [12.05 12.23]  & [130.6 134.3]  &  [161.0 215.8] && [9.14 10.57]*** & [3.40 3.45] & [136.5 140.4] &  [155.4 204.7] \\ 
  \midrule
\rowcolor{\rowcolorrevoneSec} & $\boldsymbol\epsilon${\bf-0} &&\ccg [8.42 8.78] &\ccg [73.96 74.62] &\ccg [165.3 170.4] &\ccg [130.7 142.3] &&\ccg [8.71 9.06] & \ccg[114.2 115.5] &\ccg [191.2 196.4] & \ccg[142.1 154.5] \\
\rowcolor{\rowcolorrevoneSec} & $\boldsymbol\epsilon${\bf-5} && [8.15 8.44]* & [51.83 52.29] & [161.0 165.5] & [120.7 130.5] && [8.49 8.86]*** & [87.08 87.99] & [177.4 182.5] & [132.7 141.9] \\
\rowcolor{\rowcolorrevoneSec} \multirow{-3}{*}{\rotatebox[origin=c]{90}{MPII}} & $\boldsymbol\epsilon${\bf-15} && [10.43 10.83]$^{\circ\circ\circ}$ & [134.8 136.8] & [226.0 231.0] & [186.7 205.8] && [10.02 10.39]$^{\circ\circ\circ}$ & [126.0 127.3] & [224.4 230.5] & [174.8 189.5] \\
   \bottomrule%\vspace{-7mm}
    \end{tabular}}
\whencolumns{\end{table}}{\end{table*}}

% \subsection{Data Preprocessing}
Table~\ref{tab:data-preprocessing} reports the results obtained by the different pre-processing techniques for each data set for both VGG-16 and ResNet-50. The point locations are represented by their pixel Cartesian coordinates in the case of FLD and Parse. When we evaluate a data pre-processing strategy, the images are geometrically modified and the same transformation is applied to the annotations. Consequently, the errors cannot be directly compared between two different pre-processing strategies. For instance, in the last row of Figure~\ref{fig:pre-processing} a  5-pixel error for the right elbow location may be acceptable in the case of {\model $\cancel{\boldsymbol\alpha}$-0Pad} but may correspond to a confusion with a shoulder location in the case of {\model $\boldsymbol\alpha$-0Pad}. In order to compare the pre-processing strategies, we transform all the errors into a common space. We choose common spaces such that the aspect ratio of the original images is kept. For Parse, the errors are compared in the original image space (i.e. before any resize or crop operation). In all the experiments of Section~\ref{sec:variant}, the errors are reported in the space of {\model 120x80}. This choice is justified by the fact that the MSE reported are the exact loss values used to optimize and proceed to early stopping. The drawback of this choice is that the errors on Parse obtained in Tables~\ref{tab:VGG-resnet-BN}-\ref{tab:VGG-resnet-mirroring} are not directly comparable with those of Table~\ref{tab:data-preprocessing}. In the case of FLD, $\boldsymbol\epsilon${\bf-0} space corresponds to original detections. However, as the transformations between spaces are only linear scalings, the comparison can be performed in any space without biasing the results. Therefore, we chose to report the errors in the space corresponding to $\boldsymbol\epsilon${\bf-5} to allow direct comparison with Tables ~\ref{tab:VGG-resnet-BN} to~\ref{tab:VGG-resnet-mirroring}. In the case of Biwi, as the head angle is independent of the pre-processing strategy, no transformation is required to compare the strategies.

Regarding the experiments on Biwi, we notice that the best strategy for VGG-16 is {\model 224}, whereas for ResNet-50 is {\model 64-Re}. Having said that, the differences with the second best (in terms of the confidence interval) are below 1 degree in MAE. Interestingly, we observe that in both cases two strategies are significantly worse: {\model 64-0Pad} for VGG-16 and $\model 224$ for ResNet-50. The fact that the same strategy $\model 224$ is significantly the best for VGG-16 and significantly the worst for ResNet-50 demonstrates that a serious ablation study of pre-processing strategies is required on Biwi. The experiments on the FLD data set are clearly more conclusive than the ones on Biwi. Indeed, for both architectures the $\boldsymbol\epsilon${\bf-0} strategy is significantly better and the $\boldsymbol\epsilon${\bf-50} is significantly worse. The differences range from approxiamtively 0.1 pixel (with respect to the second best strategy) to almost one pixel (with respect to the worst strategy). Regarding the experiments based on Parse, we obtain a statistical tie. Indeed, {\bf $\cancel{\boldsymbol\alpha}$-$\boldsymbol\mu$Pad} and {\bf $\cancel{\boldsymbol\alpha}$-0Pad} with VGG-16 are better than the rest, but without statistical differences between them. A similar behavior is found in ResNet-50 including {\bf 120x80}.

\addnote[preproc-MPII]{2}{The last three rows of Table~\ref{tab:data-preprocessing} are devoted to the data pre-processing of MPII. The conclusion is coherent among the two base architectures: the strategy $\boldsymbol\epsilon${\bf-5} is significantly better and the strategy $\boldsymbol\epsilon${\bf-15} is significantly worse. We hypothesize that, compared to FLD, this is due to the complexity of the dataset. After visual inspection, we observe that a non-negligible amount of images of MPII, when pre-processed with $\boldsymbol\epsilon${\bf-15}, contain body parts from other people, and this troubles the training of the network.}

\whencolumns{\begin{table}[h]}{\begin{table*}[t]}
  \centering
  \caption{Comparison of different methods on the FLD data set. Failure rate are given in percentage.  We report the median and best run behavior of the worse and best data pre-processing strategies for each baseline network.}
  \begin{tabular}{llccccccc}
  \toprule
    && Left Eye & Right Eye & Nose & Left Mouth& Right Mouth && Average\\
  \midrule
    VGG-16 $\boldsymbol\epsilon${\bf-50} && $11.65/7.23$ & $9.64/6.83$ & $16.47/13.65$ & $10.04/10.04$ & $12.45/8.43$ && $12.05/9.24$\\
    ResNet-50 $\boldsymbol\epsilon${\bf-50} && $10.44/4.02$ & $6.83/4.02$ & $5.22/7.63$ & $6.43/6.02$ & $6.83/5.22$ && $7.15/5.38$\\
  \midrule
    VGG-16 $\boldsymbol\epsilon${\bf-0} && $1.20/0.80$ & $0.40/0.00$ & $3.21/2.41$ & $3.61/2.81$ & $3.61/2.41$ && $2.41/1.69$\\
    ResNet-50 $\boldsymbol\epsilon${\bf-0} && $0.80/1.20$ & $0.00/0.00$ & $0.00/0.40$ & $2.01/0.80$ & $1.20/1.20$ && $0.80/0.72$\\
  \midrule
    Sun et al.\cite{Sun2013} && $0.67$ & $0.33$& $0.00$ & $1.16$& $0.67$ && $0.57$\\
  \bottomrule
  \end{tabular}
  \label{tab:sota-fld}
\whencolumns{\end{table}}{\end{table*}}

 \whencolumns{\begin{table}[h]}{\begin{table*}[t]}
    \centering
\caption{Comparison of different methods on the Parse database. Strict PCP scores are reported. We report the median and best run behavior of the worse and best data pre-processing strategies for each baseline network.}
  \begin{tabular}{lcccccccc}
    \toprule
	& Head & Torso & Upper Legs &Lower Legs & Upper Arms& Lower Arms && Full Body\\
    \midrule
        VGG-16 {\model $\boldsymbol\alpha$-$\boldsymbol\mu$Pad} &   $68.85/70.49$ & $85.25/83.61$ & $77.05/79.51$ & $63.11/63.93$ & $45.90/45.08$ & $40.16/42.62$ && $60.66/61.64$\\    
        ResNet-50 {\model $\boldsymbol\alpha$-$\boldsymbol\mu$Pad} &$57.4/65.6$ & $68.9/73.8$ & $71.3/74.6$ & $55.7/62.3$ & $39.3/45.1$ & $36.1/41.0$ && $53.1/58.5$ \\
    \midrule
        VGG-16 {\model $\cancel{\boldsymbol\alpha}$-0Pad} & $77.05/68.85$ & $90.16/88.52$ & $80.33/81.15$ & $63.11/67.21$ & $50.82/52.46$ & $44.26/50.00$ && $64.43/65.90$\\
        ResNet-50 $\model 120\times80$& $67.2/68.9$ & $78.7/80.3$ & $77.9/77.9$ & $65.6/63.9$ & $45.1/50.0$ & $46.7/45.9$ && $61.6/62.1$\\
    \midrule
        Andriluka et al \cite{andriluka2009pictorial} & $72.7$ & $86.3$ & $66.3$ & $60.0$ & $54.0$ & $35.6$ && $59.2$\\
        Yang \& Ramanan \cite{yang2013articulated} & $82.4$ & $82.9$ & $68.8$ & $60.5$ & $63.4$ & $42.4$ && $63.6$\\
        Pishchulin et al. \cite{pishchulin2012articulated} & $77.6$ & $90.7 $ & $80.0$ & $70.0$ & $59.3$ & $37.1 $ && $66.1$\\
        Johnson et al. \cite{johnson2010clustered} & $76.8$ & $87.6$ & $74.7$ & $67.1$ & $67.3$ & $45.8$ && $67.4$\\
        OuYang et al. \cite{ouyang2014multi} & $89.3$ & $89.3$ & $78.0$ & $72.0$ & $67.8$ & $47.8$ && $71.0$\\
        Belagiannis et al. \cite{Belagiannis2015} &$91.7$ & $98.1$ & $84.2$ & $79.3$ & $66.1$ & $41.5$ && $73.2$\\
    \bottomrule
  \end{tabular}
  \label{tab:sota-parse}
\whencolumns{\end{table}}{\end{table*}}

% \subsubsection{Discussion on Data Preprocessing and Mirroring}
The results on data pre-processing and mirroring behave quite differently. While mirroring results are consistent over data sets and networks, and therefore we recommend to systematically use mirroring for training, the results on data pre-processing require more detailed discussion. Indeed, one must be extremely careful when choosing the pre-processing on a data set. In the three cases we can see how the performance can present strong variations depending on the pre-processing used. Even more dangerous, when comparing the two architectures, the conclusions of which is best can change depending on the pre-processing. More generally, the superiority of a newly developed method 
%(potentially under review) 
with respect to the state-of-the-art may strongly depend on the data pre-processing. Ideally, standard ways to pre-process data so as to avoid unsupported conclusions should be established. In the case of Biwi, fixing one pre-processing strategy would  bias the decision either towards ResNet-50 ({\bf 64-Re} or {\bf 64-0Pad}), or towards VGG-16 ($\model 224$). But the fairest comparison would be ResNet-50 with {\bf 64-Re} against VGG-16 with $\model 224$. This indicates a clear interest on discussing different pre-processing strategies when presenting a new data set/architecture.

\section{Positioning of Vanilla Deep Regression}
\label{sec:application}
%\section{Positioning of Vanilla Deep Regression}
%\label{sec:application}
The aim of this section is to position the vanilla deep regression methods based on VGG-16 and ResNet-50 with respect to the state-of-the-art. Importantly, the point here is not to outperform all previous methods that are specifically designed for each of the studied tasks, but rather to understand how far or how close a vanilla deep regression method is from the state-of-the-art. Another question is whether or not a ``correctly fine-tuned base network'' (meaning having chosen the optimal network variant and pre-processing strategy) compares to some of the methods in the literature.

\begin{table}[t]
  \centering
  \caption{Comparison of different methods on the Biwi head-pose data set (MAE in degrees). The superscript$^\dagger$ denotes the use of extra training data (see~\cite{Liu2016} for details).  We report the median and best run of the worst and best data pre-processing strategies for each baseline network.}
  \whencolumns{%
\begin{tabular}{lcccc}
    \toprule
	& Pitch & Yaw & Roll &Mean\\
    \midrule
        VGG-16 {\bf 64-0Pad} & $10.11/4.75$ & $4.70/3.82$ & $4.16/4.18$ & $6.33/4.25$\\
        ResNet-50 {\model 224} & $4.73/4.53$ & $2.96/3.49$ & $4.91/4.10$ & $4.20/4.04$\\
        VGG-16 {\model 224} &  $4.51/\bf 4.02$ & $4.68/3.74$ & $3.22/3.28$ & $4.14/3.68$\\
        ResNet-50 {\model 64-Re} & $5.98/5.22$ & $2.39/\bf2.37$ & $3.93/4.04$ & $4.10/3.88$\\
    \midrule
        Liu et al.\cite{Liu2016}& $6.1$ & $6.0$ & $5.7 $&$5.94$\\
        Liu et al.\cite{Liu2016}$^{\dagger}$& $4.5$ & $4.3 $ & $\bf 2.4 $&$3.73$\\
        Mukherjee et al.\cite{Mukherjee2015}& $5.18$ & $5.67$ & $-$&$5.43$\\
        Drouard et al.\cite{drouard2017head}& $5.43$ & $4.24$ & $4.13$& $4.60$ \\
        Lathuiliere et al.\cite{Lathuiliere2017}& $4.68$ & $3.12$& $3.07$&$\bf 3.62$\\
    \bottomrule
  \end{tabular}
  }{%
  \resizebox{\linewidth}{!}{%
\begin{tabular}{lcccc}
    \toprule
	& Pitch & Yaw & Roll &Mean\\
    \midrule
        VGG-16 {\bf 64-0Pad} & $10.11/4.75$ & $4.70/3.82$ & $4.16/4.18$ & $6.33/4.25$\\
        ResNet-50 {\model 224} & $4.73/4.53$ & $2.96/3.49$ & $4.91/4.10$ & $4.20/4.04$\\
    \midrule
        VGG-16 {\model 224} &  $4.51/ 4.02$ & $4.68/3.74$ & $3.22/3.28$ & $4.14/3.68$\\
        ResNet-50 {\model 64-Re} & $5.98/5.22$ & $2.39/2.37$ & $3.93/4.04$ & $4.10/3.88$\\
    \midrule
        Liu et al.\cite{Liu2016}& $6.1$ & $6.0$ & $5.7 $&$5.94$\\
        Liu et al.\cite{Liu2016}$^{\dagger}$& $4.5$ & $4.3 $ & $2.4 $&$3.73$\\
        Mukherjee et al.\cite{Mukherjee2015}& $5.18$ & $5.67$ & $-$&$5.43$\\
        Drouard et al.\cite{drouard2017head}& $5.43$ & $4.24$ & $4.13$& $4.60$ \\
        Lathuiliere et al.\cite{Lathuiliere2017}& $4.68$ & $3.12$& $3.07$&$3.62$\\
    \bottomrule
  \end{tabular}
  }
  }
\label{tab:sota-biwi}
\end{table}

\begin{table}[t]
  \centering
  \caption{\comRev{ Comparison on the MPII single person body-pose data set. PCKh scores are reported.  We report the median and best run of the baseline networks.}}
  \whencolumns{%
%% \begin{tabular}{>{\color{darkgreen}}l>{\color{darkgreen}}c}
\begin{tabular}{lc}

  \toprule
	& Total PCKh\\
    \midrule
        VGG-16  & $61.6/62.5$\\
        ResNet-50  & $61.1/67.5$\\
        \midrule
        Tompson et al. \cite{tompson2014joint}& $79.6$\\
        Belagiannis et al. \cite{Belagiannis2017recurrent}& $89.7$ \\
        Newell et al. \cite{newell2016stacked}& $90.9$ \\
        Chu et al. \cite{Chu2017multi}& $91.5$\\
        Chen et al. \cite{Chen2017AdversarialPA}& $91.9$\\
        Yang et al. \cite{Yang_2017_ICCV}& $92.0$ \\
        Nie et al. \cite{Nie_2018_CVPR}& $92.4$ \\
        \bottomrule
  \end{tabular}
  }{%
    %% \resizebox{\linewidth}{!}{%
  
%% \begin{tabular}{>{\color{darkgreen}}l>{\color{darkgreen}}c}
\begin{tabular}{lc}

  \toprule
	& Total PCKh\\
    \midrule
       VGG-16  & $61.6/62.5$\\
        ResNet-50  & $61.1/67.5$\\
        \midrule
        Tompson et al. \cite{tompson2014joint}& $79.6$\\
        Belagiannis et al. \cite{Belagiannis2017recurrent}& $89.7$ \\
        Newell et al. \cite{newell2016stacked}& $90.9$ \\

        Chu et al. \cite{Chu2017multi}& $91.5$\\
        Chen et al. \cite{Chen2017AdversarialPA}& $91.9$\\
        Yang et al. \cite{Yang_2017_ICCV}& $92.0$ \\
        Nie et al. \cite{Nie_2018_CVPR}& $92.4$ \\

        \bottomrule
  \end{tabular}
  %% }
  }
\label{tab:sota-mpii}
\end{table}

In order to compare to the state-of-the-art, we employ the metric proper to each problem. Importantly, since the statistical tests are run on the per-image error, and these intermediate results are unavailable for state-of-the-art methods, we cannot run the same experimental evaluation as in previous sections. Furthermore, given that the experimental protocols of many papers are not fully detailed (we take the numbers directly from the references), we are uncertain that all row of Tables \ref{tab:sota-fld}, \ref{tab:sota-parse} and \ref{tab:sota-biwi},  are equally significant. In other words, the differences exhibited between the various state-of-the-art methods are to be taken with extreme care, since there is not other metric than the average performance (of probably one single run with a particular network design and data pre-processing strategy).

Since in the previous section we observed that the pre-processing strategy is crucial for the performance, we report the results of the two baseline architectures (Table~\ref{tab:baselines}) combined with the best and worst data pre-processing strategy for each architecture and each problem. For each of these four combinations of base architecture and pre-processing strategy, we run five network trainings and report the performance of the ``best'' and ``average'' runs (selected from the overall performance of the problem at hand, i.e., last column of the tables below). Consequently, it is possible that for a particular sub-task the ``average'' run has better performance than the ``best'' run. 

\begin{figure*}\center
\subfloat[Biwi]{\includegraphics[height=0.153\textwidth]{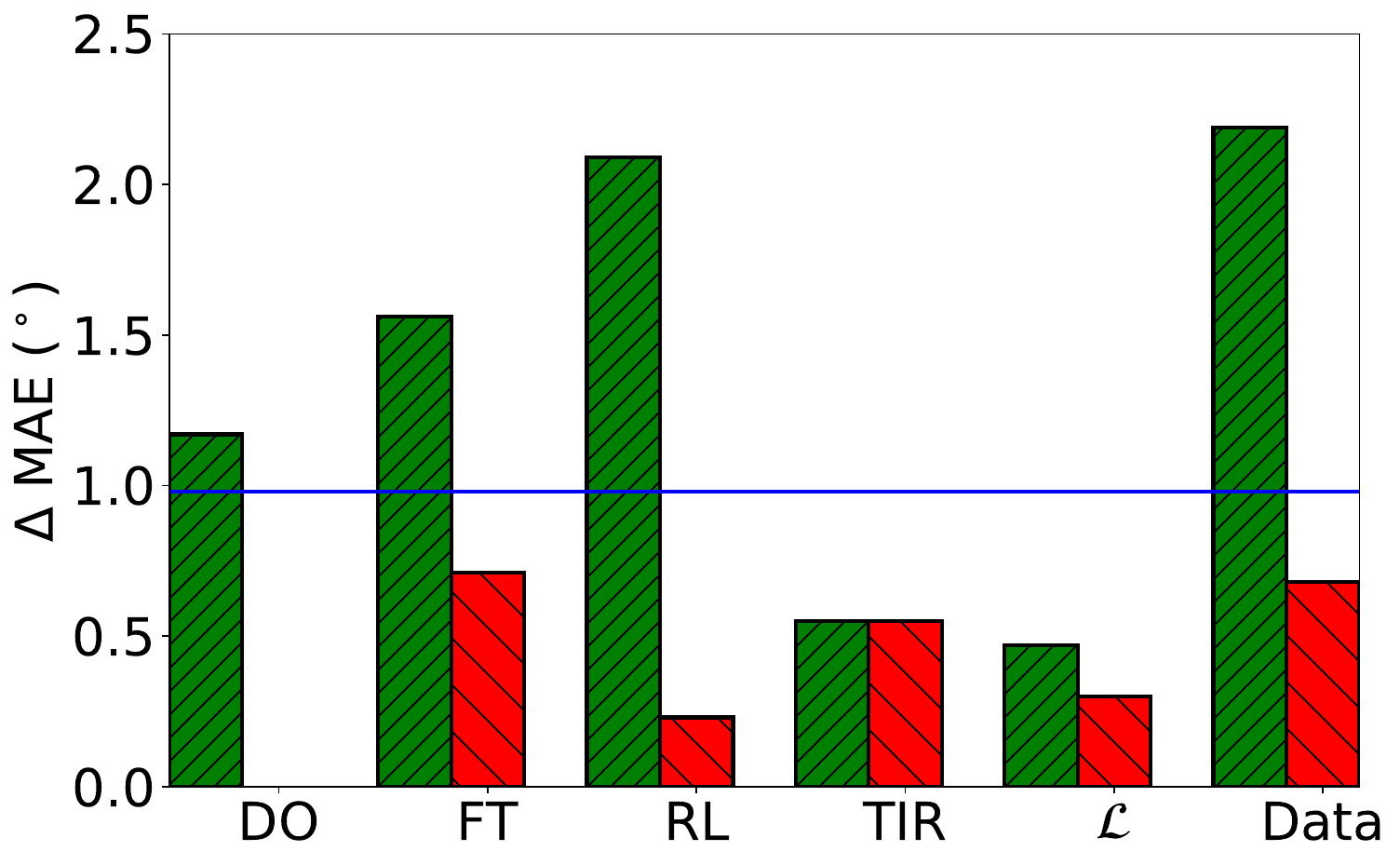}
\label{medianImprovementBiwi}} 
\subfloat[FLD]{\includegraphics[height=0.153\textwidth]{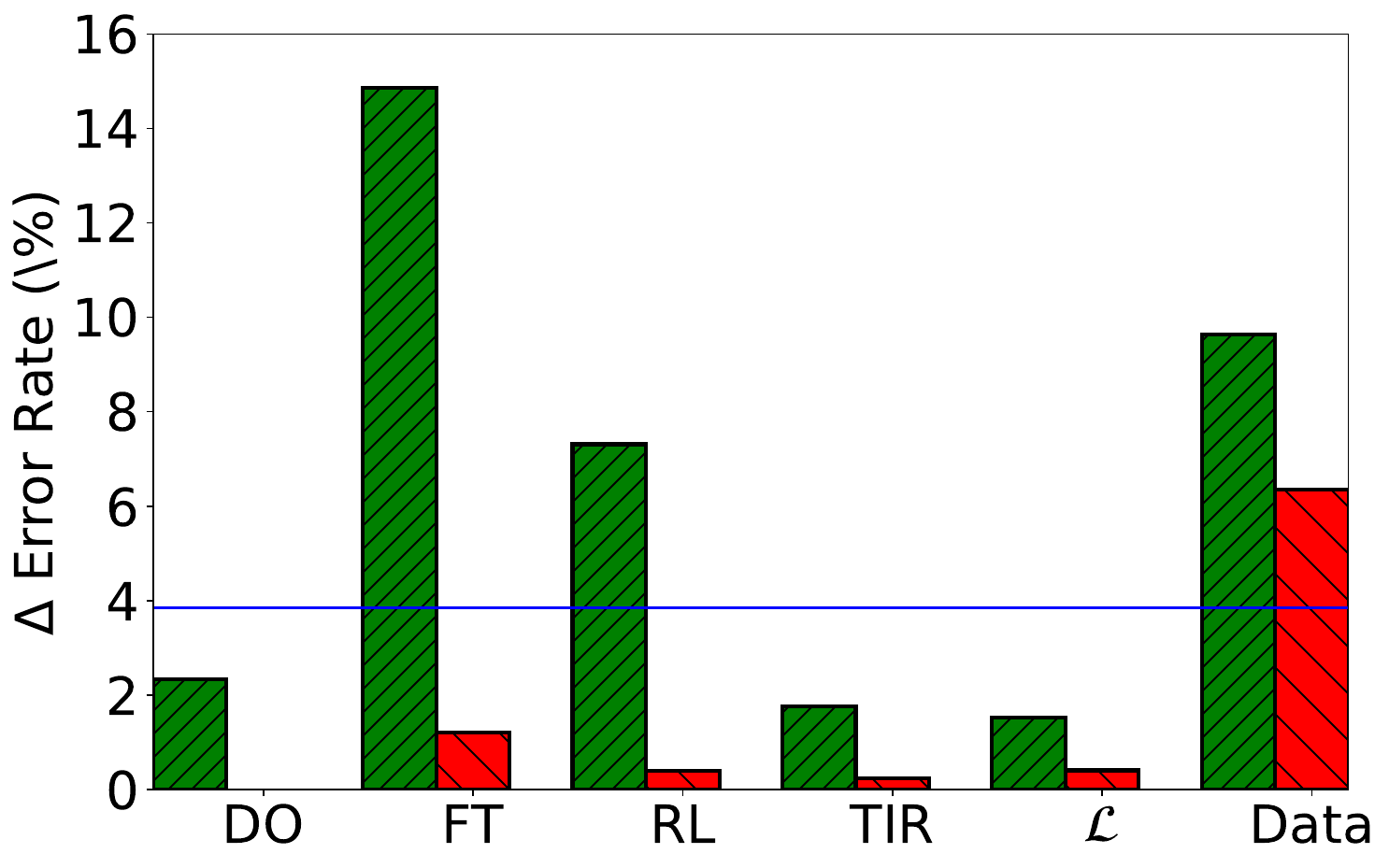}
\label{medianImprovementFLD}}
\subfloat[Parse]{\includegraphics[height=0.153\textwidth]{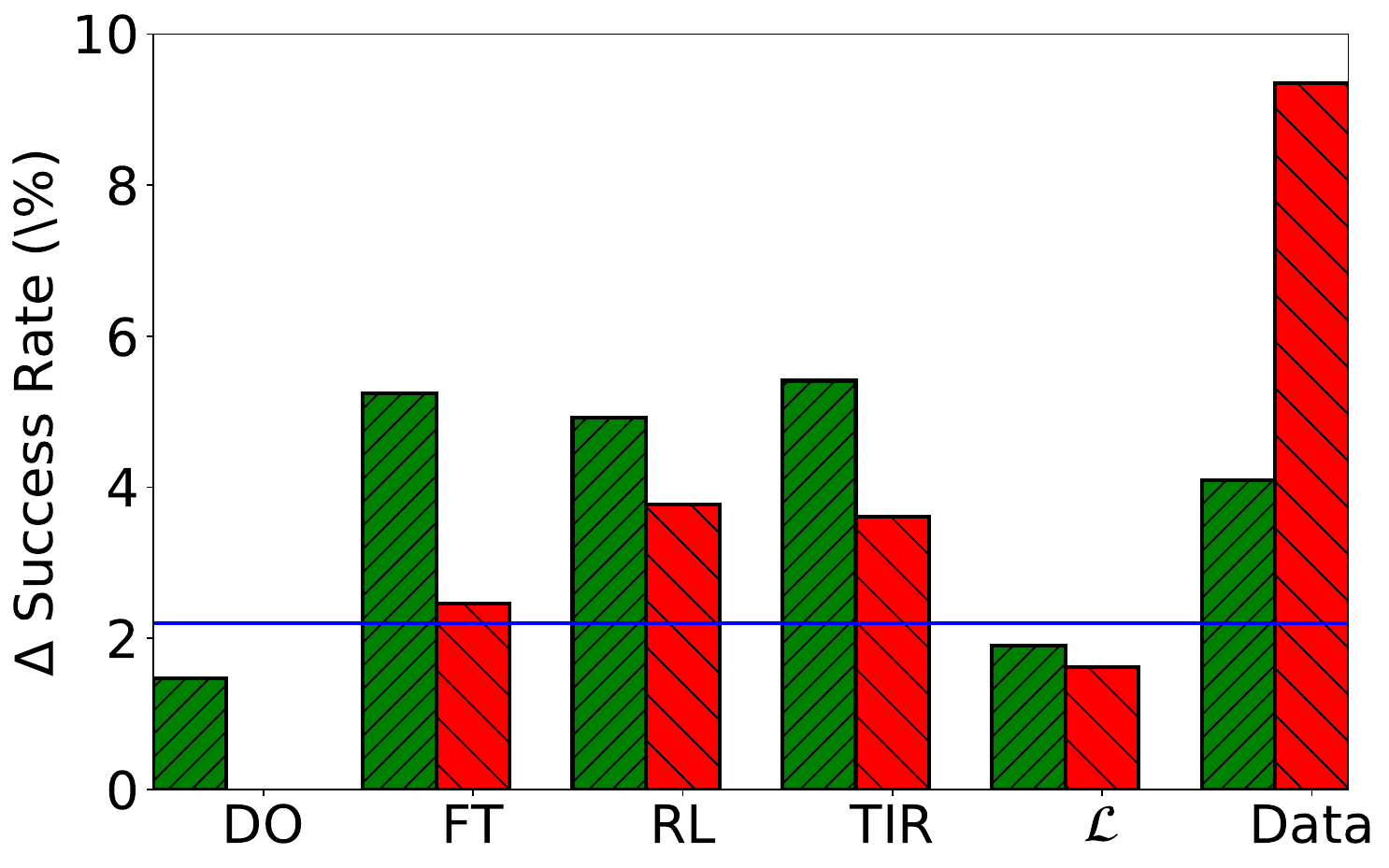}
\label{medianImprovementFBP}}
\subfloat[\comRev{MPII}]{\includegraphics[height=0.153\textwidth]{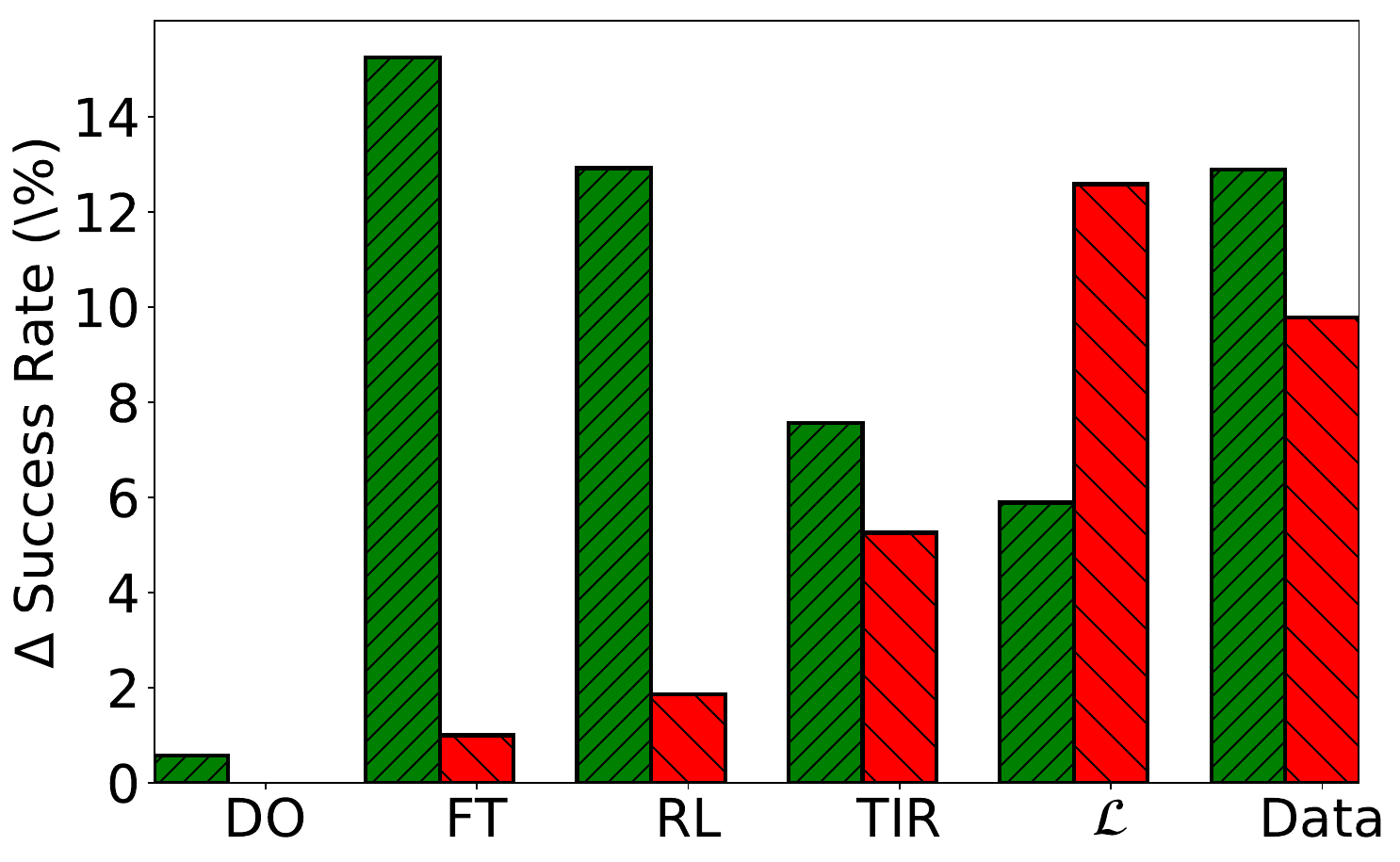}
\label{medianImprovementMPII}}

%% \subfloat[Data Pre-processing]{\includegraphics[height=0.25\textwidth]{figures/medianImprovementPreProcessing.pdf}
%% \label{medianImprovementPreProcessing}} 
%% \caption{Improvement in performance between the best and worst median runs for each configuration. The three plots on the left show the improvement on Biwi (a), FLD (b) and Parse (c) considering the five network variants used. The one on the right shows the improvement obtained in pre-processing for the same data sets. We display performance in terms of the specific metric per problem. }
\caption{Performance improvement between the best and worst median runs for each configuration on (a) Biwi, (b) FLD\comRev{, (c) Parse and (d) MPII}, for four network variants and data pre-processing (Data), using VGG-16 (green \comRev{ and ascending lines}) and ResNet-50 (red \comRev{and descending lines}). Horizontal blue lines indicate the improvement obtained by recent approaches when they outperform existing methods. Importantly, all the methods compared in this figure obtained statistically significant differences with $p<0.001$, except for DO on Parse.}

\label{medianImprovements}
\end{figure*}

Table~\ref{tab:sota-fld} reports the failure percentage of different methods on the FLD dataset  where errors 
larger than $5\%$ of the bounding box width are counted as failures. First of all, a correctly fine-tuned simple regressor is in competition with the state-of-the-art. Even if in average the best vanilla deep regression methods do not outperform the state of the art, they obtain quite decent results (e.g.\ the average run with the optimal strategy for ResNet-50 is only $0.23\%$ worse than the state of the art).

Regarding the Parse dataset, Table~\ref{tab:sota-parse} reports the results of six published methods. Performance is measured using the strict PCP (Percentage of Correctly estimated Parts) score. According to strict PCP, a limb is considered as correctely estimated if the distances between the estimated and true joint locations are smaller than $50\%$ of the limb length.  Vanilla deep regression does not outperform the state of the art. However, it is important to notice that regarding the Head, Torso, Upper Legs and Lower Legs, the best vanilla deep regression variant outperforms half of existing methods and in the case of Lower Arms, it outperforms the state of the art. On an average, vanilla deep regression competes well with half of the methods in the literature. We believe this is a very interesting result from two perspectives. On the one side, vanilla deep regression methods compete with the state-of-the-art in different tasks as long as they are properly fine tuned. On the other side, methods specifically designed for a problem may often outperform standard baselines, and therefore it is worth pursuing research in most applications. In other words, standard baselines may offer a good starting point, but developing problem-specific methods is worth to push the performances.

Table~\ref{tab:sota-biwi} reports the results on the Biwi dataset using MAE. Correctly fine-tuned deep regression is clearly competing well with the state of the art (even with a method using extra training data). Overall, the best deep regression is very close to the state-of-the-art on the Biwi dataset.

\addnote[sota-MPII]{1}{
Finally, Table~\ref{tab:sota-mpii} reports the results on the MPII the test set using the PCKh score. PCKh score is a variant of the PCP score employed for the Parse dataset, where the success threshold is set to $50\%$ of the ground-truth head height. On this dataset, we observe that the state-of-the-art clearly outperforms vanilla deep regression, suggesting that, in some cases, adopting more complex and task-specific approaches can lead to better performance. %{\color{red} We should be careful here. One of the main conclusions of the paper is that ``a standard and general-purpose network (e.g. VGG-16 or ResNet-50) adequately tuned  can achieve results close to the state-of-the-art without having to resort to much more complex and ad-hoc regression models''. Here, it seems that we contradict this idea.} 
 In particular, in the case of full-body pose estimation, in opposition to our vanilla deep regression models, state of the art models exploit several assumptions specific to the human body structure. For instance, instead of regressing only the joints coordinates, recent approaches jointly estimate the limbs locations in order to better constrain the model\cite{Belagiannis2017recurrent,Nie_2018_CVPR}. A cascade regression strategy is also often employ to enforce consistency among predicted joint locations \cite{tompson2014joint,newell2016stacked,Chen2017AdversarialPA,Belagiannis2017recurrent,Chu2017multi}. Furthermore, at training time, state of the art methods usually exploit additional landmark visibility annotation \cite{Belagiannis2017recurrent,Chen2017AdversarialPA} that are not used by our general-purpose model. As conclusion, much better results are obtained, but at the cost of introducing more complex and ad-hoc approaches.}

\section{Overall Discussion}
\label{sec:global_discussion}
%\section{Overall Discussion}
%\label{sec:global_discussion}

Figure~\ref{medianImprovements} displays the difference, in terms of problem-specific metrics (see Section~\ref{sec:application}), between the best and worst median runs per configuration. More precisely, we compute the metric for each one of the 5 runs and, from there, compute the median for each configuration. Finally, we display the difference between the best and worst medians.  The main goal of this figure is to visualize the impact in performance of each network variant and the data-preprocessing per network and problem. We exclude those unequivocal cases where one particular configuration offers a clear and systematic improvement or deterioration with respect to other configurations. These are: batch normalization (see Table~\ref{tab:VGG-resnet-BN}), $\model CB^5$ when fine-tunning ResNet-50 (see Table~\ref{tab:VGG-ResNet-FTDepth}) \comRev{ and heatmap regression (see Table~\ref{tab:TIR})} and mirroring (see Table~\ref{tab:VGG-resnet-mirroring}). The results of VGG-16 are shown in green while those of ResNet-50 are shown in red.

The first element to consider is the improvement that can be achieved when the network variants are properly selected (up to 2$^{\circ}$ MAE in Biwi, up to 14\% error rate in FLD \comRev{ and MPII}, and up to 6\% success rate in Parse). Although, in Figure~\ref{medianImprovements}, the margins of improvement may seem small, in some cases we are dealing with performance increases that would be sufficient to clearly outperform prior results. In order to give a visual reference in this regard, for each problem, we add a horizontal line that indicates the improvement obtained by recent approaches when they outperform preceding methods.  Such lines are drawn by computing the difference in performance between the best performing method in Tables~\ref{tab:sota-biwi},~\ref{tab:sota-fld} and~\ref{tab:sota-parse} for Biwi, FLD and Parse, respectively, and the second best performing method in the corresponding original papers \cite{Lathuiliere2017}, \cite{Belagiannis2015}, and \cite{Sun2013}. \addnote[mpii-disc]{1}{ Note that we do not display the horizontal line in the case of the MPII. The reason for this choice is twofold. First, the scores obtained are significantly lower than those obtained by state-of-the-art methods. Consequently, the magnitude among variants of vanilla deep regression models (around $65\%$ PCKh score) cannot be fairly compared with the variation of the state of the art (around $85\%$ PCKh score). Second, because of the MPII evaluation protocol constraints explained in Sec.~\ref{sec:problems_data sets}, we report the score on different training and test sets. } We do not suggest that, in a particular task, the precise adjustment of, for instance FT, will provide results superior to the state of the art. We only give a reference of the magnitude of the potential performance improvement (or deterioration) between different variants of the same network, and claim that this potential improvement is not only significant but also far from being negligible.

The aspect that attracts the most immediate attention is that the margin of improvement in VGG-16 is generally larger than in ResNet-50, as shown by the gap between the top of the green and red columns in Figure~\ref{medianImprovements} (with the exceptions of TIR in Biwi, data-preprocessing in Parse\comRev{, and the loss in MPII}). 
Regarding the network variants, we conclude that VGG-16 obtains highly variable results while, on the contrary, the behavior of ResNet-50 is generally more stable. From a practical point of view, a substantial improvement can be expected from the careful configuration of the VGG-16 variants. With respect to the data pre-processing, the improvement may also be quite high, and which network is subject to larger improvement is highly dependent on the problem at hand.
 
Generally speaking, the most critical factors seem to be FT, RL and data pre-processing. The only exception is VGG-16 in Parse, where TIR offers a larger improvement than data pre-processing. As a consequence, it is preferable to invest time trying different FT and RL strategies than different variants of TIR or DO, which appear to be elements with least variability between the best and worst configurations. 

Finally, we observe that the impact of proper data pre-processing is crucial for the performance of vanilla deep regression. Moreover, the margin of improvement highly depends both on the problem at hand and on the architecture. In addition, given that for some datasets the optimal data pre-processing strategy depends on the network, we argue that the pre-processing can have huge impact on the results and we draw two conclusions. First, at practical level, we strongly recommend to compare different data pre-processing strategies in order to improve the performance. Second, at a scientific level, our analysis illustrate that the employed data pre-processing procedures must be carefully detailed when it comes to reporting results. Different pre-processing techniques must be evaluated and carefully explained in order to obtain reliable conclusions and reproducible results. To be sure of the potential improvement, one must provide the outcome of statistical tests and, if possible, the confidence interval for the median performance.

\section{Conclusion}
\label{sec:conclusions}
This manuscript presents the first comprehensive study on deep regression for computer vision. Motivated by a lack of systematic means of comparing two deep regression techniques and by the proliferation of deep architectures (also for regression in computer vision), we propose an experimental protocol to soundly compare different methods. In particular, we are interested in vanilla deep regression methods --short for convolutional neural networks with a linear regression layer-- and their variants (e.g. modifying the number of fine-tuned layers or the data pre-processing strategy). Each of them is run five times and $95\%$ confidence intervals for the median performance as well as statistical testours are reported. These measures are used to discern between variations due to stochastic effects from systematic improvements. The conducted extensive experimental evaluation allows us to extract some conclusions on what to do, what not to do and what to test when employing vanilla deep regression methods for computer vision tasks.
% 
% This manuscript presents the first comprehensive study on deep neural architectures for computer vision regression. Indeed, we aim to shade some light into how to use vailla deep regressors in computer vision and which parameters are more important and should be evaluated with more care, as well as how to avoid undesirable situations. To do that, we exploited two base architectures on three different regression computer vision problems. For every configuration tested, we train the network five times, and report confidence intervals of the median of the error. Additionally, we test these results for statistical significance. In this way we provide both an absolute and a relative measure of the performance. The extensive experimental evaluation carried on allowed us to extract some global recommendations (either positive or negative) as well as point the network/data configurations that need further comparison since conclusive evidence was not found.

In addition, we have shown that\comRev{, in three out of four problems, } correctly fine-tuned deep regression networks can compete with problem-specific methods in the literature that are entirely devoted to solve only one task. Importantly, the overall set of experiments clearly points out the need of a proper data-preprocessing strategy to obtain results that are meaningful and competitive with the state-of-the-art. This behavior should encourage the community to use very clean experimental protocols that include transparent and detailed descriptions of the different pre-processing strategies used. Ideally, papers should evaluate the impact of different pre-processing strategies when presenting a new dataset or a new method. We believe this would help the community to understand deep regression for computer vision and to discern systematic behaviors from those due to chance.

\bibliographystyle{IEEEtran}
\bibliography{main}

\begin{thebibliography}{10}
\providecommand{\url}[1]{#1}
\csname url@samestyle\endcsname
\providecommand{\newblock}{\relax}
\providecommand{\bibinfo}[2]{#2}
\providecommand{\BIBentrySTDinterwordspacing}{\spaceskip=0pt\relax}
\providecommand{\BIBentryALTinterwordstretchfactor}{4}
\providecommand{\BIBentryALTinterwordspacing}{\spaceskip=\fontdimen2\font plus
\BIBentryALTinterwordstretchfactor\fontdimen3\font minus
  \fontdimen4\font\relax}
\providecommand{\BIBforeignlanguage}[2]{{%
\expandafter\ifx\csname l@#1\endcsname\relax
\typeout{** WARNING: IEEEtran.bst: No hyphenation pattern has been}%
\typeout{** loaded for the language `#1'. Using the pattern for}%
\typeout{** the default language instead.}%
\else
\language=\csname l@#1\endcsname
\fi
#2}}
\providecommand{\BIBdecl}{\relax}
\BIBdecl

\bibitem{fanelli2011real}
G.~Fanelli, J.~Gall, and L.~Van~Gool, ``Real time head pose estimation with
  random regression forests,'' in \emph{CVPR}, 2011, pp. 617--624.

\bibitem{zhu2012face}
X.~Zhu and D.~Ramanan, ``Face detection, pose estimation, and landmark
  localization in the wild,'' in \emph{CVPR}, 2012, pp. 2879--2886.

\bibitem{burgos2013robust}
X.~P. Burgos-Artizzu, P.~Perona, and P.~Doll{\'a}r, ``Robust face landmark
  estimation under occlusion,'' in \emph{ICCV}, 2013, pp. 1513--1520.

\bibitem{dantone2012real}
M.~Dantone, J.~Gall, G.~Fanelli, and L.~Van~Gool, ``Real-time facial feature
  detection using conditional regression forests,'' in \emph{CVPR}, 2012, pp.
  2578--2585.

\bibitem{agarwal20043d}
A.~Agarwal and B.~Triggs, ``3d human pose from silhouettes by relevance vector
  regression,'' in \emph{CVPR}, vol.~2, 2004, pp. II--II.

\bibitem{sun2012conditional}
M.~Sun, P.~Kohli, and J.~Shotton, ``Conditional regression forests for human
  pose estimation,'' in \emph{CVPR}, 2012, pp. 3394--3401.

\bibitem{Yan07}
S.~Yan, H.~Wang, X.~Tang, and T.~S. Huang, ``Learning auto-structured regressor
  from uncertain nonnegative labels,'' in \emph{CVPR}, 2007, pp. 1--8.

\bibitem{guo2008image}
G.~Guo, Y.~Fu, C.~R. Dyer, and T.~S. Huang, ``Image-based human age estimation
  by manifold learning and locally adjusted robust regression,'' \emph{IEEE
  TIP}, vol.~17, no.~7, pp. 1178--1188, 2008.

\bibitem{chou20132d}
C.-R. Chou, B.~Frederick, G.~Mageras, S.~Chang, and S.~Pizer, ``2d/3d image
  registration using regression learning,'' \emph{CVIU}, vol. 117, no.~9, pp.
  1095--1106, 2013.

\bibitem{niethammer2011geodesic}
M.~Niethammer, Y.~Huang, and F.-X. Vialard, ``Geodesic regression for image
  time-series,'' in \emph{MICCAI}, 2011, pp. 655--662.

\bibitem{Krizhevsky2012}
A.~Krizhevsky, I.~Sutskever, and G.~E. Hinton, ``{ImageNet Classification with
  Deep Convolutional Neural Networks},'' in \emph{NIPS}, 2012.

\bibitem{Szegedy2015}
C.~Szegedy, W.~Liu, Y.~Jia, P.~Sermanet, S.~Reed, D.~Anguelov, D.~Erhan,
  V.~Vanhoucke, and A.~Rabinovich, ``Going deeper with convolutions,'' in
  \emph{CVPR}, 2015.

\bibitem{Girshick2014}
R.~Girshick, J.~Donahue, T.~Darrell, and J.~Malik, ``{Rich Feature Hierarchies
  for Accurate Object Detection and Semantic Segmentation},'' in \emph{CVPR},
  2014.

\bibitem{Sermanet2014}
P.~Sermanet, D.~Eigen, X.~Zhang, M.~Mathieu, R.~Fergus, and Y.~Lecun,
  ``{Overfeat: Integrated recognition, localization and detection using
  convolutional networks},'' in \emph{ICLR}, 2014.

\bibitem{alameda2017viraliency}
X.~Alameda-Pineda, A.~Pilzer, D.~Xu, N.~Sebe, and E.~Ricci, ``Viraliency:
  Pooling local virality,'' in \emph{CVPR}, 2017.

\bibitem{Liu2016}
X.~Liu, W.~Liang, Y.~Wang, S.~Li, and M.~Pei, ``{3D head pose estimation with
  convolutional neural network trained on synthetic images},'' in \emph{ICIP},
  2016, pp. 1289--1293.

\bibitem{Toshev2014}
A.~Toshev and C.~Szegedy, ``{DeepPose: Human Pose Estimation via Deep Neural
  Networks},'' in \emph{CVPR}, 2014.

\bibitem{Belagiannis2015}
V.~Belagiannis, C.~Rupprecht, G.~Carneiro, and N.~Navab, ``Robust optimization
  for deep regression,'' in \emph{ICCV}, 2015.

\bibitem{Sun2013}
Y.~Sun, X.~Wang, and X.~Tang, ``Deep convolutional network cascade for facial
  point detection,'' in \emph{CVPR}, 2013.

\bibitem{lathuiliere2018deepgum}
S.~Lathuili{\`e}re, P.~Mesejo, X.~Alameda-Pineda, and R.~Horaud, ``Deepgum:
  Learning deep robust regression with a gaussian-uniform mixture model,'' in
  \emph{ECCV}, 2018, pp. 202--217.

\bibitem{hueber2015speaker}
T.~Hueber, L.~Girin, X.~Alameda-Pineda, and G.~Bailly, ``Speaker-adaptive
  acoustic-articulatory inversion using cascaded gaussian mixture regression,''
  \emph{TASLP}, vol.~23, no.~12, pp. 2246--2259, 2015.

\bibitem{liu16FashionLandmark}
Z.~Liu, S.~Yan, P.~Luo, X.~Wang, and X.~Tang, ``{Fashion Landmark Detection in
  the Wild},'' in \emph{ECCV}, 2016.

\bibitem{Lathuiliere2017}
S.~Lathuili\`{e}re, R.~Juge, P.~Mesejo, R.~Mu\~{n}oz Salinas, and R.~Horaud,
  ``{Deep Mixture of Linear Inverse Regressions Applied to Head-Pose
  Estimation},'' in \emph{CVPR}, 2017, pp. 7149--7157.

\bibitem{bulat2017far}
A.~Bulat and G.~Tzimiropoulos, ``How far are we from solving the 2d \& 3d face
  alignment problem? (and a dataset of 230,000 3d facial landmarks),'' in
  \emph{ICCV}, 2017.

\bibitem{rothe2016deep}
R.~Rothe, R.~Timofte, and L.~Van~Gool, ``Deep expectation of real and apparent
  age from a single image without facial landmarks,'' \emph{IJCV}, 2016.

\bibitem{Rogez_2017_CVPR}
G.~Rogez, P.~Weinzaepfel, and C.~Schmid, ``Lcr-net:
  Localization-classification-regression for human pose,'' in \emph{CVPR}, July
  2017.

\bibitem{LeCunBOM98}
Y.~LeCun, L.~Bottou, G.~B. Orr, and K.~M{\"{u}}ller, ``Effiicient backprop,''
  in \emph{Neural Networks: Tricks of the Trade}, 1998, pp. 9--50.

\bibitem{Nebauer1998}
C.~Nebauer, ``Evaluation of convolutional neural networks for visual
  recognition,'' \emph{IEEE TNN}, vol.~9, no.~4, pp. 685--696, 1998.

\bibitem{SzegedyVISW16}
C.~Szegedy, V.~Vanhoucke, S.~Ioffe, J.~Shlens, and Z.~Wojna, ``{Rethinking the
  Inception Architecture for Computer Vision},'' in \emph{CVPR}, 2016, pp.
  2818--2826.

\bibitem{SmithT16}
L.~N. Smith and N.~Topin, ``{Deep Convolutional Neural Network Design
  Patterns},'' \emph{CoRR}, vol. abs/1611.00847, 2016.

\bibitem{IthapuRS17}
V.~K. Ithapu, S.~N. Ravi, and V.~Singh, ``On architectural choices in deep
  learning: From network structure to gradient convergence and parameter
  estimation,'' \emph{CoRR}, vol. abs/1702.08670, 2017.

\bibitem{bengio2012}
Y.~Bengio, ``Practical recommendations for gradient-based training of deep
  architectures,'' in \emph{Neural networks: Tricks of the trade}.\hskip 1em
  plus 0.5em minus 0.4em\relax Springer, 2012, pp. 437--478.

\bibitem{GreffSKSS17}
K.~Greff, R.~K. Srivastava, J.~Koutn{\'{\i}}k, B.~R. Steunebrink, and
  J.~Schmidhuber, ``{LSTM:} {A} search space odyssey,'' \emph{{IEEE} TNNLS},
  vol.~28, no.~10, pp. 2222--2232, 2017.

\bibitem{ChandrasekharLM16}
V.~Chandrasekhar, J.~Lin, O.~Mor{\`{e}}re, H.~Goh, and A.~Veillard, ``{A
  practical guide to CNNs and Fisher Vectors for image instance retrieval},''
  \emph{Signal Processing}, vol. 128, pp. 426--439, 2016.

\bibitem{MISHKIN2017}
D.~Mishkin, N.~Sergievskiy, and J.~Matas, ``Systematic evaluation of
  convolution neural network advances on the imagenet,'' \emph{CVIU}, vol. 161,
  pp. 11--19, 2017.

\bibitem{chatfield2014return}
K.~Chatfield, K.~Simonyan, A.~Vedaldi, and A.~Zisserman, ``Return of the devil
  in the details: Delving deep into convolutional nets,'' in \emph{BMVC}, 2014.

\bibitem{SaxenaV16}
S.~Saxena and J.~Verbeek, ``Convolutional neural fabrics,'' in \emph{NIPS},
  2016, pp. 4053--4061.

\bibitem{XieY17}
L.~Xie and A.~L. Yuille, ``Genetic {CNN},'' in \emph{CVPR}, 2017.

\bibitem{wilcoxon:test}
F.~Wilcoxon, ``{Individual comparisons by ranking methods},'' \emph{Biometrics
  Bulletin}, pp. 80--83, 1945.

\bibitem{simonyan2014very}
K.~Simonyan and A.~Zisserman, ``Very deep convolutional networks for
  large-scale image recognition,'' in \emph{ICLR}, 2015.

\bibitem{he2016deep}
K.~He, X.~Zhang, S.~Ren, and J.~Sun, ``Deep residual learning for image
  recognition,'' in \emph{CVPR}, 2016, pp. 770--778.

\bibitem{Ren2015FasterRT}
S.~Ren, K.~He, R.~B. Girshick, and J.~Sun, ``Faster r-cnn: Towards real-time
  object detection with region proposal networks,'' \emph{IEEE TPAMI}, vol.~39,
  pp. 1137--1149, 2015.

\bibitem{Fanelli2013}
G.~Fanelli, M.~Dantone, J.~Gall, A.~Fossati, and L.~Gool, ``{Random Forests for
  Real Time 3D Face Analysis},'' \emph{IJCV}, vol. 101, no.~3, pp. 437--458,
  2013.

\bibitem{wang2013head}
B.~Wang, W.~Liang, Y.~Wang, and Y.~Liang, ``{Head pose estimation with combined
  2D SIFT and 3D HOG features},'' in \emph{ICIG}, 2013.

\bibitem{Mukherjee2015}
S.~Mukherjee and N.~Robertson, ``{Deep Head Pose: Gaze-Direction Estimation in
  Multimodal Video},'' \emph{IEEE TMM}, vol.~17, no.~11, pp. 2094--2107, 2015.

\bibitem{drouard2017robust}
V.~Drouard, R.~Horaud, A.~Deleforge, S.~Ba, and G.~Evangelidis, ``Robust
  head-pose estimation based on partially-latent mixture of linear
  regressions,'' \emph{IEEE TIP}, vol.~26, no.~3, pp. 1428--1440, 2017.

\bibitem{ramanan2007learning}
D.~Ramanan, ``Learning to parse images of articulated bodies,'' in \emph{NIPS},
  2007, pp. 1129--1136.

\bibitem{andriluka14cvpr}
M.~Andriluka, L.~Pishchulin, P.~Gehler, and B.~Schiele, ``2d human pose
  estimation: New benchmark and state of the art analysis,'' in \emph{CVPR},
  June 2014.

\bibitem{HinSal06}
G.~Hinton and R.~Salakhutdinov, ``Reducing the dimensionality of data with
  neural networks,'' \emph{Science}, vol. 313, no. 5786, pp. 504 -- 507, 2006.

\bibitem{Erhan2010}
D.~Erhan, Y.~Bengio, A.~Courville, P.-A. Manzagol, P.~Vincent, and S.~Bengio,
  ``{Why Does Unsupervised Pre-training Help Deep Learning?}'' \emph{JMLR},
  vol.~11, pp. 625--660, 2010.

\bibitem{Yosinski2014}
J.~Yosinski, J.~Clune, Y.~Bengio, and H.~Lipson, ``{How Transferable Are
  Features in Deep Neural Networks?}'' in \emph{NIPS}, 2014.

\bibitem{Tajbakhsh2016}
N.~Tajbakhsh, J.~Y. Shin, S.~R. Gurudu, R.~T. Hurst, C.~B. Kendall, M.~B.
  Gotway, and J.~Liang, ``{Convolutional Neural Networks for Medical Image
  Analysis: Full Training or Fine Tuning?}'' \emph{IEEE TMI}, vol.~35, no.~5,
  pp. 1299--1312, 2016.

\bibitem{goodfellow2016deep}
I.~Goodfellow, Y.~Bengio, and A.~Courville, \emph{Deep learning}, 2016.

\bibitem{yao1999evolving}
X.~Yao, ``Evolving artificial neural networks,'' \emph{Proceedings of the
  IEEE}, vol.~87, no.~9, pp. 1423--1447, 1999.

\bibitem{mesejo2015artificial}
P.~Mesejo, O.~Ib{\'a}nez, E.~Fern{\'a}ndez-Blanco, F.~Cedr{\'o}n, A.~Pazos, and
  A.~B. Porto-Pazos, ``Artificial neuron--glia networks learning approach based
  on cooperative coevolution,'' \emph{International journal of neural systems},
  vol.~25, no.~04, 2015.

\bibitem{duchi2011adaptive}
J.~Duchi, E.~Hazan, and Y.~Singer, ``Adaptive subgradient methods for online
  learning and stochastic optimization,'' \emph{JMLR}, vol.~12, no.~7, pp.
  2121--2159, 2011.

\bibitem{tieleman2012lecture}
T.~Tieleman and G.~Hinton, ``Rrmsprop: Divide the gradient by a running average
  of its recent magnitude,'' \emph{COURSERA: Neural networks for machine
  learning}, vol.~4, no.~2, pp. 26--31, 2012.

\bibitem{zeiler2012adadelta}
M.~D. Zeiler, ``Adadelta: an adaptive learning rate method,'' \emph{arXiv
  preprint arXiv:1212.5701}, 2012.

\bibitem{kingma2014adam}
D.~Kingma and J.~Ba, ``Adam: A method for stochastic optimization,''
  \emph{arXiv preprint arXiv:1412.6980}, 2014.

\bibitem{sutskever13}
I.~Sutskever, J.~Martens, G.~Dahl, and G.~Hinton, ``On the importance of
  initialization and momentum in deep learning,'' in \emph{ICML}, 2013, pp.
  1139--1147.

\bibitem{chollet2015keras}
F.~Chollet \emph{et~al.}, ``Keras,'' \url{https://github.com/fchollet/keras},
  2015.

\bibitem{fisher1925}
R.~Fisher, \emph{Statistical methods for research workers}.\hskip 1em plus
  0.5em minus 0.4em\relax Edinburgh Oliver \& Boyd, 1925.

\bibitem{Nuzzo2014}
R.~Nuzzo, ``{Scientific method: Statistical errors},'' \emph{Nature}, vol. 506,
  no. 7487, pp. 150--152, 2014.

\bibitem{Vidgen2016}
B.~Vidgen and T.~Yasseri, ``P-values: Misunderstood and misused,''
  \emph{Frontiers in Physics}, vol.~4, p.~6, 2016.

\bibitem{Sterne2001}
J.~A.~C. Sterne, D.~R. Cox, and G.~D. Smith, ``{Sifting the
  evidence{\textemdash}what{\textquoteright}s wrong with significance
  tests?Another comment on the role of statistical methods},'' \emph{BMJ}, vol.
  322, no. 7280, pp. 226--231, 2001.

\bibitem{DerracGMH11}
J.~Derrac, S.~García, D.~Molina, and F.~Herrera, ``A practical tutorial on the
  use of nonparametric statistical tests as a methodology for comparing
  evolutionary and swarm intelligence algorithms.'' \emph{Swarm and
  Evolutionary Computation}, vol.~1, no.~1, pp. 3--18, 2011.

\bibitem{Dunn1961}
O.~J. Dunn, ``{Multiple comparisons among means},'' \emph{Journal of the
  American Statistical Association}, vol.~56, pp. 52--64, 1961.

\bibitem{Holm79}
S.~Holm, ``A simple sequentially rejective multiple test procedure,''
  \emph{Scandinavian Journal of Statistics}, vol.~6, no.~2, pp. 65--70, 1979.

\bibitem{Hochberg1988}
Y.~Hochberg, ``{A Sharper Bonferroni Procedure for Multiple Tests of
  Significance},'' \emph{Biometrika}, vol.~75, no.~4, pp. 800--802, 1988.

\bibitem{Hommel1988}
G.~Hommel, ``{A stagewise rejective multiple test procedure based on a modified
  Bonferroni test},'' \emph{Biometrika}, vol.~75, no.~2, pp. 383--386, 1988.

\bibitem{Holland1987}
B.~Holland, ``{An improved sequentially rejective Bonferroni test procedure},''
  \emph{Biometrics}, vol.~43, pp. 417--423, 1987.

\bibitem{Rom1990}
D.~Rom, ``{A sequentially rejective test procedure based on a modified
  Bonferroni inequality},'' \emph{Biometrika}, vol.~77, pp. 663--665, 1990.

\bibitem{Nemenyi63}
P.~Nemenyi, ``Distribution-free multiple comparisons,'' Ph.D. dissertation,
  Princeton University, 1963.

\bibitem{conover98}
W.~Conover, \emph{{Practical Nonparametric Statistics}}.\hskip 1em plus 0.5em
  minus 0.4em\relax Kirjastus: John Wiley and Sons (WIE), 1998.

\bibitem{JMLRsrivastava14a}
N.~Srivastava, G.~Hinton, A.~Krizhevsky, I.~Sutskever, and R.~Salakhutdinov,
  ``Dropout: A simple way to prevent neural networks from overfitting,''
  \emph{JMLR}, vol.~15, pp. 1929--1958, 2014.

\bibitem{iciploss}
M.~Carvalho, B.~L. Saux, P.~Trouvé-Peloux, A.~Almansa, and F.~Champagnat, ``On
  regression losses for deep depth estimation,'' in \emph{ICIP}, 2018, pp.
  2915--2919.

\bibitem{ioffe2015batch}
S.~Ioffe and C.~Szegedy, ``Batch normalization: Accelerating deep network
  training by reducing internal covariate shift,'' in \emph{ICML}, 2015, pp.
  448--456.

\bibitem{ba2016layer}
J.~L. Ba, J.~R. Kiros, and G.~E. Hinton, ``Layer normalization,'' \emph{arXiv
  preprint arXiv:1607.06450}, 2016.

\bibitem{Belagiannis2017recurrent}
V.~Belagiannis and A.~Zisserman, ``Recurrent human pose estimation,'' in
  \emph{2017 12th IEEE International Conference on Automatic Face \& Gesture
  Recognition (FG 2017)}.\hskip 1em plus 0.5em minus 0.4em\relax IEEE, 2017,
  pp. 468--475.

\bibitem{cao2016realtime}
Z.~Cao, T.~Simon, S.-E. Wei, and Y.~Sheikh, ``Realtime multi-person 2d pose
  estimation using part affinity fields,'' \emph{arXiv preprint
  arXiv:1611.08050}, 2016.

\bibitem{newell2016stacked}
A.~Newell, K.~Yang, and J.~Deng, ``Stacked hourglass networks for human pose
  estimation,'' in \emph{European Conference on Computer Vision}.\hskip 1em
  plus 0.5em minus 0.4em\relax Springer, 2016, pp. 483--499.

\bibitem{Nie_2018_CVPR}
X.~Nie, J.~Feng, Y.~Zuo, and S.~Yan, ``Human pose estimation with parsing
  induced learner,'' in \emph{CVPR}, June 2018.

\bibitem{tompson2014joint}
J.~J. Tompson, A.~Jain, Y.~LeCun, and C.~Bregler, ``Joint training of a
  convolutional network and a graphical model for human pose estimation,'' in
  \emph{Advances in neural information processing systems}, 2014, pp.
  1799--1807.

\bibitem{Chen2017AdversarialPA}
Y.~Chen, C.~Shen, X.-S. Wei, L.~Liu, and J.~Yang, ``Adversarial posenet: A
  structure-aware convolutional network for human pose estimation,''
  \emph{ICCV}, pp. 1221--1230, 2017.

\bibitem{Chu2017multi}
X.~Chu, W.~Yang, W.~Ouyang, C.~Ma, A.~L. Yuille, and X.~Wang, ``Multi-context
  attention for human pose estimation,'' \emph{arXiv preprint
  arXiv:1702.07432}, vol.~1, no.~2, 2017.

\bibitem{drouard2017head}
V.~Drouard, R.~Horaud, A.~Deleforge, S.~Ba, and G.~Evangelidis, ``Robust
  head-pose estimation based on partially-latent mixture of linear
  regressions,'' \emph{{IEEE TIP}}, vol.~26, no.~3, pp. 1428 -- 1440, 2017.

\bibitem{andriluka2009pictorial}
M.~Andriluka, S.~Roth, and B.~Schiele, ``Pictorial structures revisited: People
  detection and articulated pose estimation,'' in \emph{CVPR}, 2009, pp.
  1014--1021.

\bibitem{yang2013articulated}
Y.~Yang and D.~Ramanan, ``Articulated human detection with flexible mixtures of
  parts,'' \emph{IEEE TPAMI}, vol.~35, no.~12, pp. 2878--2890, 2013.

\bibitem{pishchulin2012articulated}
L.~Pishchulin, A.~Jain, M.~Andriluka, T.~Thorm{\"a}hlen, and B.~Schiele,
  ``Articulated people detection and pose estimation: Reshaping the future,''
  in \emph{CVPR}.\hskip 1em plus 0.5em minus 0.4em\relax IEEE, 2012, pp.
  3178--3185.

\bibitem{johnson2010clustered}
S.~Johnson and M.~Everingham, ``Clustered pose and nonlinear appearance models
  for human pose estimation,'' in \emph{BMVC}, 2010.

\bibitem{ouyang2014multi}
W.~Ouyang, X.~Chu, and X.~Wang, ``Multi-source deep learning for human pose
  estimation,'' in \emph{CVPR}, 2014, pp. 2329--2336.

\bibitem{Yang_2017_ICCV}
W.~Yang, S.~Li, W.~Ouyang, H.~Li, and X.~Wang, ``Learning feature pyramids for
  human pose estimation,'' in \emph{ICCV}, Oct 2017.

\end{thebibliography}

\begin{IEEEbiography}[{\includegraphics[width=1in,height=1.25in,clip,keepaspectratio]{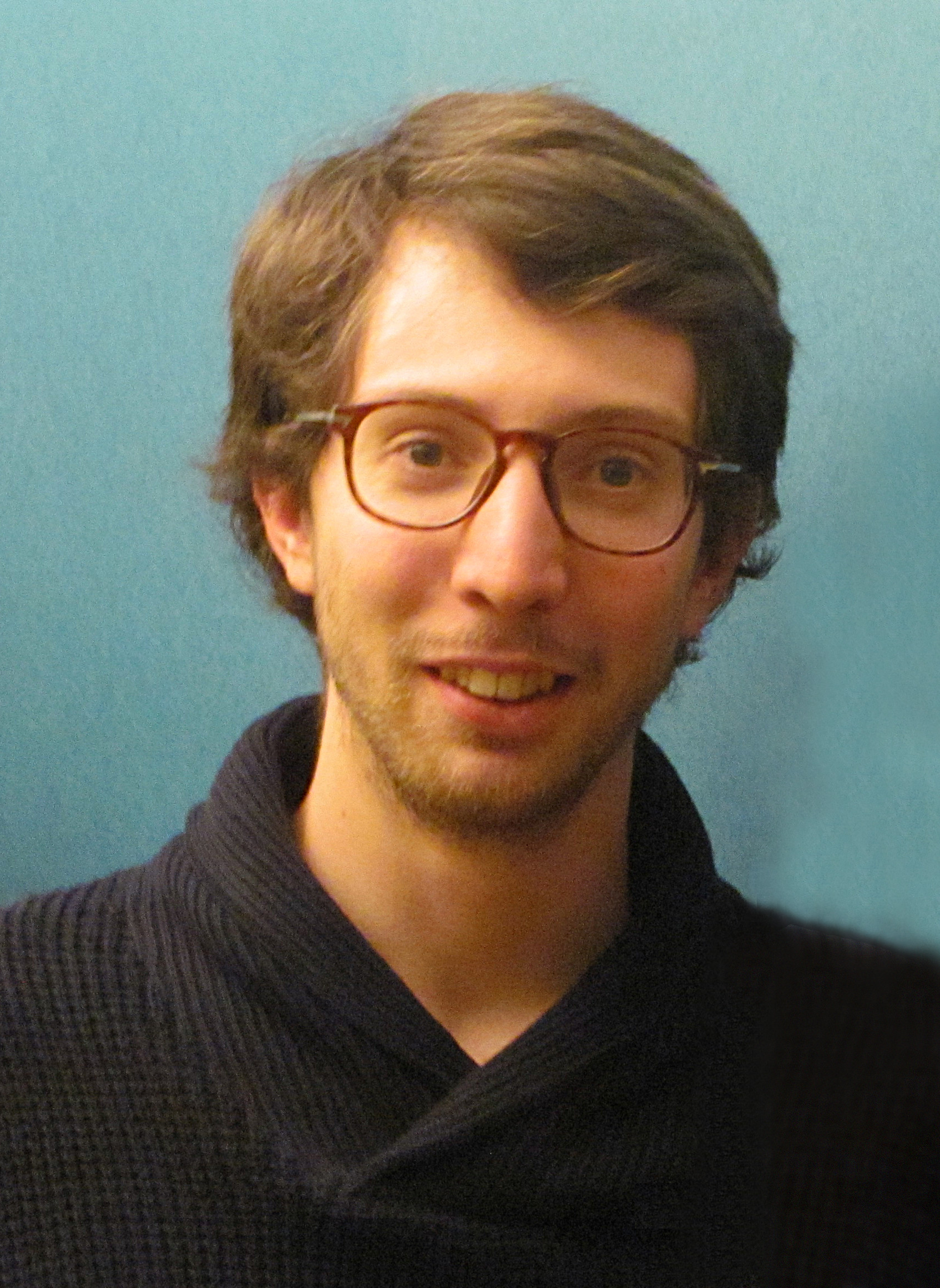}}]{St\'ephane Lathuili\`ere}
received the M.Sc. degree in applied mathematics and computer science from ENSIMAG, Grenoble Institute of Technology (Grenoble INP), France,
in 2014. He worked towards his Ph.D. in
mathematics and computer science in the Perception Team at Inria, and obtained it from Universit\'{e} Grenoble Alpes (France) in 2018. He is currently a Post-Doctoral fellow at the University of Trento. His research interests include machine learning for activity recognition, deep models for regression and reinforcement learning for audio-visual fusion in robotics.

\end{IEEEbiography}

\begin{IEEEbiography}[{\includegraphics[width=1in,height=1.25in,clip,keepaspectratio]{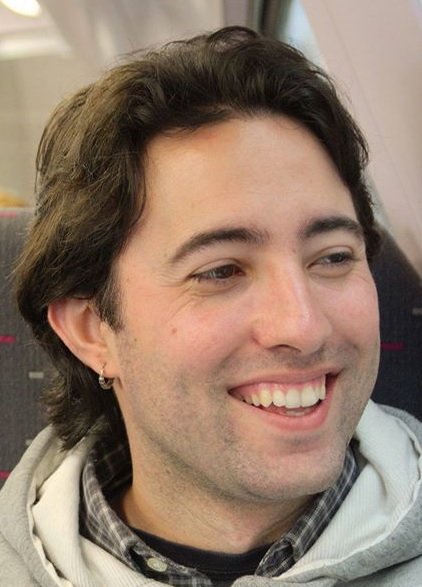}}]{Pablo Mesejo} received the M.Sc. and Ph.D. degrees in computer science respectively from Universidade da Coru\~{n}a (Spain) and Universit\`{a} degli Studi di Parma (Italy), where he was an Early Stage Researcher within the Marie Curie ITN MIBISOC. He was a post-doc at the ALCoV team of Universit\'e d'Auvergne (France) and the Mistis team of Inria Grenoble Rh\^{o}ne-Alpes (France), before joining the Perception team with a Starting Researcher Position. His research interests include computer vision, machine learning and computational intelligence techniques applied mainly to biomedical image analysis problems.

\end{IEEEbiography}

\begin{IEEEbiography}[{\includegraphics[width=1in,height=1.25in,clip,keepaspectratio]{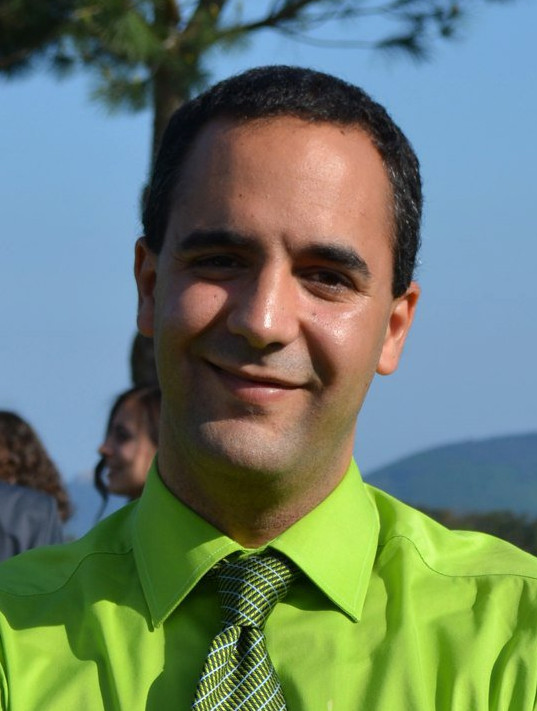}}]{Xavier Alameda-Pineda} received M.Sc.\ in Mathematics (2008), in 
Telecommunications (2009) and in Computer Science (2010). He obtained his Ph.D. in Mathematics and Computer Science from Universit\'e Joseph Fourier in 2013. Since 2016, he is a Research Scientist at Inria Grenoble, with the Perception Team. He served as Area Chair at ICCV 2017 and is the recipient of several paper awards. His scientific interests lie in computer vision, machine learning and signal processing for human behavior understanding and robotics.
\end{IEEEbiography}

\begin{IEEEbiography}[{\includegraphics[width=1in,height=1.25in,clip,keepaspectratio]{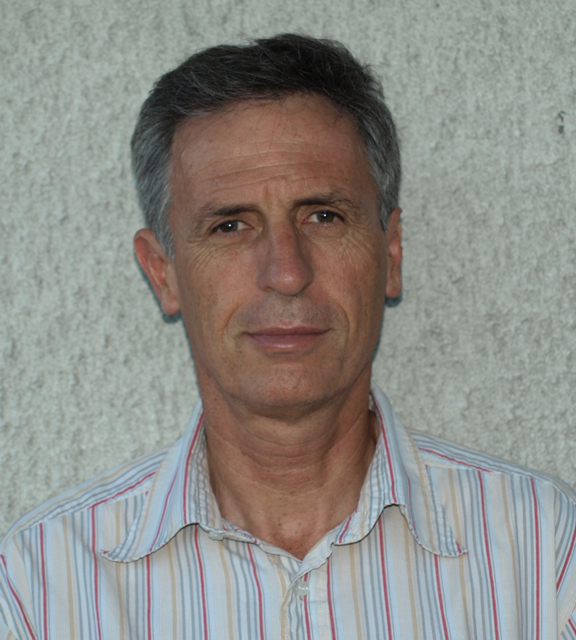}}]{Radu Horaud} received the B.Sc. degree in Electrical Engineering, the M.Sc. degree in Control Engineering, and the Ph.D. degree in Computer Science from the Institut National Polytechnique de Grenoble, France. Since 1998 he holds a position of director of research with INRIA Grenoble Rh\^one-Alpes, where he is the founder and head of the PERCEPTION team. His research interests include computer vision, machine learning, audio signal processing, audiovisual analysis, and robotics. Radu Horaud and his collaborators received numerous best paper awards. In 2013, Radu Horaud was awarded an ERC Advanced Grant for his project \textit{Vision and Hearing in Action} (VHIA) and in 2017 he was awarded an ERC Proof of Concept Grant for this project VHIALab.
\end{IEEEbiography}

\end{document}